\documentclass[]{article}
\usepackage[margin=1in]{geometry}

\usepackage{longtable}
\usepackage{url}  
\usepackage{booktabs}      
\usepackage{amsfonts}       
\usepackage{nicefrac}       
\usepackage{microtype}      
\usepackage{xcolor}         
\usepackage{amsmath}
\usepackage{mathtools}
\usepackage{amsthm}
\usepackage{amsmath}
\usepackage{amssymb}
\usepackage{bbm}
\usepackage{tikz} 
\usepackage{wrapfig}
\usepackage{graphicx}
\usepackage{floatrow}
\usepackage{enumitem}
\usepackage[suppress]{color-edits}
\usepackage{caption}
\usepackage{subcaption}
\usepackage{authblk}
\usepackage[colorlinks,allcolors=purple]{hyperref}
\PassOptionsToPackage{sort&compress}{natbib}
\usepackage[round, sort]{natbib}
\usepackage{booktabs} 
\usepackage[ruled]{algorithm2e} 
\usepackage{tikz}
\usetikzlibrary{arrows}
\usetikzlibrary{shapes.multipart}
\usepackage{caption}
\usepackage[all]{nowidow}
\usepackage[font=small,labelfont=bf]{caption}
\usepackage[utf8]{inputenc} 
\usepackage[T1]{fontenc}    
\usepackage[capitalise,nameinlink]{cleveref}
\usepackage{mdframed}
\usepackage{color,soul}
\usepackage[most]{tcolorbox}

\newtcolorbox{tabframebox}[1][]{breakable,sharp corners,boxrule=0.5pt,colback=white,halign=center,#1}

\def\cm{\texttt{CheckMate}}
\def\mc{\texttt{MathConverse}}

\title{Evaluating Language Models for Mathematics through Interactions}

\begin{document}


\author[a]{Katherine M. Collins\footnote{Equal contributions as leading authors. Correspondence: \texttt{\{kmc61, qj213\}@cam.ac.uk}.}}
\author[a]{Albert Q. Jiang{$^*$}}
\author[b]{\authorcr Simon Frieder}
\author[c]{Lionel Wong}
\author[a]{Miri Zilka}
\author[a,d,e]{Umang Bhatt}
\author[f,b]{Thomas Lukasiewicz}
\author[g]{\authorcr Yuhuai Wu\footnote{Work done while at Stanford.}} 
\author[c]{Joshua B. Tenenbaum}
\author[a]{William Hart}
\author[h,a]{Timothy Gowers}
\author[a]{Wenda Li}
\author[a,d,2]{\authorcr Adrian Weller\footnote{Equal contributions as senior authors.}}
\author[a,2]{Mateja Jamnik{$^{\ddagger}$}}

\affil[a]{University of Cambridge}
\affil[b]{University of Oxford}
\affil[c]{MIT}
\affil[d]{The Alan Turing Institute}
\affil[e]{NYU}
\affil[f]{Vienna University of Technology}
\affil[g]{x.AI}
\affil[h]{Collège de France}

\maketitle

\begin{abstract}
\noindent
There is much excitement about the opportunity to harness the power of large language models~(LLMs) when building problem-solving assistants. However, the standard methodology of evaluating LLMs relies on static pairs of inputs and outputs, and is insufficient for making an informed decision about which LLMs and under which assistive settings can they be sensibly used. Static assessment fails to account for the essential interactive element in LLM deployment, and therefore limits how we understand language model capabilities.
We introduce \cm, an adaptable prototype platform for humans to interact with and evaluate LLMs.
We conduct a study with \cm\space to evaluate three language models~(InstructGPT, ChatGPT, and GPT-4) as assistants in proving undergraduate-level mathematics, with a mixed cohort of participants from undergraduate students to professors of mathematics.
We release the resulting interaction and rating dataset, \mc.
By analysing \mc, we derive a
taxonomy of human behaviours and uncover that despite a generally positive correlation, there are notable instances of divergence between correctness and perceived helpfulness in LLM generations, amongst other findings.
Further, we garner a more granular understanding of GPT-4 mathematical problem-solving through a series of case studies, contributed by expert mathematicians. We conclude with actionable takeaways for ML practitioners and mathematicians: models that communicate uncertainty respond well to user corrections, and are more interpretable and concise may constitute better assistants. Interactive evaluation is a promising way to navigate the capability of these models; humans should be aware of language models' algebraic fallibility and discern where they are appropriate to use.
\end{abstract}

\newpage

\newpage 
\tableofcontents

\clearpage
\setlength{\parskip}{4pt}



\section{Introduction}
Foundation models~\citep{bommasani2021opportunities} -- in particular large language models~(LLMs)~\citep{brown2020language, touvron2023llama, anil2023palm} -- are increasingly human-facing, permitting users to interact with and elicit natural language responses~\citep{chatGPT, köpf2023openassistant}. Such interactive systems admit a plethora of new possibilities for human-machine collaboration~\citep{copilot, chatbotMed, dramatron}. 
However, existing evaluation approaches of LLMs that are used to help inform how to best harness their power in this human-machine interaction are insufficient. In particular, 
capability assessments for LLMs typically assume a non-interactive view: models are primarily evaluated statically with ``ground truth'' input - output pairs, and metrics are aggregated over a dataset~\citep{burnell2023}, which may be misaligned with their use cases.

\begin{figure*}[t!]
\begin{subfigure}[b]{.45\textwidth}
  \centering
  \includegraphics[width=.9\linewidth,keepaspectratio]{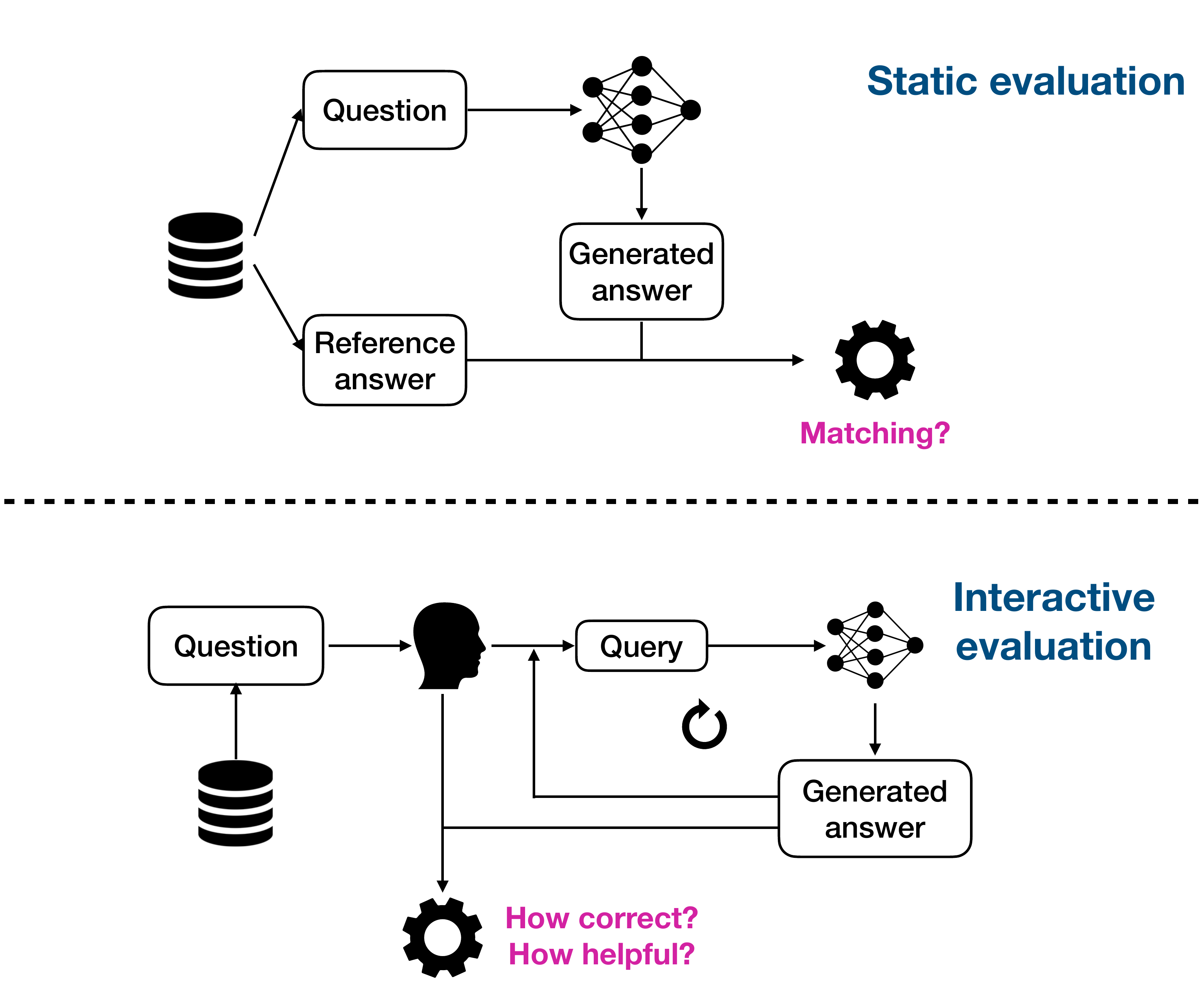}
  \caption{}
  \label{fig:ipe}
\end{subfigure}%
\begin{subfigure}[b]{.55\textwidth}
        \centering
        \includegraphics[width=\linewidth]{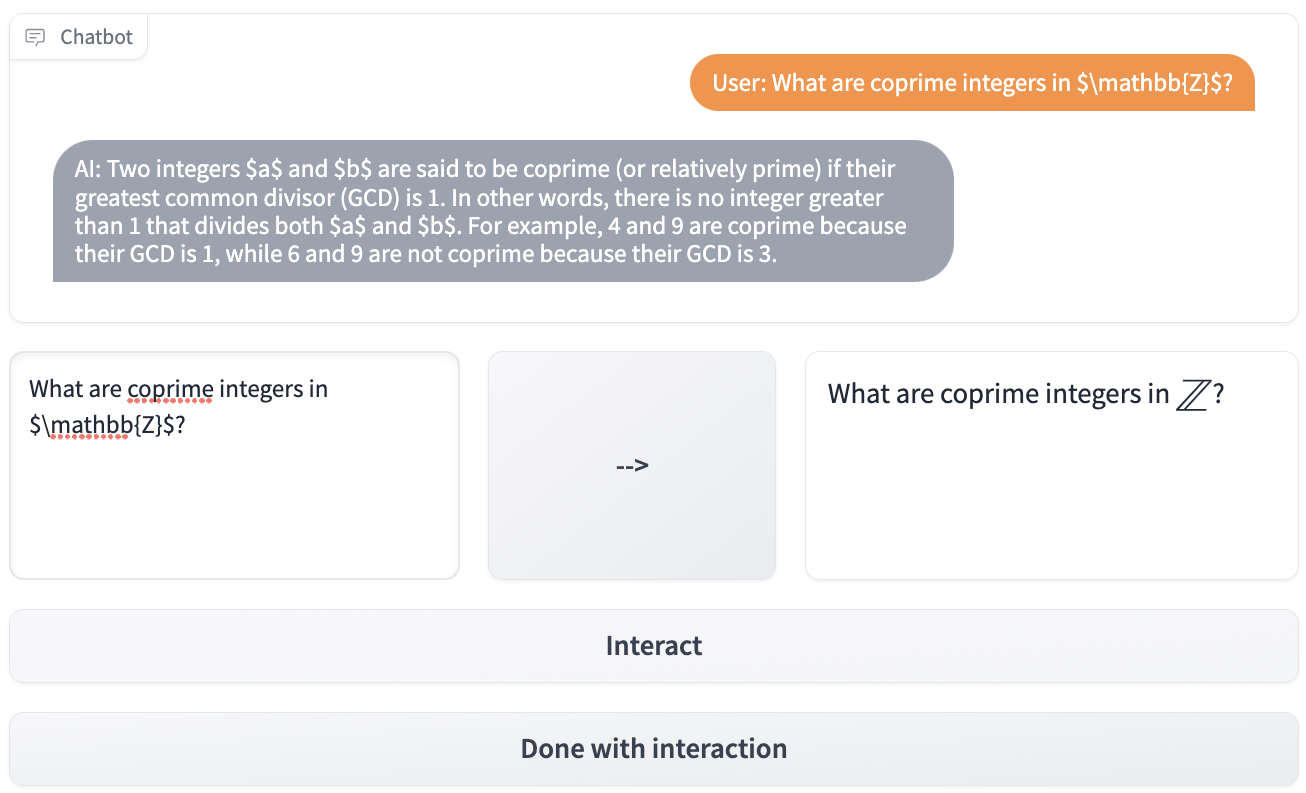}
        \caption{}
        \label{fig:checkmateinterface}
    
\end{subfigure}
\caption{\textbf{a)} Contrasting typical static evaluation (top) with interactive evaluation (bottom), wherein a human iteratively queries a model and rates the quality of responses. \textbf{b)} Example chat interface from \texttt{CheckMate} to interact with an LLM. The participant is presented with the problem and a reminder of particularly relevant instructions, and enters their chat below (which they can see compiled in LaTeX). Chat history is presented for the user to refer to.
}
\label{fig:interface}
\end{figure*}

To address this problem, we argue in this paper that \textit{interactive} and \textit{dynamic} evaluation of LLMs~\citep{leeInteractive, shen2023parachute} is essential for grasping their capabilities. We carry out such an interactive and dynamic evaluation of how humans use LLMs for assistance~(see Figure~\ref{fig:ipe} for an overview of how interactive evaluation differs from conventional static evaluation) to better characterise their limitations, undesirable behaviours, and potential harms.

Evaluating LLM interactions is especially warranted in the case of informal mathematical theorem proving, wherein an agent is given a mathematical theorem and needs to propose a proof that is acceptable to the mathematical community.
Informal theorem proving is special in that there is a formal notion of correctness at its core, yet most things are expressed in natural language~(informally). Important quality measures for the task include helpfulness and correctness, neither of which can be satisfactorily captured by automatic metrics~(e.g., BLEU and ROUGE scores)~\citep{papineni2002bleu, lin2004rouge, golovneva2022roscoe}. Letting humans communicate and evaluate these systems is crucial for their assessment. 
Further, mathematics is an interactive practice. Recent works~\citep{welleck2022naturalprover, first2023baldur} have shown that LLMs can benefit from feedback on their previously-produced language output (i.e., ``generations'') in mathematical tasks, and this benefit can only be seen in multi-turn interactive evaluations.
Hence, we choose mathematics to best highlight the value of human interactive evaluations. But \cm\space and our broader evaluation methodology can be extended to other domains and chatbot-based systems (see Supplement).

Concretely, we apply two methods to analyse human-LLM mathematical reasoning interactions: (1) structured evaluation, that is, rating every LLM generation in a conversation; and (2) free-form evaluation, where expert mathematicians conduct instance-level case studies. The latter approach is inspired by the burgeoning literature of involving domain experts alongside ML practitioners in understanding model behaviour~\citep{mcgrath2022acquisition, davies2021advancing}, and directly responds to the call from Burnell et al.~\citep{burnell2023} to engage with non-ML scholars to better understand LLM systems. Our study is interdisciplinary at its core.

Despite the large number of LLM-based chatbots, there is a paucity of open and unified platforms for eliciting fine-grained evaluations of interactions with users at scale.
Hence, we develop a lightweight interactive evaluation platform that is highly adaptable, called \cm\footnote{The name alludes to the interactive manner of the evaluation, or ``checking'', to be the kind you may do with a ``mate''.}.
We leverage \cm{} to conduct an empirical study on undergraduate-level theorem proving~(see an example problem in \textit{SI An Example Survey Problem}), over a suite of popular language models -- InstructGPT~\citep{ouyang2022training}, ChatGPT~\citep{chatGPT},\footnote{When we refer to ChatGPT in the text, we mean ``gpt-3.5-turbo'', according to the OpenAI \href{https://platform.openai.com/docs/api-reference/chat/create}{API}.} and GPT-4~\citep{openai2023gpt4}. We release the resulting interactions and evaluations on 261 human-model interactions in a new dataset called \mc, from which we derive a preliminary taxonomy of user behaviours. 
We do not claim completeness for our taxonomy, because of the limited size of \mc.
Our study is particularly compelling as it not only engages a group of participants with a wide range of mathematical experience, but the level of problem difficulty is higher than what is typically explored~\citep{gsm8k, mathqa, minif2f}.
We emphasise that \texttt{CheckMate} can be conveniently extended to domains other than mathematics. 
We also invite three expert mathematicians to contribute in-depth interaction case studies to help better characterise current LLM mathematical reasoning capabilities. Throughout, we emphasise that we are not trying to draw broad conclusions across the entire LLM landscape. Rather, we aim to highlight the feasibility and value of incorporating interactions into the evaluation process, particularly when involving domain experts, and to elucidate potential human and model behaviour patterns specifically in mathematics.\\

\noindent\textbf{Our three key contributions are:} 
\begin{itemize}[leftmargin=*]
    \setlength\itemsep{0em}
    \item We introduce an adaptable platform, \cm, for evaluating language models by their interactions with human users.
    We demonstrate that \textit{scalable} and \textit{valuable} dynamic interactive evaluations are feasible by applying \cm{} to evaluate three language models on mathematical theorem proving.
    \item With interactions and evaluations collected from \cm{} via a mixed cohort study, we derive a taxonomy of user behaviours which identify crucial expected abilities of LLM-based mathematical assistants.
    We release the dataset of \cm{} interactions and evaluations, \mc.\footnote{They can be found on:  \href{https://github.com/collinskatie/checkmate}{https://github.com/collinskatie/checkmate}.}
    \item Through case studies conducted by expert mathematicians, we add empirical evidence for several weaknesses of the LLMs that we explore, including algebraic manipulations, over-verbosity, and over-reliance on memorised solutions. We urge solutions from ML practitioners to these challenges (such as better communication of uncertainty and ability to update user corrections) and suggest good practices for LLM users (e.g., to heed caution when inspecting generations, as mistakes can be subtle). We encourage further interactive evaluation with LLMs, in mathematics and beyond, to inform how, when, and whether to deploy these models in assistive settings.
\end{itemize}

\section{Results}

We first present results from both of our evaluation methods -- structured multi-step interactive ratings and the free-form instance-based evaluation (see Figure~\ref{fig:test}) -- before synthesising the key insights across studies.

\subsection{Observations from Applying \texttt{CheckMate}} 

We highlight our primary findings from applying \texttt{CheckMate} in the domain of mathematics, through which we collect and introduce the \texttt{MathConverse} dataset (see example interface screen in Figure \ref{fig:checkmateinterface} and the \textit{Supplementary Information (SI)}).

\subsubsection*{Systems Optimised for Chat are Preferred}
\label{preferences}

Participants were \textit{not informed} of which model they were interacting with and rating. Nonetheless, we nicely observe in Figure~\ref{fig:post-int-prefs} that, as expected, models optimised for chat (ChatGPT and GPT-4) are consistently rated as preferable to those not (InstructGPT) with GPT-4 being most frequently favoured and much less often least preferable; i.e., the lower bound (``worst case'') behaviour of \mbox{GPT-4} is consistently better than the other models (e.g., the model is never ranked as the worst assistant). We emphasise that these evaluations are not meant to be definitive assessments of model performance, but rather, highlight that interactive evaluation can yield a more nuanced understanding of model behaviour beyond the common ``snapshot'' evaluation on standard benchmark datasets.



\subsubsection*{Perceived Utility Per Model}

We next look at individual interactions. Participants were asked to rate the mathematical correctness and perceived helpfulness of each generation; we depict the helpfulness and correctness ratings across models in Figure~\ref{fig:decomp-ratings-by-model}. These data further reveal distinctions across models; notably, GPT-4 achieves consistently high helpfulness ratings, underscoring its potential perceived utility.

Here, the generation\footnote{This generation is from GPT-4; recall, participants were not told which model they were interacting with and rating, or even which models they could have been interacting with.} is rated to have ``Multiple critical maths errors''~(correctness score 2), while being ``Somewhat useful''~(helpfulness score 4): This response from the assistant is indeed plagued with errors and misunderstandings, but it also contains the structure of a correct proof as well as the Rank-Nullity Theorem which is useful if used properly. 

We observe in Figure~\ref{fig:corr_v_help} that the perceived helpfulness and correctness correlate positively for all three models, although interestingly some
generations can be deemed completely mathematically correct, but not particularly helpful (e.g., 6 correctness, 3 helpfulness). This can occur, for instance, for overly verbose responses, e.g. the example in \textit{SI An Overly Verbose Response}.

These data further assert the value of multi-factorial LLM evaluations, beyond a single scalar ``goodness'' score. We expand on these two points in our expert case studies.

\subsection{Taxonomising User Behaviour from \texttt{MathConverse}}

Additionally, we qualitatively study the \textit{characteristics} of the participants' interactions with the LLMs. A core motivation 
is to get a
grasp of how mathematicians actually \textit{use} these systems in the wild -- what queries they may initiate, and how they follow-up with the model over the course of the interaction.  Following Lee et al.~\citep{lee2022evaluating}, we offer a preliminary taxonomy of the kinds of queries found in our data. All interaction traces are released anonymously in our repository to support further study of human-machine interaction, along with our annotated taxonomy. Details on how we constructed and annotated such a taxonomy are included in the Methods and \textit{SI Additional Details on Taxonomy Construction}. 


\paragraph{Initial Interaction Behaviour} 

We find that participants typically take one of four approaches in their first query to the AI assistant. Remarkably, \textbf{over 90\% of participants' first interaction for each problem fall into one of the following interaction behaviour types}:

\begin{enumerate}
    \item Seeking specific definitions of a concept mentioned in the problem (e.g., ``\texttt{Definition of Hall subgroup}'' or ``\texttt{What is the definition of `nullity' in linear algebra?}'').
    \item Asking a general question about mathematics related to the problem (e.g., ``\texttt{When is a plane in  $\mathbb{R}^3$ parallel to another plane in  $\mathbb{R}^3$}'' or ``\texttt{In mathematics, what does it mean to Let $A \in K^{n \times n}$}'').
    \item Simply copy-pasting the entire problem statement, or a slight rephrasing of the original statement, optionally with prepended instructions (e.g., ``\texttt{Can you assist me in proving the following statement? [...]}'').
    \item Prompting the model for a single step of the problem,  rather than the entire problem all at once (e.g.,  ``\texttt{We will first prove a lemma, let us call it Lemma 1 [...]}'').
\end{enumerate}
%

\paragraph{Mid-Interaction Behaviour} 

We observe a wider array of interaction modes after the first interaction. In addition to repeats of the above, such as users asking for more definitions, we find the following general patterns:

\begin{enumerate}[start=5] 
    \item Asking a clarifying question (e.g., ``\texttt{Does it hold even when $p$ is not a prime number?}'').
    \item Correcting the model output, occasionally with a clarifying question (e.g., ``\texttt{I understand. But your example is misleading. In your example, $f$ has degree $2$ and it has $2$ roots, so it does not represent a valid counterexample. Can you show an example in which a polynomial has more roots than its degree?}'').
    \item Asking for clarification about the generation from the model (e.g., what a particular symbol means — ``\texttt{What is $\tau$ here?}'').
    \item Asking \textit{why} the model did something (e.g., ``\texttt{so why do you need to add the whole set at step 2?}'').
    \item \textit{Implicitly} correcting the model (e.g., ``\texttt{That sounds like there being a homeomorphism. But a contraction is not a homeomorphism?}'').
    \item Asking for instances of a particular construction (e.g., ``\texttt{Can you exhibit an example to demonstrate that?}'').
\end{enumerate}

We also find that a few participants would ask the model to ``continue'' if it stopped midway through, and some participants seemed to attempt popular prompt engineering tricks, for example, attempting to get the model to ``restart'' by telling it to forget what it had done before: ``\texttt{Forget what you've said before and try again. Start with $n^x + n^y = n^z$, divide both sides by $n^z$, and reason from there.}''. Further, we note that one participant asked for \textit{intuition} about a concept: ``\texttt{What is the intuition behind the proof of this statement?}''

Here, the model (GPT-4) provided a response rated as ``definitely helpful'', indicative of a potential exciting assistive case going forwards. In addition to revealing the kinds of interactions that mathematicians may make to help motivate the design of tools better equipped to handle such interactions (e.g., when participants ask for clarification), we see these observations as pointers to broader public education as to what AI systems can be leveraged to help with and how to best query for this help (such as particular prompt techniques).

\subsubsection*{Rating Dynamics over the Interaction Trace}

\begin{figure*}[t!]
\centering
\begin{subfigure}{.4\linewidth}
\centering
    \includegraphics[width=\linewidth]{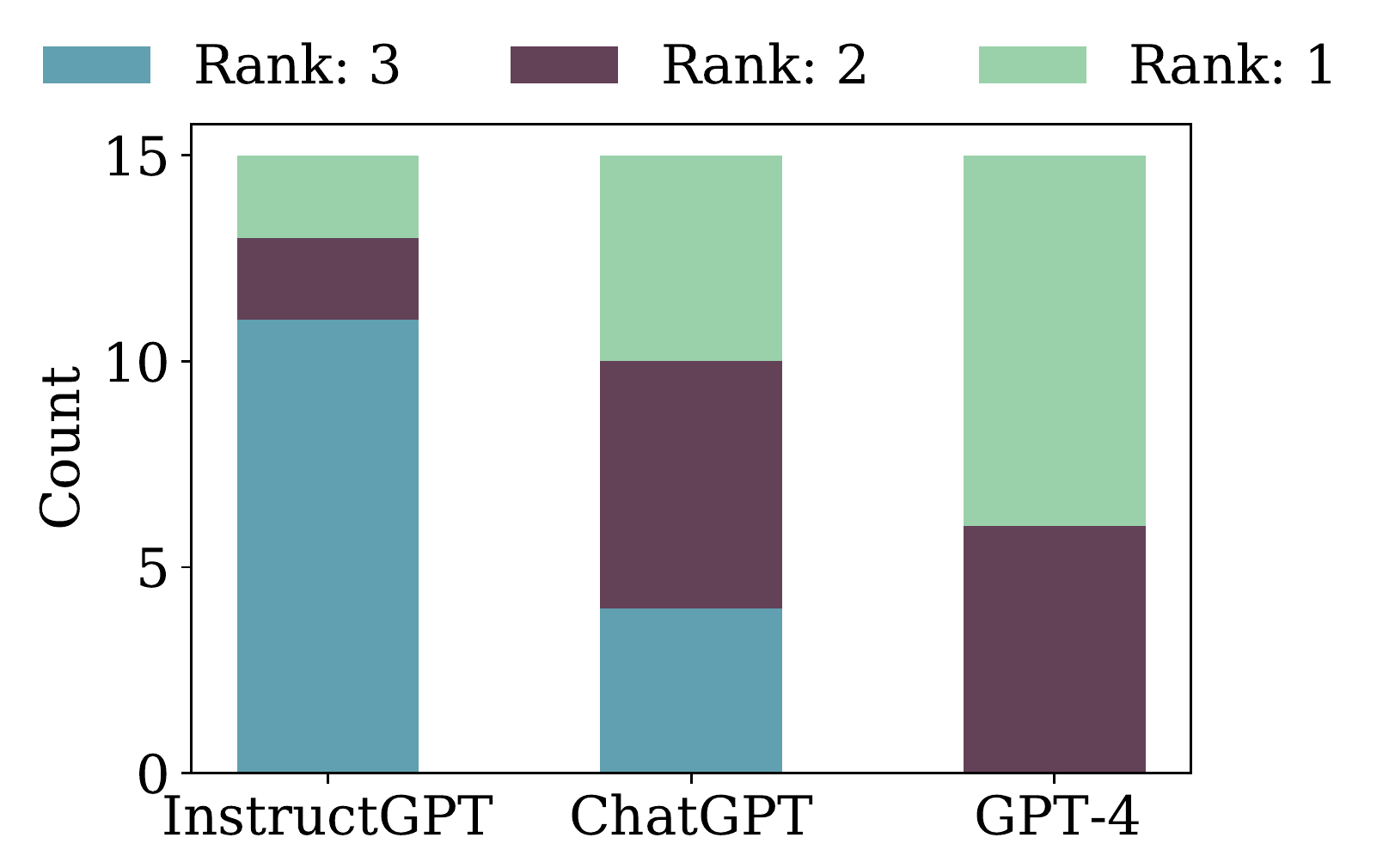}
    \caption{}
    \label{fig:post-int-prefs}
\end{subfigure}
\begin{subfigure}{.5\linewidth}
\centering
    \includegraphics[width=\linewidth]{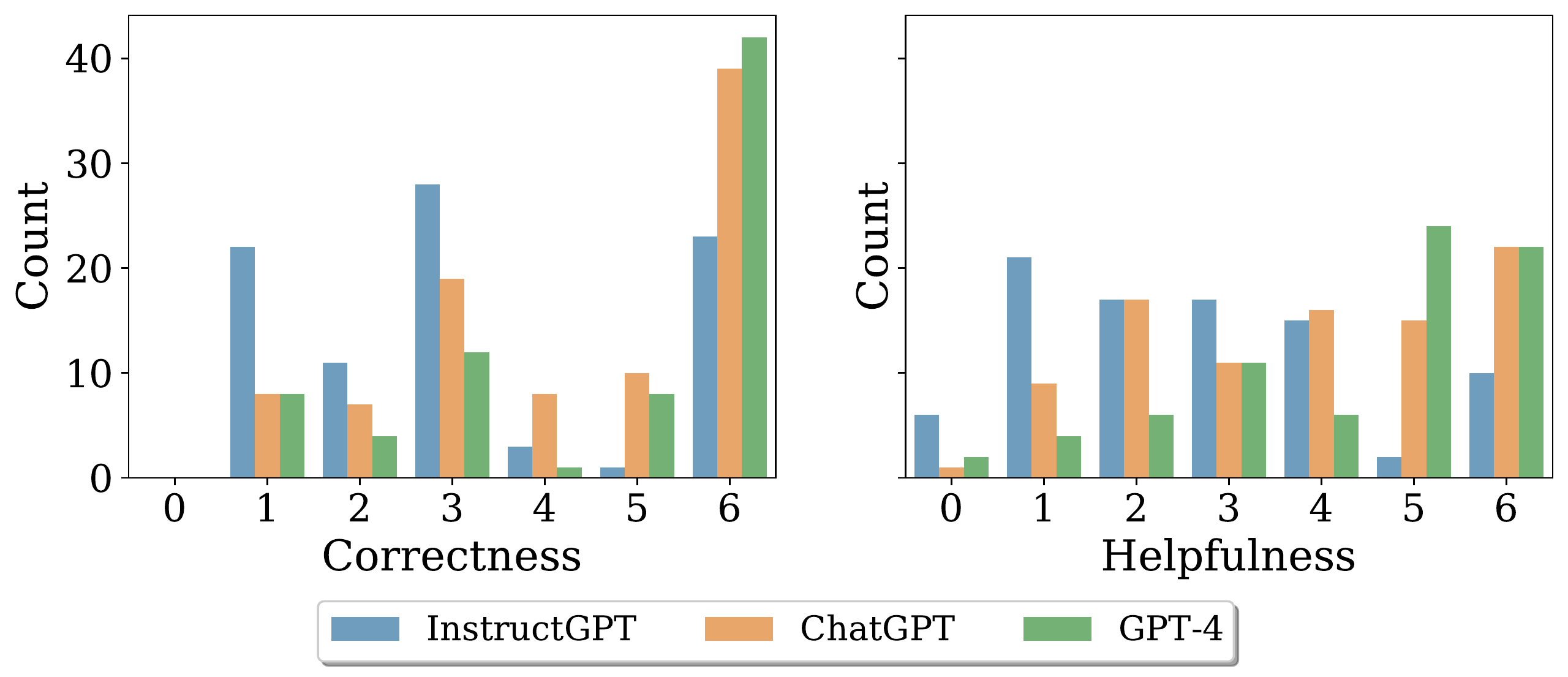}
    \caption{}
    \label{fig:decomp-ratings-by-model}
\end{subfigure}
\begin{subfigure}{0.55\linewidth}
  \centering
    \includegraphics[height=2in]{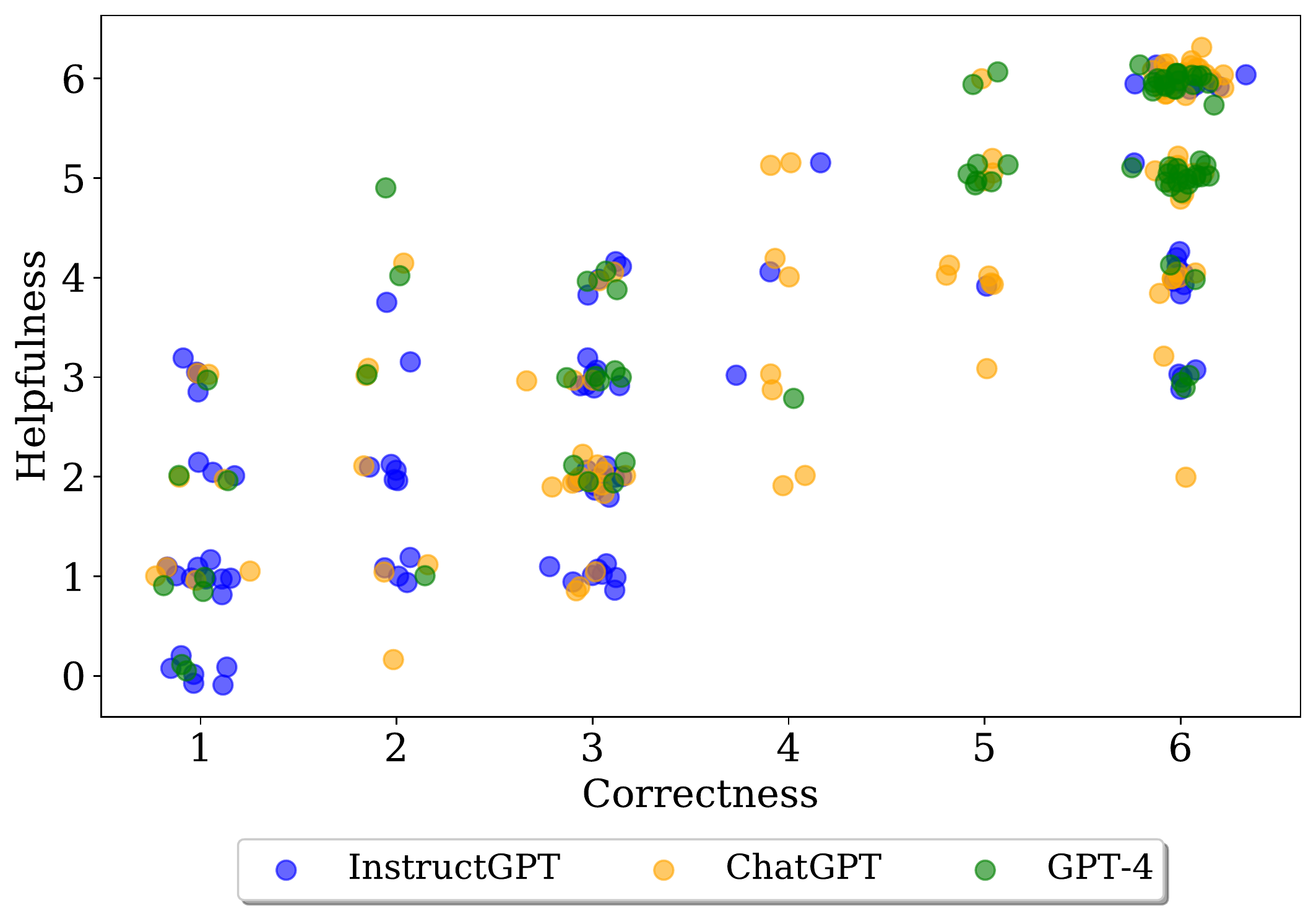}
    \caption{}
    \label{fig:corr_v_help}
\end{subfigure}
\begin{subfigure}{0.43\linewidth}
\vspace{0.2in}
    \includegraphics[height=2in]{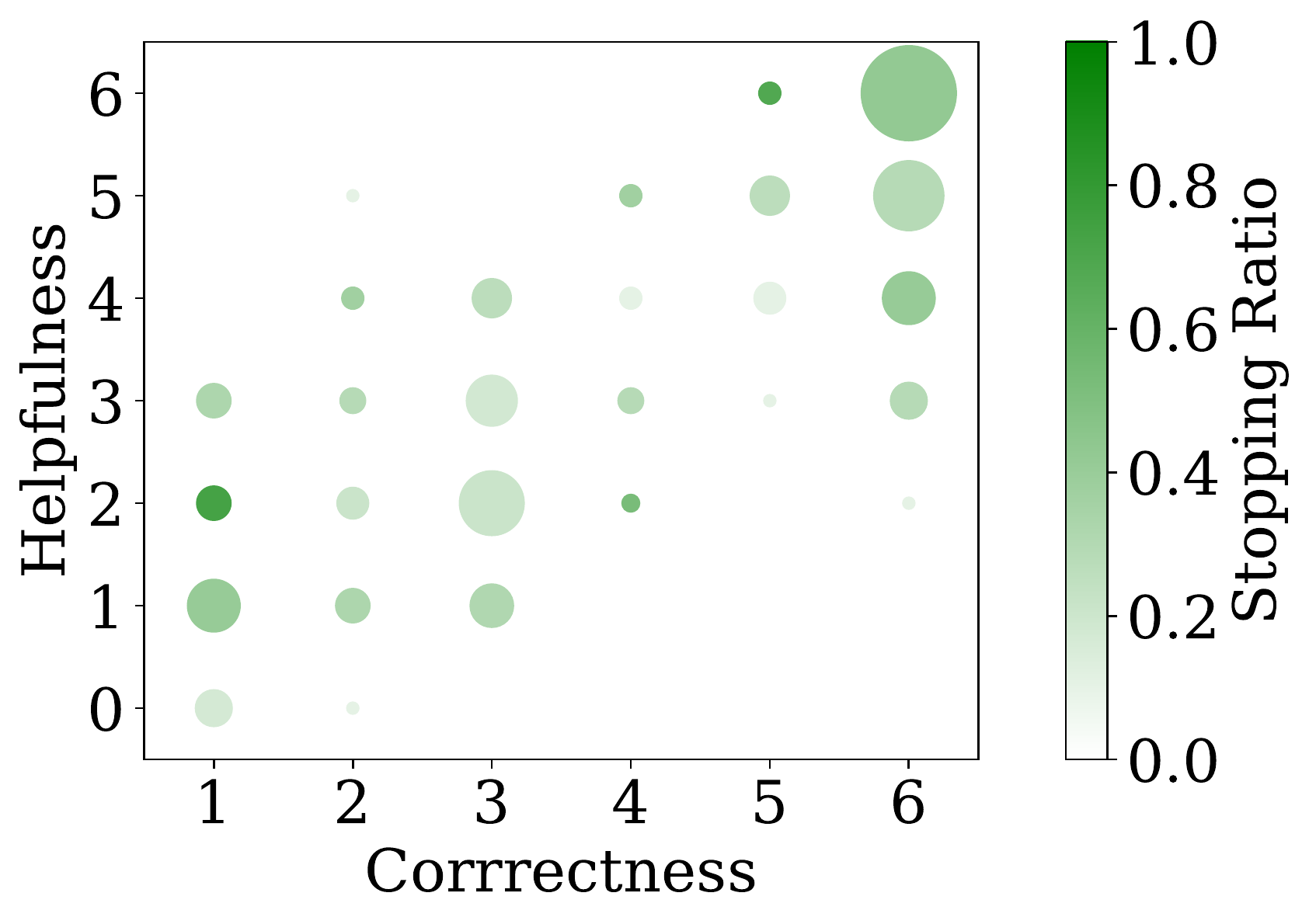}
    \caption{}
    \label{fig:why-stop}
\end{subfigure}%
\caption{\label{fig:test}(a) Post-interactive ranks across models about which model the participant preferred as a mathematical assistant (lower rank is better). Ties were allowed and are included: participants were permitted to assign the same rank to multiple models (see \textit{SI Additional Survey Observations}). (b) Mathematical correctness and perceived helpfulness scores (all scores are an integer $\in \{0, 1, ..., 6\}$; higher is better) received for each model. Full details about the text associated with the scales of each score are included in the \textit{SI Additional Survey Details}. (c) Comparing participants' scores of the mathematical correctness against perceived helpfulness of each models' generations. 
 Each dot is a score for a single human-model interaction. We add slight jitter for visual ease given that points overlap. Interestingly, we observe cases where the perceived helpfulness and correctness of a generation diverge, i.e., particular instances can be deemed incorrect yet somewhat helpful, or correct, but somewhat unhelpful. (d) The relationship between correctness and helpfulness scores and whether the step is terminal (i.e., the step after which the participant stopped interacting for a particular problem). The size of the bubbles indicates the number of that particular score pair (correctness, helpfulness). For a fixed score pair, the opacity indicates the ratio of stopping steps, i.e., the number of terminal steps divided by the number of total steps.}
\end{figure*}

As noted, we observe that several participants attempt to correct the model's output or ask for clarification. Sometimes these occurrences would go on for a few successive trials; we refer to such correction-mistake interaction ruts as ``frustration cycles''. We can see some of this behaviour by inspecting the rating dynamics across interaction traces. In \textit{SI Figure S6}, we see that in general, participants' ratings begin to fall off over the course of interactions, and through 
Figure~\ref{fig:why-stop},
we see that participants seem to stop when both ratings are higher than 4, indicating the model clearly can solve the problem and ``assist'' them), or with both ratings lower than 2 (indicative of the model completely failing to provide any further useful mathematical knowledge). We include participant testimonials about why they chose to stop in  \textit{SI Post-Survey Testimonials from Participants}. 

\subsection{Investigations into the \texttt{MathConverse} Annotated Taxonomy}
\label{query taxonomy ratings}

We build an annotated taxonomy by coding each user query; details are included in the Methods below, as well as in \textit{SI Additional Details on Taxonomy Construction}. The taxonomy enables us, and other researchers, to understand the kinds of queries users make. We ask a few questions here: (1) how do the queries made in the first interaction compare to those in the second; (2) is there a relationship between the kinds of queries made by participants who had extensive prior experience interacting with AI systems versus those who did not; (3) and is there a relationship between particular interaction types and the scores assigned to the models' corresponding responses?

\begin{figure}[t!]
    \centering
    \begin{subfigure}{\linewidth}
        \centering
        \includegraphics[width=\linewidth]{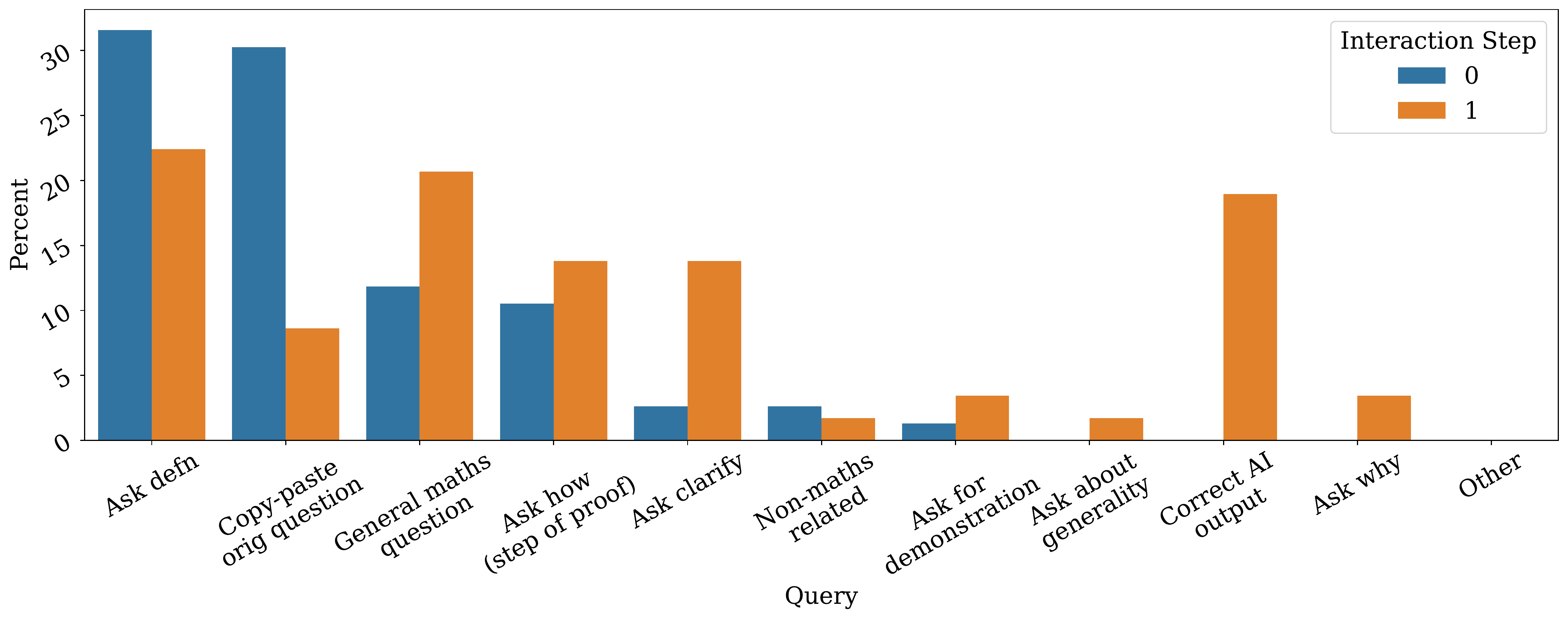}
        \caption{}
        \label{fig:timestep_query_profile}
        \vspace{0.1in}
    \end{subfigure}
    
    \begin{subfigure}{\linewidth}
        \centering
        \includegraphics[width=\linewidth]{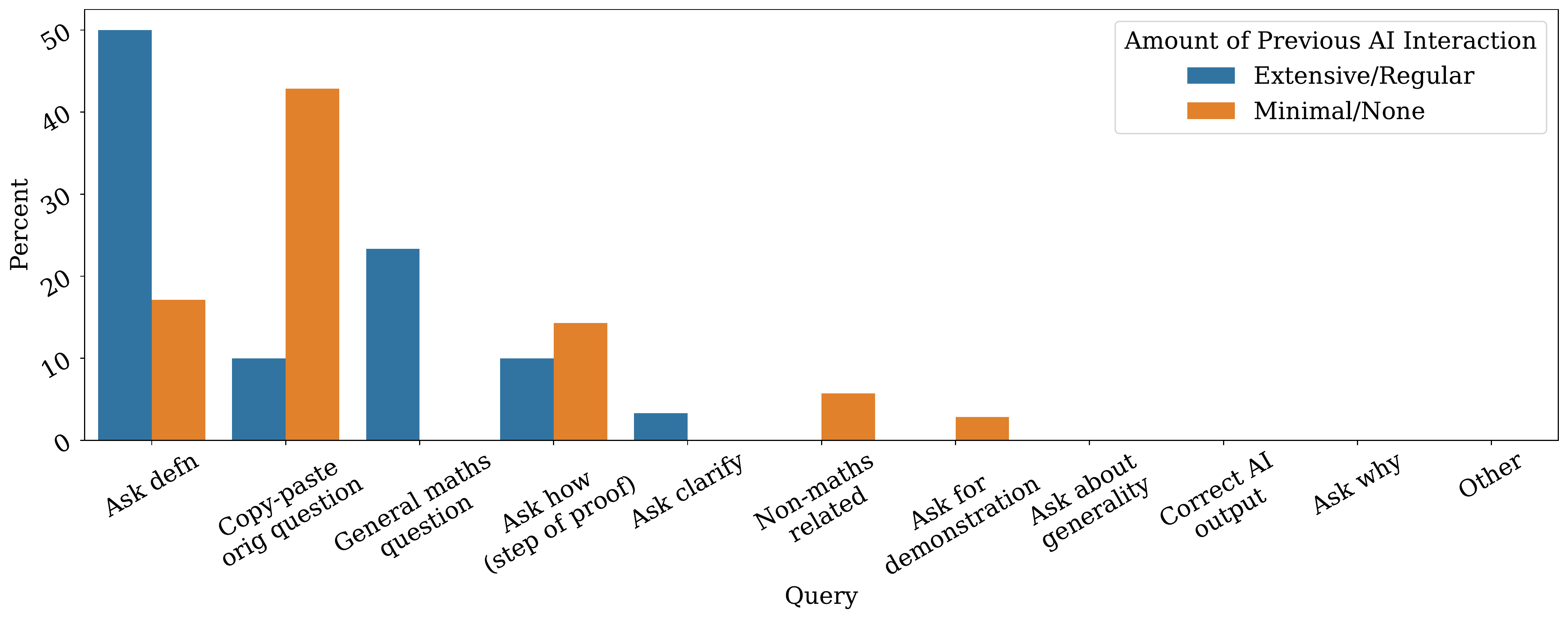}
        \caption{}
        \label{fig:expertise_query_profile}
    \end{subfigure}
\caption{(a) Query profiles as a function of the interaction step. Users prefer to ask for definitions or general mathematics questions, and to paste in the full text, in the first interaction, compared to correcting the model's output, asking why, etc. in the second interaction. Interaction step 0 is the initial interaction; step 1 is the query made after receiving the first AI response back to the query made in step 0. (b) Query profiles -- for the first interaction step (i.e., step 0) -- as function of the amount of experience the user has with AI systems prior to participating.}
\label{fig:taxonomy_query_profiles}
\end{figure}

We address questions (1) and (2) by investigating ``query profiles'' -- a ``signature'' of the kinds of queries made by a user (or subset of users) over a particular interaction duration. We compare query profiles for the kinds of queries participants made in the first interaction versus the second in Figure~\ref{fig:timestep_query_profile} to address question (1). We observe a marked difference across query profiles between the first and the second interaction, indicating the importance of studying AI systems behaviour through interactions; query patterns can \textit{evolve} in time. 

To address question (2), we notice in Figure~\ref{fig:expertise_query_profile} that users who had minimal prior AI expertise (i.e., responded that they either had never interacted with an AI system, or had only done so rarely) were more likely to simply paste in the full prompt text, in contrast to those with more experience with AI systems who more regularly asked for decomposed tasks (e.g., asking for a definition or help on a single step of the proof). These differences in behaviour hint at the potential importance of improving general AI literacy across users about the kinds of prompting behaviour that induces desirable responses.


We explore question (3) in \textit{SI Additional Investigations in the \texttt{MathConverse} Taxonomy} and find that particular queries are associated with differential correctness and helpfulness ratings, hinting at the importance. Further systematic study of particular query types and model performance for queries involving human corrections remain future work.

\subsection{Qualitative Observations from Case Studies}

We next include takeaways provided by each expert in relation to their freeform interactions. We then synthesise the core insights across the case studies.

\subsection{Per-Expert Conclusions} 
Each domain expert provided takeaway remarks following their interactions with GPT-4.

\subsubsection*{Dr. William Hart}
Whilst GPT-4 is able to regurgitate some very commonly found elementary number theoretical material and can handle straightforward problems, it has a major difficulty with algebraic manipulation and little or no ability to work on unseen problems that require backtracking, proving intermediate lemmas or extensive planning.

This is clear when GPT-4 is asked to show that \[3 = \sqrt {1 + 2 \sqrt {1 + 3 \sqrt { 1 + \cdots} } }. \tag{\textbf{SI ProofWiki prob. 28}}\]

A consistent problem here is an inability to write down a correct expression for a recursive relation to describe the nested radical. GPT-4 seems to be convinced that the expression under each square root is the same, so that if we write the initial expression $3 = \sqrt{A}$ then we also have $3 = \sqrt{1 + 2\sqrt{A}}$ and $3 = \sqrt{1 + 2\sqrt{1 + 3\sqrt{A}}}$, etc. To probe a little deeper, GPT-4 was instead prompted in a direction that might allow it to make partial progress. The hint was given to try peeling the expression on the right hand side one square root at a time, working backwards from the desired result that the full nested radical should have the value 3 to see if some pattern could be found in the values of the inner nested radicals. It was easy to prompt it so that it heads in that direction but on every generation it made hopeless algebraic and numerical errors, once again illustrating that very often what holds it back is high school algebra rather than the depth of the mathematics.

\subsubsection*{Dr. Wenda Li}
We found GPT-4's performance at variations of several ProofWiki problems quite satisfactory: it can reliably retrieve definitions of concepts used in the problem as well as in its own proof; it can correctly assess whether loosening certain assumptions breaks the proof; it can also instantiate variables quite robustly, given the opportunity of inspection of its own answers.
There have been debates~\citep{stochasticParrots, piantadosi2022meaning} about to what extent, shall we say, language models ``understand'', given the nature of their stochastic generation. In our study, we find a couple of simple\footnote{Note, ``simple'' here is relative to that of a trained mathematician.} cases where the language-model-based assistant possesses the mathematical understanding of assumptions and variable instantiations beyond mere memorisation.

For instance, we asked GPT-4 to solve the following standard probability theory problem: Let $X$ be a random variable. Assume $\mathsf{E} (X) = \mu$ for some $\mu \in \mathbb{R}$ and $\mathsf{var} (X) = \sigma^2$ for some $\sigma^2 \in \mathbb{R}_{> 0}$. Show that for all $k > 0$: $\Pr \left({\left| {X - \mu}\right| \geq k \sigma}\right) \leq \dfrac {1}{k^2}.$
GPT-4 started by stating that we can use Chebyshev’s inequality, and then re-stated the problem in an almost identical way but with different variable names. This demonstrates a
certain level of variable unification. 
We then checked whether the assistant knew how to instantiate variables by asking it whether the proof still holds when the following concrete values are assigned to $k$: $2$, $\sqrt{2}, \sqrt{2}-1, \sqrt{2}-2$, and $(\sqrt{2}-2)^2$. Human inspection finds the assistant's behaviour to be correct.
The assistant can handle concrete calculations even when $k$ is a relatively complicated number (e.g., $\sqrt{2} - 1$). The model also knows that the previous derivation cannot be carried out when $k=\sqrt{2} - 2$, a negative number. 

\subsubsection*{Professor Timothy Gowers}
Many of the strengths and weaknesses we observed in GPT-4 are ones that have been commented on several times (in connection with other LLMs as well). For instance, it is not good at calculation, it has a tendency to ignore facts that do not support its main conclusion (even if it itself has generated those facts), and to invent facts that do support it~\citep{azamfirei2023large}. 

When it comes to building examples, it has another weakness, which is that instead of using a process of reasoning to constrain what the example can look like and only then exhibiting the example, it prefers to start by exhibiting the example and then provide the justification that it has the desired properties. If its initial suggestion is correct, then this may be all right (though its justifications are not always correct), but often the example it gives is not correct, and it typically follows it up with a ``justification'' that to a trained mathematician has very obvious flaws. This behaviour supports the stochastic parrot view, since mathematical constructions are typically presented in the literature in the same unhelpful style -- first the example, and then the verification that it works.

One can try to use prompt engineering to discourage GPT-4 from organising its answers in this way, for example, asking the model not to provide an example immediately but to describe a general strategy first, and only then the implementation of the strategy, or to pretend to be a student and ask it to offer hints. While these did improve its output somewhat, they did not make a significant enough difference to affect our general conclusions, so we mainly used more straightforward prompts. 

A further feature that has a negative effect on the experience of interacting with GPT-4 is that even when it gives correct answers, it often describes them and checks their properties in a laborious case-by-case way, and often those checks are not even necessary. For instance, if asked to construct a 3-regular graph with eight vertices, it will typically list all the vertices and edges, and then for each vertex it will say what its neighbours are and comment that there are three of them. Or if it needs the fact that the matrix $\begin{pmatrix}0&1\\0&0\\ \end{pmatrix}$ squares to the zero matrix, instead of just asserting that it does, it will write
\[\begin{pmatrix}0&1\\0&0\\ \end{pmatrix}^2=\begin{pmatrix}0&1\\0&0\\ \end{pmatrix}\begin{pmatrix}0&1\\0&0\\ \end{pmatrix}=\begin{pmatrix}0&0\\0&0\\ \end{pmatrix},\]
which is adding virtually nothing to a bald assertion, since no details are given of the computation of the matrix product (not that one would want them). Similarly, it sometimes ``verifies'' that a matrix is symmetric by writing out that matrix and its transpose and noting that the two matrices it has written out are identical.

In the more positive direction, there were several questions that GPT-4 answered well in a way that is hard to dismiss as mere parroting. To give a simple example, if it is asked to differentiate a function, it will typically do so very competently, suggesting at least some ability to generalise. (A possible explanation for why it is good at differentiation and bad at arithmetic is that there are many more worked examples of differentiation, complete with step-by-step explanations, than there are worked examples of arithmetic.) More examples are discussed in \textit{SI Interactive Case Studies with Experts}, with some speculations about why GPT-4 is good at them.

\subsection{Key Findings} 

We now synthesise the key findings across our two evaluations: insights drawn from our \texttt{MathConverse} dataset collected by applying \texttt{CheckMate} in the domain of mathematics, and expert case studies, where mathematicians deeply engage with GPT-4 to solve problems from particular angles. 


\subsubsection*{[Key finding 1] Correctness and helpfulness of model responses are related, but can diverge in interesting ways}   
\label{sec:corr_and_help_rel}

When a human poses a query, what is the relationship between the perceived helpfulness and correctness of a model's answer? We find that, across all human-model interactions, helpfulness and correctness ratings are highly correlated (with Pearson correlation coefficient $r = 0.83$). This finding corroborates a similar observation in \citep{naturalprover}, wherein although correctness lags behind perceived usefulness, for both per-step and fully-generated proofs, the two qualities are intimately related. This trend underscores an important point: for mathematical language models to be useful assistants, a core quality is that they should consistently produce largely mathematically correct responses.
We also observe an interesting phenomenon at the extremes~(see Figure~\ref{fig:corr_v_help}): there are cases where generations are considered incorrect but helpful, or correct but unhelpful~(see Examples 4 and 3 respectively in \textit{SI Investigating the Boundary between Easy and Hard Problems}).
For instance, models can generate overly verbose answers that are deemed entirely mathematically correct and only moderately helpful.
In expert case study Problem Perturbation to Probe Memorisation in \textit{SI Interactive Case Studies with Experts}, we see models can be good at providing definitions and interestingly can produce helpful scaffolding for a problem (e.g., the right strategy or first few steps), even if details of the full proof are incorrect.
These instances reveal that to create useful assistants, increasing the mathematical correctness of these models alone is not sufficient.




\subsubsection*{[Key finding 2] Lack of verification can induce errors}
\label{misled-main}

The \cm\space ratings are determined from a first-person perspective: participants rate the generations they receive. But what if a participant cannot verify mathematical correctness? Before interacting with models, participants are asked to indicate their confidence at solving the problem on their own. We find instances where participants who indicated low confidence (i.e., confidence $<= 3$; see rating scale in \textit{SI Additional Details on Survey Set-Up}) in being able to solve the problem on their own ended up rating the generation as completely correct \textit{even when it was not}. 
For such examples, as well as participant testimonials of this behaviour, see \textit{SI Additional Survey Observations}. 
In case studies, even if the model does produce a correct solution, this solution is not always arrived at in a seemingly ``human-like'' way; for instance, the model may follow a guess-and-check approach rather than forward planning (e.g., Examples 1, 4, and 7 in \textit{SI Interactive Case Studies with Experts}). However, guess-and-check cannot work well if one cannot ``check'' solutions; indeed, we see that in general, challenges with algebraic manipulation plague in- and out-of-distribution performance~(e.g., examples in \textit{SI Interactive Case Studies with Expert, Number Theory Evaluation}). Further, our studies highlight issues in interaction even when the \textit{human user} attempts to correct the model. One case study illuminated intriguing behaviour when the model was queried about uncertainty: the model began to apologise despite having been correct (see \textit{SI Interactive Case Studies with Expert, Problem Perturbation to Probe Memorisation}).




\subsubsection*{[Key finding 3] The double-edged sword of reliance on memorised solutions}
Memorisation of important concepts and definitions is an important aspect of acquiring world knowledge. The \mc\space taxonomy revealed that queries about mathematical definitions are frequently made by humans, and responses received are considered one of the most helpful among all the categories.
In contrast to definitions, solutions to specific problems should ideally be understood in a fashion that can generalise and not be completely memorised.
By probing GPT-4 capabilities on slightly novel problems or those which involve building examples, we notice the model's tendency to over-rely on plausibly memorised examples or patterns. We caution that we cannot be definitely sure whether these examples are indeed ``memorised`` without direct access to the models' training data. However, from the behaviour, we have a strong suspicion this is the case.


\section{Discussion}

We compile key observations from both our structured and instance-based evaluations into a series of actionable takeaways, which -- given the interdisciplinary nature of our study -- we hope will appeal to a wide audience.
We tailor these takeaways to audiences from different fields. 
To offer balance, we first note that the best LLMs we investigate do demonstrate \textit{some non-trivial ability} in collaborating helpfully and correctly with users on undergraduate-level mathematical problems~(see Figure~\ref{fig:decomp-ratings-by-model}).
Should the user be able to assess the validity of LLM-generated responses, they can meaningfully assist on some problems.
Even if the answers are memorised and can be found somewhere on the internet, LLMs have the advantage of being flexible in their inputs and outputs over traditional search engines. We then close with limitations of our methodology.

\subsection{Takeaways for ML Developers}

\paragraph{Enable Models to Communicate \textit{Calibrated} Uncertainty and Uptake Corrections} We observe cases where people attempted to correct the model when it made an error, the model apologised, and proceeded to give an answer without the necessary corrections or asking for clarification
The pattern often repeated itself until the user seemed to get bored and abort. To improve user experience, systems that can adequately respond to user corrections, for example, through uncertainty calibration~\citep{hullman2018pursuit,vasconcelos2023generation,liu2016communicating}, are compelling~\citep{akyurek2023rl4f,kocielnik2019will,wilder2021learning,mitchell2022memory,meng2022locating}. Indeed, in the models we explored, it was not clear when the model was unsure. We include a discussion with participants about these challenges in a post-survey questionnaire (see \textit{SI Post-Survey Testimonials from Participants}). Communicating uncertainty is critical to ensure users know when they can trust the model output~\citep{bhatt2021uncertainty,hullman2018pursuit} and help calibrate appropriate levels of trust~\citep{umangTrust,spiegelhalter_risk_2017}. However, obtaining accurate, calibrated uncertainty estimates from LLMs can be a difficult endeavour~\citep{xiao-etal-2022-uncertainty,si2022prompting}.

\paragraph{Enable Provision of Rationales} Several participants in \mc{} asked ``why'' a model undertook a particular proof step. Expanding on the justification for a choice could be a valuable educational tool. Generating compelling explanations, on-the-fly and on-request -- provided those explanations are indeed representative and not misleading ~\citep{bhatt2020explainable, zelikman2022star, sevastjanova2022beware, kıcıman2023causal, wu2023interpretability} -- seem promising and desirable to explore to further boost the utility of these systems in partnership with mathematicians.

\paragraph{Strive for Conciseness} Both our survey and our expert case studies find that -- while mathematical correctness appears to often be a foundation for useful assistance in higher-level mathematics -- it is not always sufficient. Responses that were overly verbose were sometimes deemed less helpful. Designing systems that generate concise responses to mathematical queries seems a promising future direction, best also coupled with the capability of showing its ``work'' if needed (related to rationales, see above). The applicability of this to other domains than mathematics remains to be investigated: it may be that responses of different degrees of verbosity are preferred in different domains (e.g., in medicine, longer responses laden with empathy may be preferable~\citep{chatbotMed}).

\subsection{Takeaways for Mathematicians (Students, Educators, and Researchers)}
\paragraph{Pay attention!} Large language models are capable of generating remarkably compelling natural language -- an incredible technical feat which ought not to be dismissed and can be helpful as we see in both our studies. However, such prowess belies the potential for coaxing the reader into not recognising errors. Be careful not to fall into the trap of lazy checking (see \textit{SI Post-Survey Testimonials from Participants} in participant testimonials). This is worth keeping in mind for users learning from or evaluating the generations of LLMs, for example, students and assignment markers. It is worth being cognisant of the risk of automation bias, that is, where a user may inappropriately over-rely on the output of a model simply because it came from a model~\citep{Cummings2004AutomationBI}.

\paragraph{Take a Nuanced View of When These Models Can Help} Reinforcing similar findings from~\citep{frieder2023mathematical}, we observe in this work that LLMs can be useful for retrieving definitions (see \textit{SI Additional Taxonomy Observations}) and can occasionally provide a valuable scaffolding for how to approach a problem (see \textit{SI Additional Survey Observations} and \textit{SI Interactive Case Studies with Experts}). It is important not to presume that a model which performs well in one realm of the task space will surely perform well in another~\citep{bhatt2023learning, kelly2023capturing}. Counterintuitively -- a la Moravec's Paradox~\citep{moravec1988mind} -- it is possible that models will succeed at tasks perceived challenging by humans, but fail at tasks humans consider easy (e.g., derivation versus algebraic manipulation). Mathematicians can take advantage of our query profiles to learn how experienced peers adapt their behaviours to get more out of language models for mathematics.

\paragraph{Be Cautious When Using Current LLMs (Alone) for Heavy Algebra} In particular, our studies further underscore the challenges of present models at algebraic manipulation, corroborating prior work~\citep{bubeck2023sparks, frieder2023mathematical, dziri2023faith}. We believe it is therefore important that mathematicians take care if using these systems for tasks which involve substantial algebra. 
We do not explore plug-ins~\citep{gptPlugins} in this paper, nor alternative hybrid neuro-symbolic approaches (e.g.,~\citep{dsp, timManifesto, peano, lemmaReasoning, kazemi2022lambada, wong2023word}), which may prove a useful salve for some of this failure mode. 

\subsection{Takeaways for LLM Development, Evaluation, and Deployment}
 
We conclude with broad takeaways for anyone developing, evaluating or considering deploying LLMs in practice.

\paragraph{Carefully Discern When Assistance is Needed (or Even Worth Utilising)} To build complementary systems~\citep{wilder2020learning}, understanding when an AI-based assistant is helpful is of utmost importance: seldom will such an assistant be helpful in all settings~\citep{bhatt2023learning}. An important question will be in which settings such an assistant can be useful without undermining the agency of the mathematician, for example, of the kind already being proposed when considering using LLMs in coursework~\citep{baLLMwork}. Future work would benefit from considering how to build usable assistants that optimise for complementarity, providing support as and when needed~\citep{miller2023explainable}.

\paragraph{Collaboration between ML Practitioners and Domain Experts is Valuable} Conducting investigations in partnership with domain experts can be especially fruitful for characterising model behaviour~\citep{mcgrath2022acquisition, davies2021advancing, dramatron}, particularly by designing entirely new tasks, as our expert case studies demonstrate. We encourage forming such interdisciplinary partnerships in and beyond mathematics.

\paragraph{Incorporate Interactivity into LLM Capability Assessments} To truly comprehend the landscape of an LLM's capabilities, we believe it is paramount to incorporate interactive evaluations. Our work further drives home the importance of interactive evaluation as a way to gain deeper insights into the strengths and weaknesses of these models and probe characteristics which may be preferable for assistive settings. However, as we highlight here, the interactive study of LLMs not only serves to characterise model behaviour, but it identifies ways in which \textit{humans} may themselves choose to interact with these models and actually use these systems~\citep{ringer2020replica}. A wave of works increasingly illuminates the sensitivity of these models to the choice of prompts ~\citep{weichain, zhou2022least, yao2023tree}. As such, it is important to consider the form and content of queries that humans may use to interact with these systems both to design systems more adapted to particular user queries, and to inform users of best practices. It may be valuable for system maintainers to recognise whether or not users are leveraging these tactics, in order to help better inform the techniques for boosting the quality of the response for their query.

We hope to see more works like ours and \citep{lee2022evaluating, lee2022coauthor, dramatron} that study LLMs in the context of human-computer interactions. \cm\space offers a place to start, potentially complemented by free-form evaluation of the kind we conduct in our expert case studies.

\section{Limitations}

While our study reveals insights into how mathematicians may use language models -- and opens doors for future interactive evaluation -- our survey is simply an \textit{initial step} in evaluating LLMs for mathematical assistance. Our sample size is small but informative; we consider \texttt{MathConverse} to be a preliminary dataset to spark further methodological and deployment-time considerations. Additionally, we ask each participant to rate generations provided during their own interaction trace; while this permits first-person evaluation of the kind called for in \citep{leeInteractive}, for those who do not already know how to solve the problem this means that they may be falsely judging the correctness of the generation. A sensible next step would be two-fold: deploying our evaluation platform with students who have not already solved such problems, and sending the interaction traces off for external evaluation as well. We also encourage a reassessment of mathematician interactions over time; it is quite possible -- in fact likely -- that the kinds of interactions humans make with these systems will evolve as their capabilities grow. Additionally, our taxonomy categories are non-exhaustive; alternative categorisations are possible. Nevertheless, we found our categorisation sufficient to draw helpful findings. 

And while our case studies offer invaluable insight into the prompting behaviour of expert mathematicians and further characterisation of model performance, each individual may bring to bear their own expectations about models' strengths and weaknesses, which could seep into the way probing behaviour is chosen and model outputs are interpreted. We emphasise that, as with the survey results, these insights ought not to be taken as a firm testament about the capabilities nor potential of these models, much less all language models (we only consider a handful, and all from the OpenAI family). Rather, we hope the evaluation toolkit expanded on in our studies pave the way for further research into the use of LLMs as assistants for problem-solving, in mathematics and beyond.

\section{Conclusion}

As LLMs are increasingly deployed in human-facing settings where they may serve as assistants, it is paramount that evaluation of their efficacy fundamentally includes evaluation in an interactive context~\citep{leeInteractive}. As we demonstrate, these interactive evaluations can be structured (e.g., leveraging \texttt{CheckMate}) or free-form (e.g., through sourced domain expert or target user interactions). LLMs, and foundation models broadly, are complex and often surprising in their behaviour; so are humans. Hence characterising potential failure modes 
in LLM and human interactions necessitates a multi-factorial evaluation approach, which includes both interactive evaluation and classical, static-snapshot evaluation \citep{burnell2023}. Through our study, we extract insights which we hope can inform careful design and deployment when considering leveraging LLM-based mathematics assistants and reasoning engines. We believe that our study paves the way for further evaluation of the use of foundation models in mathematics and other domains, particularly through closer collaboration with domain experts.

\section{Methods}
\label{sec:Methods}

\subsection{\texttt{CheckMate}: Adaptable Platform for Interactive Evaluation}
\label{sec:checkmate-intro}

We introduce \texttt{CheckMate} as an adaptable platform to support \textit{interactive} evaluation of language models.\footnote{Base code can be found in our repository: \href{https://github.com/collinskatie/checkmate}{https://github.com/collinskatie/checkmate}.} Humans can interact with and rate text generated by language models, and \texttt{CheckMate} records the ``interaction traces''.\footnote{We borrow the terminology of ~\citep{lee2022coauthor, lee2022evaluating} for ``interaction traces''.}

We design \texttt{CheckMate} to support two flavours of evaluation: studying the interactions with a single model, and studying preference across a bank of models. First, we introduce the rating scheme for a single model. Then, we discuss how we support comparative evaluation over a suite of models. We focus on the domain of mathematical theorem proving; however, \texttt{CheckMate} can be extended more broadly (see \textit{SI User Guide for \texttt{CheckMate}}). 

\subsubsection*{Evaluation for a Single Model}  
\label{checkmate-for-a-single-model}

Evaluation begins with permitting the participant to freely interact with the model in order to solve a problem. We encourage participants to imagine they are trying to solve the problem -- and elicit assistance. The participant can continue to explore assistance for up to 20 interaction exchanges.\footnote{We chose a limit of 20 expecting that participants may fatigue beyond that point and to guard against the possibility that a participant could try to interact unfettered with the model for an extended period of time.} When the participant is satisfied with the level of assistance (or sufficiently unsatisfied that they wish to terminate the interaction), they proceed to evaluate \textit{each step} of their entire interaction. 

We design \cm\space to support a multi-dimensional evaluation over the interaction trace for the successive human query-model generation pairs. At present, the platform is designed with a mix of Likert scales and radio buttons (see \textit{SI Additional Details on Survey Set-Up and User Guide for \texttt{CheckMate}}). However, \cm\space can be readily extended with alternative rating types, for instance, to handle individual error profiling \citep{naturalprover} or additional interaction metrics as proposed in \citep{lee2022evaluating, shen2023parachute}, if desired. 

\subsubsection*{Comparative Evaluation Across Models}

With an ever-growing suite of language models available for humans to leverage, it is important to compare capabilities -- and how these compare to previous versions. When done, such comparisons typically involve single snapshots. \texttt{CheckMate} permits the study of preference \textit{over the interaction trace} and can serve as a valuable tool to explore the evolution of assistance potential. 

In \texttt{CheckMate}, participants provide a rank order over which model they preferred interacting with, after they have interacted with two or more models. This instantiation of the platform is set-up such that participants interact with a different task per model (to avoid ``bleed over'' effects when considering the same problem multiple times). However, alternative designs, for example, rating models per task, or subsampling the models to evaluate, are possible adaptations to our paradigm (see \textit{SI User Guide for \texttt{CheckMate}}).
Importantly, participants are \textit{blind} to which model they are evaluating at any time; this ensures they are not biased by preconceived notions of which model may be more performative. 

In the rank order, participants can assign the same rank if they are unsure which model they prefer. Future work could consider more expansive comparative preference evaluation. We provide further details on \texttt{CheckMate} and hosting our survey in the \textit{SI User Guide for \texttt{CheckMate}}.

\subsubsection*{Instantiating \texttt{CheckMate} for Mathematics to Collect \texttt{MathConverse}}



Recall, survey participants are asked to prove a mathematical statement and to use an AI system to assist them in any way to carry out this task. As the interaction is free-form, interactions can range from asking for help on the entire problem, to clarifying definitions, or asking for an explanation for a particular generated proof step. Participants are not provided with possible interaction behaviours in advance to avoid priming. When the participant is satisfied with the level of assistance (or sufficiently unsatisfied that they wish to terminate the interaction), they proceed to evaluate \textit{each step} of their entire interaction. Participants solve a different problem for three models (Instruct-GPT, ChatGPT and GPT-4), where the order of the models is shuffled and participants are blind to which model they are interacting with.

We next describe our task set-up over which we conduct evaluations. The study was conducted under the approval of the University of Cambridge Computer Science Ethics Division. Example interface screens of \texttt{CheckMate} for mathematics are included in the the Supplement. 

\subsubsection*{Tasks} We select $54$ problems from ProofWiki, a corpus of \textit{undergraduate-level} mathematics problems.\footnote{\url{https://proofwiki.org}} Nine problems are selected from each of six mathematics topics (linear algebra, number theory, probability theory, algebra, topology, and group theory). We select these topics to span a range of subject areas in typical undergraduate mathematical curricula. The problems can be found at \url{https://github.com/collinskatie/checkmate/tree/main/data/problems}.

\subsubsection*{Rating Scales}

Participants evaluate the \textit{perceived helpfulness} and \textit{mathematical correctness} of each step, selecting one ``preference'' and one ``quality'' metric, as defined in \citep{lee2022evaluating}. Cognitive load and biases are kept in mind at each stage of the design, for example, lightening the number of ratings per page, and randomising model rating order to reduce possible ordering effects. Ratings are provided on a 7-point Likert scale, with the width chosen to ameliorate potential rating collapse (i.e., the phenomenon where participants hesitate to use scale endpoints~\citep{bishop2015use}). Further, we select only two factors per step to avoid excess cognitive load while rating. Before responding, participants specify their confidence in being able to solve the problem on their own. After interacting with the three models on three different problems, participants are shown the full interaction traces with each model and (blindly) indicate their rating about which model they would prefer as an assistant (blindly) via a dropdown bar. We include full details of the scales in \textit{SI Additional Details on Survey Set-Up}. For all quantitative analyses -- with the exception of analysing users' stopping behaviour in Figure~\ref{fig:why-stop} -- we filter out generations rated as zero for mathematical correctness, as that means that no mathematically-relevant content was included; we find that these are typically responses to greetings or exclamations (e.g., after the user has thanked the model -- see released data). 

\subsubsection*{Language Model Selection and Set-Up} Participants evaluate three popular language models: InstructGPT~\citep{ouyang2022training}, ChatGPT~\citep{chatGPT}, and GPT-4~\citep{openai2023gpt4} in chat mode\footnote{Our study design began in February 2023, shorty after GPT-4 was released.}.
As new language models are introduced, the methodology of designing optimal prompts is rapidly evolving (e.g., to name a few
~\citep{weichain, zhou2022least, yao2023tree}). Since we are studying how \textit{real domain users} (i.e., mathematicians) would interact with these systems \textit{in-the-wild}, we keep a sparse base prompt, only asking the model to be a helpful mathematical assistant in the prompt. Further details for the experimental setup can be found in \textit{SI Additional Details on Survey Set-Up}.

\subsubsection*{Participants} We recruit mathematician volunteers to participate in our evaluation. In total, we received 25 entries comprising 261 human-model interactions; while this could comprise 25 unique participants, we did not store a unique participant identifier, for example, the IP address, for privacy reasons (see \textit{SI Additional Details on Survey Set-Up}), thus we cannot confirm that these are exactly 25 unique individuals. The mathematicians have experience levels ranging from current undergraduate students up to expert mathematics professors; for participants without a formal mathematics degree, they likely have some exposure to high-level mathematics (see \textit{SI Participant Recruitment and Additional Details}). Each participant chooses one of the six topics and can evaluate as many questions as they like (up to the maximum of 9). Note, the range of expertise of our participants -- up to world-class experts -- coupled with the fact that our problems sit at a level where students majoring in mathematics might find them in textbooks or as exercises, means that some participants may be able to solve the problems already, others may not. If a participant knows how to solve the problem, we ask that they \textit{imagine} what kind of assistance they would like had they been at the experience level of someone who does not know how to solve the problem. More details on recruitment, data processing, and expertise can be found in \textit{SI Additional Details on Survey Set-Up}.

\subsection{Deriving a Taxonomy from \texttt{MathConverse}}
We observe a wide spectrum of mathematicians' interactions with AI assistants in \texttt{MathConverse}. We derive an initial taxonomy of these interactions and annotate each interaction according to the taxonomy. To build the taxonomy, a subset of our author team manually inspected each interaction (i.e., the user's query) and identified 10 broad categories in which most interactions seemed to fall. These categories are specified in the Taxonomising User Behaviour from MathConverse' subsection of the Results. We included an additional ``Other'' bucket for queries which did not fall into one of the 10; cases that fell into such a category were discussed amongst the annotators. Four authors then manually annotated each user query into these buckets. Annotators were asked to mark whether an interaction fell into a bucket, with an option to specify whether they were unsure. Each interaction was annotated by a single annotator; however, in the case of confusion or discrepancies, annotators discussed and came to an agreed coding. 
We release the annotated taxonomy in our \href{https://github.com/collinskatie/checkmate}{repository}. Full instructions given to annotators are included in the \textit{SI Additional Details on Taxonomy Construction}. 




\subsection{Interactive Case Studies with Experts}
\label{case-studies}

While structured interactive evaluation permits nice quantitative findings, to deeply understand the capability of LLMs -- in the context of mathematics and beyond -- free-form interaction, like instance-level evaluation~\citep{burnell2023} can be particularly revealing. Here we want to 
expands the scope of the evaluation, attempting to locate the boundary between problems that GPT-4 finds easy and those it finds hard. In our quantitative study with \texttt{CheckMate} we observed a close relationship between mathematical correctness and perceived usefulness. But since correlation is not causation, we further explore the broader \textit{mathematical reasoning capabilities} of these models as a bedrock to inform their utility as proof assistants.

We reiterate that we are not aiming to single out GPT-4 for criticism. Rather, our goal is 1) to offer one of the first real expert mathematician interactive case studies with LLMs to help guide the design of better mathematical assistants and inform their safe, trustworthy use by helping characterise their limitations, 2) to pave the way for further interactive evaluations, and 3) to highlight patterns of human-computer interaction not previously known to the community, particularly when the humans interacting are domain-leading experts. We hope the work will be of interest to ML engineers and researchers, cognitive scientists, human-computer interaction specialists, mathematicians, educators, and beyond. 

A complete transcript of interactions for each case study example is included in the Supplementary Materials. We maintain the original text of each case study author for authenticity, with only minor edits for precision and coherence. For the interactions with ProofWiki problems, we also host them with a \href{https://albertqjiang.github.io/case_studies/pages/proofwiki_21.html}{sample explorer} for clearer visualisation.


First, our recruited experts conduct a deeper dive into some of the problems we explored in our previous evaluation. Specifically, we use the problems as a playground to explore how much the model seems to ``know'' about relevant concepts and further characterise what interactions can yield better (or worse) performance and assistance experience. We focus on GPT-4~(in chat mode) because it showed the strongest overall performance in our quantitative study with \texttt{CheckMate}. The first case study is provided by Dr.\ William Hart, a number theorist by training; the second is primarily contributed by Dr.\ Wenda Li, a formal mathematics expert, and the third one was conducted with Prof.\ Timothy Gowers, a Fields medalist and Professor in mathematics.




\section*{Acknowledgements}
Special thanks to Fabian Gloeckle, for support and informative discussions throughout the project. We additionally thank in no particular order Anshula Gandhi, Jonas Bayer, Angeliki Koutsoukou-Argyraki, Fredy Yip, Mariusz Mirek, Gabriel Poesia, Noah Goodman, Valerie Chen, Nina Grgić-Hlača, Ilia Sucholutsky, Shoaib Ahmed Siddiqui, and Ced Zhang for valuable discussions around assistive systems in AI and mathematics, and LLM evaluation. KMC gratefully acknowledges support from the Marshall Commission and the Cambridge Trust.
AQJ acknowledges support from the Peterhouse Graduate Studentship with gratitude. 
SF and TL acknowledge  support  from the Alan Turing Institute under the EPSRC grant EP/N510129/1.  
MZ acknowledges support from the Leverhulme Trust grant ECF-2021-429. 
UB acknowledges support  from  DeepMind  and  the  Leverhulme Trust  via  the  Leverhulme  Centre  for  the  Future  of  Intelligence  (CFI),  and  from ELSA: European Lighthouse on Secure and Safe AI via the Alan Turing Institute. 
TL acknowledges  support  from the AXA Research Fund and  the EU TAILOR grant 952215. 
WL is supported by the ERC Advanced Grant ALEXANDRIA (Project GA 742178). 
AW  acknowledges  support  from  a  Turing  AI  Fellowship  under grant  EP/V025279/1,  The  Alan  Turing  Institute,  and  the Leverhulme Trust via CFI. 
MJ acknowledges support from the EPSRC under grant EP/T019603/1.


\bibliography{pnas-sample}

\begin{thebibliography}{112}
\providecommand{\natexlab}[1]{#1}
\providecommand{\url}[1]{\texttt{#1}}
\expandafter\ifx\csname urlstyle\endcsname\relax
  \providecommand{\doi}[1]{doi: #1}\else
  \providecommand{\doi}{doi: \begingroup \urlstyle{rm}\Url}\fi

\bibitem[Abid et~al.(2019)Abid, Abdalla, Abid, Khan, Alfozan, and Zou]{abid2019gradio}
A.~Abid, A.~Abdalla, A.~Abid, D.~Khan, A.~Alfozan, and J.~Zou.
\newblock {Gradio: Hassle-Free Sharing and Testing of ML Models in the Wild}.
\newblock \emph{arXiv preprint arXiv:1906.02569}, 2019.

\bibitem[Aitken et~al.(1998)Aitken, Gray, Melham, and Thomas]{aitken1998interactive}
J.~S. Aitken, P.~Gray, T.~Melham, and M.~Thomas.
\newblock Interactive theorem proving: An empirical study of user activity.
\newblock \emph{Journal of Symbolic Computation}, 25\penalty0 (2):\penalty0 263--284, 1998.

\bibitem[Aky{\"u}rek et~al.(2023)Aky{\"u}rek, Aky{\"u}rek, Madaan, Kalyan, Clark, Wijaya, and Tandon]{akyurek2023rl4f}
A.~F. Aky{\"u}rek, E.~Aky{\"u}rek, A.~Madaan, A.~Kalyan, P.~Clark, D.~Wijaya, and N.~Tandon.
\newblock {RL4F: Generating Natural Language Feedback with Reinforcement Learning for Repairing Model Outputs}.
\newblock \emph{arXiv preprint arXiv:2305.08844}, 2023.

\bibitem[Amini et~al.(2019)Amini, Gabriel, Lin, Koncel{-}Kedziorski, Choi, and Hajishirzi]{mathqa}
A.~Amini, S.~Gabriel, S.~Lin, R.~Koncel{-}Kedziorski, Y.~Choi, and H.~Hajishirzi.
\newblock Mathqa: Towards interpretable math word problem solving with operation-based formalisms.
\newblock In J.~Burstein, C.~Doran, and T.~Solorio, editors, \emph{Proceedings of the 2019 Conference of the North American Chapter of the Association for Computational Linguistics: Human Language Technologies, {NAACL-HLT} 2019, Minneapolis, MN, USA, June 2-7, 2019, Volume 1 (Long and Short Papers)}, pages 2357--2367. Association for Computational Linguistics, 2019.
\newblock \doi{10.18653/v1/n19-1245}.
\newblock URL \url{https://doi.org/10.18653/v1/n19-1245}.

\bibitem[Anil et~al.(2023)Anil, Dai, Firat, Johnson, Lepikhin, Passos, Shakeri, Taropa, Bailey, Chen, Chu, Clark, Shafey, Huang, Meier-Hellstern, Mishra, Moreira, Omernick, Robinson, Ruder, Tay, Xiao, Xu, Zhang, Abrego, Ahn, Austin, Barham, Botha, Bradbury, Brahma, Brooks, Catasta, Cheng, Cherry, Choquette-Choo, Chowdhery, Crepy, Dave, Dehghani, Dev, Devlin, Díaz, Du, Dyer, Feinberg, Feng, Fienber, Freitag, Garcia, Gehrmann, Gonzalez, Gur-Ari, Hand, Hashemi, Hou, Howland, Hu, Hui, Hurwitz, Isard, Ittycheriah, Jagielski, Jia, Kenealy, Krikun, Kudugunta, Lan, Lee, Lee, Li, Li, Li, Li, Li, Lim, Lin, Liu, Liu, Maggioni, Mahendru, Maynez, Misra, Moussalem, Nado, Nham, Ni, Nystrom, Parrish, Pellat, Polacek, Polozov, Pope, Qiao, Reif, Richter, Riley, Ros, Roy, Saeta, Samuel, Shelby, Slone, Smilkov, So, Sohn, Tokumine, Valter, Vasudevan, Vodrahalli, Wang, Wang, Wang, Wang, Wieting, Wu, Xu, Xu, Xue, Yin, Yu, Zhang, Zheng, Zheng, Zhou, Zhou, Petrov, and Wu]{anil2023palm}
R.~Anil, A.~M. Dai, O.~Firat, M.~Johnson, D.~Lepikhin, A.~Passos, S.~Shakeri, E.~Taropa, P.~Bailey, Z.~Chen, E.~Chu, J.~H. Clark, L.~E. Shafey, Y.~Huang, K.~Meier-Hellstern, G.~Mishra, E.~Moreira, M.~Omernick, K.~Robinson, S.~Ruder, Y.~Tay, K.~Xiao, Y.~Xu, Y.~Zhang, G.~H. Abrego, J.~Ahn, J.~Austin, P.~Barham, J.~Botha, J.~Bradbury, S.~Brahma, K.~Brooks, M.~Catasta, Y.~Cheng, C.~Cherry, C.~A. Choquette-Choo, A.~Chowdhery, C.~Crepy, S.~Dave, M.~Dehghani, S.~Dev, J.~Devlin, M.~Díaz, N.~Du, E.~Dyer, V.~Feinberg, F.~Feng, V.~Fienber, M.~Freitag, X.~Garcia, S.~Gehrmann, L.~Gonzalez, G.~Gur-Ari, S.~Hand, H.~Hashemi, L.~Hou, J.~Howland, A.~Hu, J.~Hui, J.~Hurwitz, M.~Isard, A.~Ittycheriah, M.~Jagielski, W.~Jia, K.~Kenealy, M.~Krikun, S.~Kudugunta, C.~Lan, K.~Lee, B.~Lee, E.~Li, M.~Li, W.~Li, Y.~Li, J.~Li, H.~Lim, H.~Lin, Z.~Liu, F.~Liu, M.~Maggioni, A.~Mahendru, J.~Maynez, V.~Misra, M.~Moussalem, Z.~Nado, J.~Nham, E.~Ni, A.~Nystrom, A.~Parrish, M.~Pellat, M.~Polacek, A.~Polozov, R.~Pope, S.~Qiao, E.~Reif, B.~Richter,
  P.~Riley, A.~C. Ros, A.~Roy, B.~Saeta, R.~Samuel, R.~Shelby, A.~Slone, D.~Smilkov, D.~R. So, D.~Sohn, S.~Tokumine, D.~Valter, V.~Vasudevan, K.~Vodrahalli, X.~Wang, P.~Wang, Z.~Wang, T.~Wang, J.~Wieting, Y.~Wu, K.~Xu, Y.~Xu, L.~Xue, P.~Yin, J.~Yu, Q.~Zhang, S.~Zheng, C.~Zheng, W.~Zhou, D.~Zhou, S.~Petrov, and Y.~Wu.
\newblock {PaLM 2 Technical Report}, 2023.

\bibitem[Ayers et~al.(2023)Ayers, Poliak, Dredze, Leas, Zhu, Kelley, Faix, Goodman, Longhurst, Hogarth, and Smith]{chatbotMed}
J.~W. Ayers, A.~Poliak, M.~Dredze, E.~C. Leas, Z.~Zhu, J.~B. Kelley, D.~J. Faix, A.~M. Goodman, C.~A. Longhurst, M.~Hogarth, and D.~M. Smith.
\newblock {Comparing Physician and Artificial Intelligence Chatbot Responses to Patient Questions Posted to a Public Social Media Forum}.
\newblock \emph{JAMA Internal Medicine}, 04 2023.
\newblock ISSN 2168-6106.
\newblock \doi{10.1001/jamainternmed.2023.1838}.
\newblock URL \url{https://doi.org/10.1001/jamainternmed.2023.1838}.

\bibitem[Azamfirei et~al.(2023)Azamfirei, Kudchadkar, and Fackler]{azamfirei2023large}
R.~Azamfirei, S.~R. Kudchadkar, and J.~Fackler.
\newblock Large language models and the perils of their hallucinations.
\newblock \emph{Critical Care}, 27\penalty0 (1):\penalty0 1--2, 2023.

\bibitem[Ba and Wang(2023)]{baLLMwork}
J.~Ba and B.~Wang.
\newblock Csc413/2516 winter 2023 university of toronto, assignment 1, 2023.
\newblock URL \url{https://uoft-csc413.github.io/2023/assets/assignments/a1.pdf}.

\bibitem[Beckert et~al.(2015)Beckert, Grebing, and B{\"o}hl]{beckert2015usability}
B.~Beckert, S.~Grebing, and F.~B{\"o}hl.
\newblock A usability evaluation of interactive theorem provers using focus groups.
\newblock In \emph{Software Engineering and Formal Methods: SEFM 2014 Collocated Workshops: HOFM, SAFOME, OpenCert, MoKMaSD, WS-FMDS, Grenoble, France, September 1-2, 2014, Revised Selected Papers 12}, pages 3--19. Springer, 2015.

\bibitem[Bender et~al.(2021{\natexlab{a}})Bender, Gebru, McMillan-Major, and Shmitchell]{bender2021dangers}
E.~M. Bender, T.~Gebru, A.~McMillan-Major, and S.~Shmitchell.
\newblock On the dangers of stochastic parrots: Can language models be too big?
\newblock In \emph{Proceedings of the 2021 ACM conference on fairness, accountability, and transparency}, pages 610--623, 2021{\natexlab{a}}.

\bibitem[Bender et~al.(2021{\natexlab{b}})Bender, Gebru, McMillan-Major, and Shmitchell]{stochasticParrots}
E.~M. Bender, T.~Gebru, A.~McMillan-Major, and S.~Shmitchell.
\newblock On the dangers of stochastic parrots: Can language models be too big?
\newblock In \emph{Proceedings of the 2021 ACM Conference on Fairness, Accountability, and Transparency}, FAccT '21, page 610–623, New York, NY, USA, 2021{\natexlab{b}}. Association for Computing Machinery.
\newblock ISBN 9781450383097.
\newblock \doi{10.1145/3442188.3445922}.

\bibitem[Bhatt et~al.(2020)Bhatt, Xiang, Sharma, Weller, Taly, Jia, Ghosh, Puri, Moura, and Eckersley]{bhatt2020explainable}
U.~Bhatt, A.~Xiang, S.~Sharma, A.~Weller, A.~Taly, Y.~Jia, J.~Ghosh, R.~Puri, J.~M. Moura, and P.~Eckersley.
\newblock Explainable machine learning in deployment.
\newblock In \emph{Proceedings of the 2020 conference on fairness, accountability, and transparency}, pages 648--657, 2020.

\bibitem[Bhatt et~al.(2021)Bhatt, Antor{\'a}n, Zhang, Liao, Sattigeri, Fogliato, Melan{\c{c}}on, Krishnan, Stanley, Tickoo, et~al.]{bhatt2021uncertainty}
U.~Bhatt, J.~Antor{\'a}n, Y.~Zhang, Q.~V. Liao, P.~Sattigeri, R.~Fogliato, G.~Melan{\c{c}}on, R.~Krishnan, J.~Stanley, O.~Tickoo, et~al.
\newblock Uncertainty as a form of transparency: Measuring, communicating, and using uncertainty.
\newblock In \emph{Proceedings of the 2021 AAAI/ACM Conference on AI, Ethics, and Society}, pages 401--413, 2021.

\bibitem[Bhatt et~al.(2023)Bhatt, Chen, Collins, Kamalaruban, Kallina, Weller, and Talwalkar]{bhatt2023learning}
U.~Bhatt, V.~Chen, K.~M. Collins, P.~Kamalaruban, E.~Kallina, A.~Weller, and A.~Talwalkar.
\newblock Learning personalized decision support policies.
\newblock \emph{arXiv e-prints}, pages arXiv--2304, 2023.

\bibitem[Bishop and Herron(2015)]{bishop2015use}
P.~A. Bishop and R.~L. Herron.
\newblock Use and misuse of the likert item responses and other ordinal measures.
\newblock \emph{International journal of exercise science}, 8\penalty0 (3):\penalty0 297, 2015.

\bibitem[Bledsoe(1977)]{Bledsoe1977NonResolutionTP}
W.~W. Bledsoe.
\newblock Non-resolution theorem proving.
\newblock \emph{Artif. Intell.}, 9:\penalty0 1--35, 1977.

\bibitem[Bommasani et~al.(2021)Bommasani, Hudson, Adeli, Altman, Arora, von Arx, Bernstein, Bohg, Bosselut, Brunskill, et~al.]{bommasani2021opportunities}
R.~Bommasani, D.~A. Hudson, E.~Adeli, R.~Altman, S.~Arora, S.~von Arx, M.~S. Bernstein, J.~Bohg, A.~Bosselut, E.~Brunskill, et~al.
\newblock On the opportunities and risks of foundation models.
\newblock \emph{arXiv preprint arXiv:2108.07258}, 2021.

\bibitem[Brown et~al.(2020)Brown, Mann, Ryder, Subbiah, Kaplan, Dhariwal, Neelakantan, Shyam, Sastry, Askell, et~al.]{brown2020language}
T.~Brown, B.~Mann, N.~Ryder, M.~Subbiah, J.~D. Kaplan, P.~Dhariwal, A.~Neelakantan, P.~Shyam, G.~Sastry, A.~Askell, et~al.
\newblock Language models are few-shot learners.
\newblock \emph{NeurIPS}, 33:\penalty0 1877--1901, 2020.

\bibitem[Bubeck et~al.(2023)Bubeck, Chandrasekaran, Eldan, Gehrke, Horvitz, Kamar, Lee, Lee, Li, Lundberg, Nori, Palangi, Ribeiro, and Zhang]{bubeck2023sparks}
S.~Bubeck, V.~Chandrasekaran, R.~Eldan, J.~Gehrke, E.~Horvitz, E.~Kamar, P.~Lee, Y.~T. Lee, Y.~Li, S.~Lundberg, H.~Nori, H.~Palangi, M.~T. Ribeiro, and Y.~Zhang.
\newblock Sparks of artificial general intelligence: Early experiments with gpt-4, 2023.

\bibitem[Bundy(1983)]{Bundy1983TheCM}
A.~Bundy.
\newblock \emph{The Computer Modelling of Mathematical Reasoning}.
\newblock Academic Press, 1983.

\bibitem[Bundy(1988)]{Bundy1988TheUO}
A.~Bundy.
\newblock The use of explicit plans to guide inductive proofs.
\newblock In \emph{CADE}, 1988.

\bibitem[Bundy et~al.(1993)Bundy, Stevens, Harmelen, Ireland, and Smaill]{Bundy1993RipplingAH}
A.~Bundy, A.~Stevens, F.~V. Harmelen, A.~Ireland, and A.~Smaill.
\newblock Rippling: A heuristic for guiding inductive proofs.
\newblock \emph{Artif. Intell.}, 62:\penalty0 185--253, 1993.

\bibitem[Burnell et~al.(2023)Burnell, Schellaert, Burden, Ullman, Martinez-Plumed, Tenenbaum, Rutar, Cheke, Sohl-Dickstein, Mitchell, Kiela, Shanahan, Voorhees, Cohn, Leibo, and Hernandez-Orallo]{burnell2023}
R.~Burnell, W.~Schellaert, J.~Burden, T.~D. Ullman, F.~Martinez-Plumed, J.~B. Tenenbaum, D.~Rutar, L.~G. Cheke, J.~Sohl-Dickstein, M.~Mitchell, D.~Kiela, M.~Shanahan, E.~M. Voorhees, A.~G. Cohn, J.~Z. Leibo, and J.~Hernandez-Orallo.
\newblock Rethink reporting of evaluation results in ai.
\newblock \emph{Science}, 380\penalty0 (6641):\penalty0 136--138, 2023.
\newblock \doi{10.1126/science.adf6369}.
\newblock URL \url{https://www.science.org/doi/abs/10.1126/science.adf6369}.

\bibitem[Casper et~al.(2023)Casper, Davies, Shi, Krendl~Gilbert, Scheurer, Rando, Freedman, Korbak, Lindner, Freire, et~al.]{casper2023open}
S.~Casper, X.~Davies, C.~Shi, T.~Krendl~Gilbert, J.~Scheurer, J.~Rando, R.~Freedman, T.~Korbak, D.~Lindner, P.~Freire, et~al.
\newblock Open problems and fundamental limitations of reinforcement learning from human feedback.
\newblock \emph{arXiv e-prints}, pages arXiv--2307, 2023.

\bibitem[Christiano et~al.(2017)Christiano, Leike, Brown, Martic, Legg, and Amodei]{christiano2017deep}
P.~F. Christiano, J.~Leike, T.~Brown, M.~Martic, S.~Legg, and D.~Amodei.
\newblock Deep reinforcement learning from human preferences.
\newblock \emph{Advances in neural information processing systems}, 30, 2017.

\bibitem[Clark et~al.(2018)Clark, Ross, Tan, Ji, and Smith]{clark+etal:18}
E.~Clark, A.~S. Ross, C.~Tan, Y.~Ji, and N.~A. Smith.
\newblock Creative writing with a machine in the loop: Case studies on slogans and stories.
\newblock In \emph{Proceedings of IUI}, 2018.

\bibitem[Cobbe et~al.(2021)Cobbe, Kosaraju, Bavarian, Hilton, Nakano, Hesse, and Schulman]{gsm8k}
K.~Cobbe, V.~Kosaraju, M.~Bavarian, J.~Hilton, R.~Nakano, C.~Hesse, and J.~Schulman.
\newblock Training verifiers to solve math word problems.
\newblock \emph{CoRR}, abs/2110.14168, 2021.
\newblock URL \url{https://arxiv.org/abs/2110.14168}.

\bibitem[Cohn and Hernandez-Orallo(2023)]{cohn2023dialectical}
A.~G. Cohn and J.~Hernandez-Orallo.
\newblock Dialectical language model evaluation: An initial appraisal of the commonsense spatial reasoning abilities of llms.
\newblock \emph{arXiv preprint arXiv:2304.11164}, 2023.

\bibitem[Cummings(2004)]{Cummings2004AutomationBI}
M.~L. Cummings.
\newblock Automation bias in intelligent time critical decision support systems.
\newblock In \emph{AIAA 1st Intelligent Systems Technical Conference}, 2004.
\newblock URL \url{https://api.semanticscholar.org/CorpusID:10328335}.

\bibitem[Davies et~al.(2021)Davies, Veli{\v{c}}kovi{\'c}, Buesing, Blackwell, Zheng, Toma{\v{s}}ev, Tanburn, Battaglia, Blundell, Juh{\'a}sz, et~al.]{davies2021advancing}
A.~Davies, P.~Veli{\v{c}}kovi{\'c}, L.~Buesing, S.~Blackwell, D.~Zheng, N.~Toma{\v{s}}ev, R.~Tanburn, P.~Battaglia, C.~Blundell, A.~Juh{\'a}sz, et~al.
\newblock Advancing mathematics by guiding human intuition with ai.
\newblock \emph{Nature}, 600\penalty0 (7887):\penalty0 70--74, 2021.

\bibitem[Davis et~al.(1962)Davis, Logemann, and Loveland]{davis1962machine}
M.~Davis, G.~Logemann, and D.~Loveland.
\newblock A machine program for theorem-proving.
\newblock \emph{Communications of the ACM}, 5\penalty0 (7):\penalty0 394--397, 1962.

\bibitem[de~Moura and Bj{\o}rner(2008)]{Moura2008Z3AE}
L.~M. de~Moura and N.~S. Bj{\o}rner.
\newblock Z3: An efficient smt solver.
\newblock In \emph{International Conference on Tools and Algorithms for Construction and Analysis of Systems}, 2008.

\bibitem[Delahaye(2000)]{delahaye2000tactic}
D.~Delahaye.
\newblock A tactic language for the system coq.
\newblock In M.~Parigot and A.~Voronkov, editors, \emph{Logic for Programming and Automated Reasoning, 7th International Conference, {LPAR} 2000, Reunion Island, France, November 11-12, 2000, Proceedings}, volume 1955 of \emph{Lecture Notes in Computer Science}, pages 85--95. Springer, 2000.
\newblock \doi{10.1007/3-540-44404-1\_7}.
\newblock URL \url{https://doi.org/10.1007/3-540-44404-1\_7}.

\bibitem[Dohan et~al.(2022)Dohan, Xu, Lewkowycz, Austin, Bieber, Lopes, Wu, Michalewski, Saurous, Sohl-Dickstein, et~al.]{dohan2022language}
D.~Dohan, W.~Xu, A.~Lewkowycz, J.~Austin, D.~Bieber, R.~G. Lopes, Y.~Wu, H.~Michalewski, R.~A. Saurous, J.~Sohl-Dickstein, et~al.
\newblock Language model cascades.
\newblock \emph{ICML}, 2022.

\bibitem[Dziri et~al.(2023)Dziri, Lu, Sclar, Li, Jian, Lin, West, Bhagavatula, Bras, Hwang, et~al.]{dziri2023faith}
N.~Dziri, X.~Lu, M.~Sclar, X.~L. Li, L.~Jian, B.~Y. Lin, P.~West, C.~Bhagavatula, R.~L. Bras, J.~D. Hwang, et~al.
\newblock Faith and fate: Limits of transformers on compositionality.
\newblock \emph{arXiv preprint arXiv:2305.18654}, 2023.

\bibitem[Felty(1993)]{felty1993implementing}
A.~P. Felty.
\newblock Implementing tactics and tacticals in a higher-order logic programming language.
\newblock \emph{J. Autom. Reason.}, 11\penalty0 (1):\penalty0 41--81, 1993.
\newblock \doi{10.1007/BF00881900}.
\newblock URL \url{https://doi.org/10.1007/BF00881900}.

\bibitem[First et~al.(2023)First, Rabe, Ringer, and Brun]{first2023baldur}
E.~First, M.~N. Rabe, T.~Ringer, and Y.~Brun.
\newblock Baldur: Whole-proof generation and repair with large language models.
\newblock \emph{CoRR}, abs/2303.04910, 2023.
\newblock \doi{10.48550/arXiv.2303.04910}.
\newblock URL \url{https://doi.org/10.48550/arXiv.2303.04910}.

\bibitem[Frieder et~al.(2023)Frieder, Pinchetti, Griffiths, Salvatori, Lukasiewicz, Petersen, Chevalier, and Berner]{frieder2023mathematical}
S.~Frieder, L.~Pinchetti, R.-R. Griffiths, T.~Salvatori, T.~Lukasiewicz, P.~C. Petersen, A.~Chevalier, and J.~Berner.
\newblock {Mathematical Capabilities of ChatGPT}, 2023.

\bibitem[Ganesalingam and Gowers(2013)]{timprover2013}
M.~Ganesalingam and W.~T. Gowers.
\newblock A fully automatic problem solver with human-style output.
\newblock \emph{CoRR}, abs/1309.4501, 2013.
\newblock URL \url{http://arxiv.org/abs/1309.4501}.

\bibitem[Github(2021)]{copilot}
Github.
\newblock Github copilot · your ai pair programmer, 2021.
\newblock URL \url{https://github.com/features/copilot/}.

\bibitem[Golovneva et~al.(2022)Golovneva, Chen, Poff, Corredor, Zettlemoyer, Fazel{-}Zarandi, and Celikyilmaz]{golovneva2022roscoe}
O.~Golovneva, M.~Chen, S.~Poff, M.~Corredor, L.~Zettlemoyer, M.~Fazel{-}Zarandi, and A.~Celikyilmaz.
\newblock {ROSCOE:} {A} suite of metrics for scoring step-by-step reasoning.
\newblock \emph{CoRR}, abs/2212.07919, 2022.
\newblock \doi{10.48550/arXiv.2212.07919}.
\newblock URL \url{https://doi.org/10.48550/arXiv.2212.07919}.

\bibitem[Gowers(2022)]{timManifesto}
W.~T. Gowers.
\newblock How can it be feasible to find proofs?, 2022.
\newblock URL \url{https://drive.google.com/file/d/1-FFa6nMVg18m1zPtoAQrFalwpx2YaGK4/view}.

\bibitem[Hales(2014)]{flyspeck}
T.~Hales.
\newblock The flyspeck project, Aug 2014.
\newblock URL \url{https://code.google.com/archive/p/flyspeck/}.

\bibitem[Hales(2005)]{Hales2005APO}
T.~C. Hales.
\newblock A proof of the kepler conjecture.
\newblock \emph{Annals of Mathematics}, 162:\penalty0 1063--1183, 2005.

\bibitem[Han et~al.(2022)Han, Rute, Wu, Ayers, and Polu]{proofCoTrain}
J.~M. Han, J.~Rute, Y.~Wu, E.~W. Ayers, and S.~Polu.
\newblock Proof artifact co-training for theorem proving with language models.
\newblock In \emph{The Tenth International Conference on Learning Representations, {ICLR} 2022, Virtual Event, April 25-29, 2022}. OpenReview.net, 2022.
\newblock URL \url{https://openreview.net/forum?id=rpxJc9j04U}.

\bibitem[Harrison et~al.(2014)Harrison, Urban, and Wiedijk]{harrison2014history}
J.~Harrison, J.~Urban, and F.~Wiedijk.
\newblock History of interactive theorem proving.
\newblock In \emph{Computational Logic}, volume~9, pages 135--214, 2014.

\bibitem[{Hugging Face}(2021)]{hfSpace}
{Hugging Face}.
\newblock Hugging face spaces, 2021.
\newblock URL \url{https://huggingface.co/spaces}.

\bibitem[Hullman et~al.(2018)Hullman, Qiao, Correll, Kale, and Kay]{hullman2018pursuit}
J.~Hullman, X.~Qiao, M.~Correll, A.~Kale, and M.~Kay.
\newblock In pursuit of error: A survey of uncertainty visualization evaluation.
\newblock \emph{IEEE transactions on visualization and computer graphics}, 25\penalty0 (1):\penalty0 903--913, 2018.

\bibitem[Jiang et~al.(2022{\natexlab{a}})Jiang, Li, Tworkowski, Czechowski, Odrzyg{\'{o}}zdz, Milos, Wu, and Jamnik]{thor}
A.~Q. Jiang, W.~Li, S.~Tworkowski, K.~Czechowski, T.~Odrzyg{\'{o}}zdz, P.~Milos, Y.~Wu, and M.~Jamnik.
\newblock Thor: Wielding hammers to integrate language models and automated theorem provers.
\newblock \emph{CoRR}, abs/2205.10893, 2022{\natexlab{a}}.
\newblock \doi{10.48550/arXiv.2205.10893}.
\newblock URL \url{https://doi.org/10.48550/arXiv.2205.10893}.

\bibitem[Jiang et~al.(2022{\natexlab{b}})Jiang, Welleck, Zhou, Li, Liu, Jamnik, Lacroix, Wu, and Lample]{dsp}
A.~Q. Jiang, S.~Welleck, J.~P. Zhou, W.~Li, J.~Liu, M.~Jamnik, T.~Lacroix, Y.~Wu, and G.~Lample.
\newblock Draft, sketch, and prove: Guiding formal theorem provers with informal proofs.
\newblock \emph{CoRR}, abs/2210.12283, 2022{\natexlab{b}}.
\newblock \doi{10.48550/arXiv.2210.12283}.
\newblock URL \url{https://doi.org/10.48550/arXiv.2210.12283}.

\bibitem[Kazemi et~al.(2022)Kazemi, Kim, Bhatia, Xu, and Ramachandran]{kazemi2022lambada}
S.~M. Kazemi, N.~Kim, D.~Bhatia, X.~Xu, and D.~Ramachandran.
\newblock Lambada: Backward chaining for automated reasoning in natural language.
\newblock \emph{arXiv preprint arXiv:2212.13894}, 2022.

\bibitem[Kelly et~al.(2023)Kelly, Kumar, Smyth, and Steyvers]{kelly2023capturing}
M.~Kelly, A.~Kumar, P.~Smyth, and M.~Steyvers.
\newblock Capturing humans' mental models of ai: An item response theory approach.
\newblock \emph{FAccT}, 2023.

\bibitem[Kocielnik et~al.(2019)Kocielnik, Amershi, and Bennett]{kocielnik2019will}
R.~Kocielnik, S.~Amershi, and P.~N. Bennett.
\newblock Will you accept an imperfect ai? exploring designs for adjusting end-user expectations of ai systems.
\newblock In \emph{Proceedings of the 2019 CHI Conference on Human Factors in Computing Systems}, pages 1--14, 2019.

\bibitem[Köpf et~al.(2023)Köpf, Kilcher, von Rütte, Anagnostidis, Tam, Stevens, Barhoum, Duc, Stanley, Nagyfi, ES, Suri, Glushkov, Dantuluri, Maguire, Schuhmann, Nguyen, and Mattick]{köpf2023openassistant}
A.~Köpf, Y.~Kilcher, D.~von Rütte, S.~Anagnostidis, Z.-R. Tam, K.~Stevens, A.~Barhoum, N.~M. Duc, O.~Stanley, R.~Nagyfi, S.~ES, S.~Suri, D.~Glushkov, A.~Dantuluri, A.~Maguire, C.~Schuhmann, H.~Nguyen, and A.~Mattick.
\newblock Openassistant conversations -- democratizing large language model alignment, 2023.

\bibitem[Kıcıman et~al.(2023)Kıcıman, Ness, Sharma, and Tan]{kıcıman2023causal}
E.~Kıcıman, R.~Ness, A.~Sharma, and C.~Tan.
\newblock Causal reasoning and large language models: Opening a new frontier for causality, 2023.

\bibitem[Lample et~al.(2022)Lample, Lacroix, Lachaux, Rodriguez, Hayat, Lavril, Ebner, and Martinet]{lample2022hypertree}
G.~Lample, T.~Lacroix, M.-A. Lachaux, A.~Rodriguez, A.~Hayat, T.~Lavril, G.~Ebner, and X.~Martinet.
\newblock Hypertree proof search for neural theorem proving.
\newblock \emph{Advances in Neural Information Processing Systems}, 35:\penalty0 26337--26349, 2022.

\bibitem[Lee et~al.(2022{\natexlab{a}})Lee, Liang, and Yang]{lee2022coauthor}
M.~Lee, P.~Liang, and Q.~Yang.
\newblock Coauthor: Designing a human-ai collaborative writing dataset for exploring language model capabilities.
\newblock In \emph{Proceedings of the 2022 CHI Conference on Human Factors in Computing Systems}, pages 1--19, 2022{\natexlab{a}}.

\bibitem[Lee et~al.(2022{\natexlab{b}})Lee, Srivastava, Hardy, Thickstun, Durmus, Paranjape, Gerard-Ursin, Li, Ladhak, Rong, Wang, Kwon, Park, Cao, Lee, Bommasani, Bernstein, and Liang]{leeInteractive}
M.~Lee, M.~Srivastava, A.~Hardy, J.~Thickstun, E.~Durmus, A.~Paranjape, I.~Gerard-Ursin, X.~L. Li, F.~Ladhak, F.~Rong, R.~E. Wang, M.~Kwon, J.~S. Park, H.~Cao, T.~Lee, R.~Bommasani, M.~Bernstein, and P.~Liang.
\newblock Evaluating human-language model interaction, 2022{\natexlab{b}}.
\newblock URL \url{https://arxiv.org/abs/2212.09746}.

\bibitem[Lee et~al.(2022{\natexlab{c}})Lee, Srivastava, Hardy, Thickstun, Durmus, Paranjape, Gerard-Ursin, Li, Ladhak, Rong, et~al.]{lee2022evaluating}
M.~Lee, M.~Srivastava, A.~Hardy, J.~Thickstun, E.~Durmus, A.~Paranjape, I.~Gerard-Ursin, X.~L. Li, F.~Ladhak, F.~Rong, et~al.
\newblock Evaluating human-language model interaction.
\newblock \emph{arXiv preprint arXiv:2212.09746}, 2022{\natexlab{c}}.

\bibitem[Lewkowycz et~al.(2022)Lewkowycz, Andreassen, Dohan, Dyer, Michalewski, Ramasesh, Slone, Anil, Schlag, Gutman{-}Solo, Wu, Neyshabur, Gur{-}Ari, and Misra]{minerva}
A.~Lewkowycz, A.~Andreassen, D.~Dohan, E.~Dyer, H.~Michalewski, V.~V. Ramasesh, A.~Slone, C.~Anil, I.~Schlag, T.~Gutman{-}Solo, Y.~Wu, B.~Neyshabur, G.~Gur{-}Ari, and V.~Misra.
\newblock Solving quantitative reasoning problems with language models.
\newblock \emph{CoRR}, abs/2206.14858, 2022.
\newblock \doi{10.48550/arXiv.2206.14858}.
\newblock URL \url{https://doi.org/10.48550/arXiv.2206.14858}.

\bibitem[Li et~al.(2023)Li, Chen, Sharma, and Andreas]{li2023lampp}
B.~Z. Li, W.~Chen, P.~Sharma, and J.~Andreas.
\newblock Lampp: Language models as probabilistic priors for perception and action.
\newblock \emph{arXiv e-prints}, pages arXiv--2302, 2023.

\bibitem[Li et~al.(2022)Li, Poesia, Costilla{-}Reyes, Goodman, and Solar{-}Lezama]{lemmaReasoning}
Z.~Li, G.~Poesia, O.~Costilla{-}Reyes, N.~D. Goodman, and A.~Solar{-}Lezama.
\newblock {LEMMA:} bootstrapping high-level mathematical reasoning with learned symbolic abstractions.
\newblock \emph{CoRR}, abs/2211.08671, 2022.
\newblock \doi{10.48550/arXiv.2211.08671}.
\newblock URL \url{https://doi.org/10.48550/arXiv.2211.08671}.

\bibitem[Lin(2004)]{lin2004rouge}
C.-Y. Lin.
\newblock {ROUGE}: A package for automatic evaluation of summaries.
\newblock In \emph{Text Summarization Branches Out}, pages 74--81, Barcelona, Spain, July 2004. Association for Computational Linguistics.
\newblock URL \url{https://aclanthology.org/W04-1013}.

\bibitem[Lipkin et~al.(2023)Lipkin, Wong, Grand, and Tenenbaum]{lipkin2023evaluating}
B.~Lipkin, L.~Wong, G.~Grand, and J.~B. Tenenbaum.
\newblock Evaluating statistical language models as pragmatic reasoners.
\newblock \emph{arXiv preprint arXiv:2305.01020}, 2023.

\bibitem[Liu et~al.(2016)Liu, Bartz, and Duke]{liu2016communicating}
B.~F. Liu, L.~Bartz, and N.~Duke.
\newblock Communicating crisis uncertainty: A review of the knowledge gaps.
\newblock \emph{Public relations review}, 42\penalty0 (3):\penalty0 479--487, 2016.

\bibitem[Lu et~al.(2022)Lu, Qiu, Yu, Welleck, and Chang]{mathsSurvey}
P.~Lu, L.~Qiu, W.~Yu, S.~Welleck, and K.~Chang.
\newblock A survey of deep learning for mathematical reasoning.
\newblock \emph{CoRR}, abs/2212.10535, 2022.
\newblock \doi{10.48550/arXiv.2212.10535}.
\newblock URL \url{https://doi.org/10.48550/arXiv.2212.10535}.

\bibitem[McGrath et~al.(2022)McGrath, Kapishnikov, Toma{\v{s}}ev, Pearce, Wattenberg, Hassabis, Kim, Paquet, and Kramnik]{mcgrath2022acquisition}
T.~McGrath, A.~Kapishnikov, N.~Toma{\v{s}}ev, A.~Pearce, M.~Wattenberg, D.~Hassabis, B.~Kim, U.~Paquet, and V.~Kramnik.
\newblock Acquisition of chess knowledge in alphazero.
\newblock \emph{Proceedings of the National Academy of Sciences}, 119\penalty0 (47):\penalty0 e2206625119, 2022.

\bibitem[Meng et~al.(2022)Meng, Bau, Andonian, and Belinkov]{meng2022locating}
K.~Meng, D.~Bau, A.~Andonian, and Y.~Belinkov.
\newblock Locating and editing factual associations in gpt.
\newblock \emph{Advances in Neural Information Processing Systems}, 35:\penalty0 17359--17372, 2022.

\bibitem[Mikula et~al.(2023)Mikula, Antoniak, Tworkowski, Jiang, Zhou, Szegedy, Kucinski, Milos, and Wu]{magnushammer}
M.~Mikula, S.~Antoniak, S.~Tworkowski, A.~Q. Jiang, J.~P. Zhou, C.~Szegedy, L.~Kucinski, P.~Milos, and Y.~Wu.
\newblock Magnushammer: {A} transformer-based approach to premise selection.
\newblock \emph{CoRR}, abs/2303.04488, 2023.
\newblock \doi{10.48550/arXiv.2303.04488}.
\newblock URL \url{https://doi.org/10.48550/arXiv.2303.04488}.

\bibitem[Miller(2023)]{miller2023explainable}
T.~Miller.
\newblock Explainable ai is dead, long live explainable ai! hypothesis-driven decision support, 2023.

\bibitem[Mirowski et~al.(2023)Mirowski, Mathewson, Pittman, and Evans]{dramatron}
P.~Mirowski, K.~W. Mathewson, J.~Pittman, and R.~Evans.
\newblock Co-writing screenplays and theatre scripts with language models: Evaluation by industry professionals.
\newblock In \emph{Proceedings of the 2023 CHI Conference on Human Factors in Computing Systems}, CHI '23, New York, NY, USA, 2023. Association for Computing Machinery.
\newblock ISBN 9781450394215.
\newblock \doi{10.1145/3544548.3581225}.
\newblock URL \url{https://doi.org/10.1145/3544548.3581225}.

\bibitem[Mitchell et~al.(2022)Mitchell, Lin, Bosselut, Manning, and Finn]{mitchell2022memory}
E.~Mitchell, C.~Lin, A.~Bosselut, C.~D. Manning, and C.~Finn.
\newblock Memory-based model editing at scale.
\newblock In \emph{International Conference on Machine Learning}, pages 15817--15831. PMLR, 2022.

\bibitem[Moravec(1988)]{moravec1988mind}
H.~Moravec.
\newblock \emph{Mind children: The future of robot and human intelligence}.
\newblock Harvard University Press, 1988.

\bibitem[Newell and Simon(1956)]{newell1956logic}
A.~Newell and H.~Simon.
\newblock The logic theory machine--a complex information processing system.
\newblock \emph{IRE Transactions on information theory}, 2\penalty0 (3):\penalty0 61--79, 1956.

\bibitem[OpenAI(2022)]{chatGPT}
OpenAI.
\newblock Introducing chatgpt, 2022.
\newblock URL \url{https://openai.com/blog/chatgpt}.

\bibitem[OpenAI(2023{\natexlab{a}})]{gptPlugins}
OpenAI.
\newblock Chatgpt plugins, 2023{\natexlab{a}}.
\newblock URL \url{https://openai.com/blog/chatgpt-plugins}.

\bibitem[OpenAI(2023{\natexlab{b}})]{openai2023gpt4}
OpenAI.
\newblock Gpt-4 technical report, 2023{\natexlab{b}}.

\bibitem[Ouyang et~al.(2022)Ouyang, Wu, Jiang, Almeida, Wainwright, Mishkin, Zhang, Agarwal, Slama, Ray, Schulman, Hilton, Kelton, Miller, Simens, Askell, Welinder, Christiano, Leike, and Lowe]{ouyang2022training}
L.~Ouyang, J.~Wu, X.~Jiang, D.~Almeida, C.~L. Wainwright, P.~Mishkin, C.~Zhang, S.~Agarwal, K.~Slama, A.~Ray, J.~Schulman, J.~Hilton, F.~Kelton, L.~Miller, M.~Simens, A.~Askell, P.~Welinder, P.~Christiano, J.~Leike, and R.~Lowe.
\newblock Training language models to follow instructions with human feedback, 2022.

\bibitem[Papineni et~al.(2002)Papineni, Roukos, Ward, and Zhu]{papineni2002bleu}
K.~Papineni, S.~Roukos, T.~Ward, and W.~Zhu.
\newblock {BLEU}: a method for automatic evaluation of machine translation.
\newblock In \emph{Proceedings of the 40th Annual Meeting of the Association for Computational Linguistics, July 6-12, 2002, Philadelphia, PA, {USA}}, pages 311--318. {ACL}, 2002.
\newblock \doi{10.3115/1073083.1073135}.
\newblock URL \url{https://aclanthology.org/P02-1040/}.

\bibitem[Paulson(2010)]{paulson2010three}
L.~C. Paulson.
\newblock Three years of experience with sledgehammer, a practical link between automatic and interactive theorem provers.
\newblock In R.~A. Schmidt, S.~Schulz, and B.~Konev, editors, \emph{Proceedings of the 2nd Workshop on Practical Aspects of Automated Reasoning, PAAR-2010, Edinburgh, Scotland, UK, July 14, 2010}, volume~9 of \emph{EPiC Series in Computing}, pages 1--10. EasyChair, 2010.
\newblock \doi{10.29007/tnfd}.
\newblock URL \url{https://doi.org/10.29007/tnfd}.

\bibitem[Piantadosi and Hill(2022)]{piantadosi2022meaning}
S.~T. Piantadosi and F.~Hill.
\newblock Meaning without reference in large language models.
\newblock \emph{CoRR}, abs/2208.02957, 2022.
\newblock \doi{10.48550/arXiv.2208.02957}.
\newblock URL \url{https://doi.org/10.48550/arXiv.2208.02957}.

\bibitem[Poesia and Goodman(2022)]{peano}
G.~Poesia and N.~D. Goodman.
\newblock Peano: Learning formal mathematical reasoning.
\newblock \emph{CoRR}, abs/2211.15864, 2022.
\newblock \doi{10.48550/arXiv.2211.15864}.
\newblock URL \url{https://doi.org/10.48550/arXiv.2211.15864}.

\bibitem[Polu and Sutskever(2020)]{polu2020}
S.~Polu and I.~Sutskever.
\newblock Generative language modeling for automated theorem proving.
\newblock \emph{CoRR}, abs/2009.03393, 2020.
\newblock URL \url{https://arxiv.org/abs/2009.03393}.

\bibitem[Polu et~al.(2022)Polu, Han, Zheng, Baksys, Babuschkin, and Sutskever]{polu2022}
S.~Polu, J.~M. Han, K.~Zheng, M.~Baksys, I.~Babuschkin, and I.~Sutskever.
\newblock Formal mathematics statement curriculum learning.
\newblock \emph{CoRR}, abs/2202.01344, 2022.
\newblock URL \url{https://arxiv.org/abs/2202.01344}.

\bibitem[Radford et~al.(2019)Radford, Wu, Child, Luan, Amodei, and Sutskever]{Radford2019LanguageMA}
A.~Radford, J.~Wu, R.~Child, D.~Luan, D.~Amodei, and I.~Sutskever.
\newblock Language models are unsupervised multitask learners.
\newblock In \emph{OpenAI blog}, 2019.

\bibitem[Ringer et~al.(2020)Ringer, Sanchez{-}Stern, Grossman, and Lerner]{ringer2020replica}
T.~Ringer, A.~Sanchez{-}Stern, D.~Grossman, and S.~Lerner.
\newblock Replica: {REPL} instrumentation for coq analysis.
\newblock In J.~Blanchette and C.~Hritcu, editors, \emph{Proceedings of the 9th {ACM} {SIGPLAN} International Conference on Certified Programs and Proofs, {CPP} 2020, New Orleans, LA, USA, January 20-21, 2020}, pages 99--113. {ACM}, 2020.
\newblock \doi{10.1145/3372885.3373823}.
\newblock URL \url{https://doi.org/10.1145/3372885.3373823}.

\bibitem[Schulz(2002)]{Schulz2002EA}
S.~Schulz.
\newblock E - a brainiac theorem prover.
\newblock \emph{AI Commun.}, 15:\penalty0 111--126, 2002.

\bibitem[Sevastjanova and El-Assady(2022)]{sevastjanova2022beware}
R.~Sevastjanova and M.~El-Assady.
\newblock Beware the rationalization trap! when language model explainability diverges from our mental models of language.
\newblock \emph{arXiv preprint arXiv:2207.06897}, 2022.

\bibitem[Shen and Wu(2023)]{shen2023parachute}
H.~Shen and T.~Wu.
\newblock Parachute: Evaluating interactive human-lm co-writing systems.
\newblock \emph{arXiv e-prints}, pages arXiv--2303, 2023.

\bibitem[Si et~al.(2022)Si, Gan, Yang, Wang, Wang, Boyd-Graber, and Wang]{si2022prompting}
C.~Si, Z.~Gan, Z.~Yang, S.~Wang, J.~Wang, J.~Boyd-Graber, and L.~Wang.
\newblock Prompting gpt-3 to be reliable.
\newblock \emph{arXiv preprint arXiv:2210.09150}, 2022.

\bibitem[Spiegelhalter(2017)]{spiegelhalter_risk_2017}
D.~Spiegelhalter.
\newblock Risk and {{Uncertainty Communication}}.
\newblock \emph{Annual Review of Statistics and Its Application}, 4\penalty0 (1):\penalty0 31--60, 2017.
\newblock ISSN 2326-8298, 2326-831X.
\newblock \doi{10.1146/annurev-statistics-010814-020148}.
\newblock URL \url{http://www.annualreviews.org/doi/10.1146/annurev-statistics-010814-020148}.

\bibitem[Sutcliffe and Suttner(2001)]{sutcliffe2001evaluating}
G.~Sutcliffe and C.~Suttner.
\newblock Evaluating general purpose automated theorem proving systems.
\newblock \emph{Artificial intelligence}, 131\penalty0 (1-2):\penalty0 39--54, 2001.

\bibitem[Taori et~al.(2023)Taori, Gulrajani, Zhang, Dubois, Li, Guestrin, Liang, and Hashimoto]{alpaca}
R.~Taori, I.~Gulrajani, T.~Zhang, Y.~Dubois, X.~Li, C.~Guestrin, P.~Liang, and T.~B. Hashimoto.
\newblock Stanford alpaca: An instruction-following llama model.
\newblock \url{https://github.com/tatsu-lab/stanford_alpaca}, 2023.

\bibitem[Tarski(1969)]{tarski1969truth}
A.~Tarski.
\newblock Truth and proof.
\newblock \emph{Scientific American}, 220\penalty0 (6):\penalty0 63--77, 1969.

\bibitem[Touvron et~al.(2023)Touvron, Lavril, Izacard, Martinet, Lachaux, Lacroix, Rozi{\`e}re, Goyal, Hambro, Azhar, et~al.]{touvron2023llama}
H.~Touvron, T.~Lavril, G.~Izacard, X.~Martinet, M.-A. Lachaux, T.~Lacroix, B.~Rozi{\`e}re, N.~Goyal, E.~Hambro, F.~Azhar, et~al.
\newblock Llama: Open and efficient foundation language models.
\newblock \emph{arXiv e-prints}, pages arXiv--2302, 2023.

\bibitem[Vasconcelos et~al.(2023)Vasconcelos, Bansal, Fourney, Liao, and Wortman~Vaughan]{vasconcelos2023generation}
H.~Vasconcelos, G.~Bansal, A.~Fourney, Q.~V. Liao, and J.~Wortman~Vaughan.
\newblock Generation probabilities are not enough: Exploring the effectiveness of uncertainty highlighting in ai-powered code completions.
\newblock \emph{arXiv e-prints}, pages arXiv--2302, 2023.

\bibitem[Wang(1960)]{wangMaths}
H.~Wang.
\newblock Toward mechanical mathematics.
\newblock \emph{IBM J. Res. Dev.}, 4\penalty0 (1):\penalty0 2–22, jan 1960.
\newblock ISSN 0018-8646.
\newblock \doi{10.1147/rd.41.0002}.
\newblock URL \url{https://doi.org/10.1147/rd.41.0002}.

\bibitem[Wei et~al.(2022)Wei, Wang, Schuurmans, Bosma, Xia, Chi, Le, Zhou, et~al.]{weichain}
J.~Wei, X.~Wang, D.~Schuurmans, M.~Bosma, F.~Xia, E.~H. Chi, Q.~V. Le, D.~Zhou, et~al.
\newblock Chain-of-thought prompting elicits reasoning in large language models.
\newblock In \emph{Advances in Neural Information Processing Systems}, 2022.

\bibitem[Welleck et~al.(2022{\natexlab{a}})Welleck, Liu, Lu, Hajishirzi, and Choi]{naturalprover}
S.~Welleck, J.~Liu, X.~Lu, H.~Hajishirzi, and Y.~Choi.
\newblock Naturalprover: Grounded mathematical proof generation with language models, 2022{\natexlab{a}}.
\newblock URL \url{https://arxiv.org/abs/2205.12910}.

\bibitem[Welleck et~al.(2022{\natexlab{b}})Welleck, Liu, Lu, Hajishirzi, and Choi]{welleck2022naturalprover}
S.~Welleck, J.~Liu, X.~Lu, H.~Hajishirzi, and Y.~Choi.
\newblock Naturalprover: Grounded mathematical proof generation with language models.
\newblock In \emph{NeurIPS}, 2022{\natexlab{b}}.
\newblock URL \url{http://papers.nips.cc/paper\_files/paper/2022/hash/1fc548a8243ad06616eee731e0572927-Abstract-Conference.html}.

\bibitem[Wilder et~al.(2021{\natexlab{a}})Wilder, Horvitz, and Kamar]{wilder2020learning}
B.~Wilder, E.~Horvitz, and E.~Kamar.
\newblock Learning to complement humans.
\newblock In \emph{Proceedings of the Twenty-Ninth International Conference on International Joint Conferences on Artificial Intelligence}, pages 1526--1533, 2021{\natexlab{a}}.

\bibitem[Wilder et~al.(2021{\natexlab{b}})Wilder, Horvitz, and Kamar]{wilder2021learning}
B.~Wilder, E.~Horvitz, and E.~Kamar.
\newblock Learning to complement humans.
\newblock In \emph{Proceedings of the Twenty-Ninth International Conference on International Joint Conferences on Artificial Intelligence}, pages 1526--1533, 2021{\natexlab{b}}.

\bibitem[Wong et~al.(2023)Wong, Grand, Lew, Goodman, Mansinghka, Andreas, and Tenenbaum]{wong2023word}
L.~Wong, G.~Grand, A.~K. Lew, N.~D. Goodman, V.~K. Mansinghka, J.~Andreas, and J.~B. Tenenbaum.
\newblock From word models to world models: Translating from natural language to the probabilistic language of thought.
\newblock \emph{arXiv preprint arXiv:2306.12672}, 2023.

\bibitem[Wu et~al.(2022)Wu, Jiang, Li, Rabe, Staats, Jamnik, and Szegedy]{autoform}
Y.~Wu, A.~Q. Jiang, W.~Li, M.~N. Rabe, C.~Staats, M.~Jamnik, and C.~Szegedy.
\newblock Autoformalization with large language models.
\newblock \emph{CoRR}, abs/2205.12615, 2022.
\newblock \doi{10.48550/arXiv.2205.12615}.
\newblock URL \url{https://doi.org/10.48550/arXiv.2205.12615}.

\bibitem[Wu et~al.(2023)Wu, Geiger, Potts, and Goodman]{wu2023interpretability}
Z.~Wu, A.~Geiger, C.~Potts, and N.~D. Goodman.
\newblock Interpretability at scale: Identifying causal mechanisms in alpaca.
\newblock \emph{arXiv preprint arXiv:2305.08809}, 2023.

\bibitem[Xiao et~al.(2022)Xiao, Liang, Bhatt, Neiswanger, Salakhutdinov, and Morency]{xiao-etal-2022-uncertainty}
Y.~Xiao, P.~P. Liang, U.~Bhatt, W.~Neiswanger, R.~Salakhutdinov, and L.-P. Morency.
\newblock Uncertainty quantification with pre-trained language models: A large-scale empirical analysis.
\newblock In \emph{Findings of the Association for Computational Linguistics: EMNLP 2022}, pages 7273--7284, Abu Dhabi, United Arab Emirates, Dec. 2022. Association for Computational Linguistics.
\newblock URL \url{https://aclanthology.org/2022.findings-emnlp.538}.

\bibitem[Yao et~al.(2023)Yao, Yu, Zhao, Shafran, Griffiths, Cao, and Narasimhan]{yao2023tree}
S.~Yao, D.~Yu, J.~Zhao, I.~Shafran, T.~L. Griffiths, Y.~Cao, and K.~Narasimhan.
\newblock Tree of thoughts: Deliberate problem solving with large language models, 2023.

\bibitem[Zelikman et~al.(2022{\natexlab{a}})Zelikman, Huang, Poesia, Goodman, and Haber]{parsel}
E.~Zelikman, Q.~Huang, G.~Poesia, N.~D. Goodman, and N.~Haber.
\newblock Parsel: {A} unified natural language framework for algorithmic reasoning.
\newblock \emph{CoRR}, abs/2212.10561, 2022{\natexlab{a}}.
\newblock \doi{10.48550/arXiv.2212.10561}.
\newblock URL \url{https://doi.org/10.48550/arXiv.2212.10561}.

\bibitem[Zelikman et~al.(2022{\natexlab{b}})Zelikman, Wu, Mu, and Goodman]{zelikman2022star}
E.~Zelikman, Y.~Wu, J.~Mu, and N.~D. Goodman.
\newblock Star: Bootstrapping reasoning with reasoning, 2022{\natexlab{b}}.

\bibitem[Zerilli et~al.(2022)Zerilli, Bhatt, and Weller]{umangTrust}
J.~Zerilli, U.~Bhatt, and A.~Weller.
\newblock How transparency modulates trust in artificial intelligence.
\newblock \emph{Patterns}, 3\penalty0 (4):\penalty0 100455, 2022.

\bibitem[Zheng et~al.(2022)Zheng, Han, and Polu]{minif2f}
K.~Zheng, J.~M. Han, and S.~Polu.
\newblock {miniF2F: a cross-system benchmark for formal Olympiad-level mathematics}.
\newblock In \emph{The Tenth International Conference on Learning Representations, {ICLR} 2022, Virtual Event, April 25-29, 2022}. OpenReview.net, 2022.
\newblock URL \url{https://openreview.net/forum?id=9ZPegFuFTFv}.

\bibitem[Zhou et~al.(2023)Zhou, Sch{\"a}rli, Hou, Wei, Scales, Wang, Schuurmans, Bousquet, Le, and Chi]{zhou2022least}
D.~Zhou, N.~Sch{\"a}rli, L.~Hou, J.~Wei, N.~Scales, X.~Wang, D.~Schuurmans, O.~Bousquet, Q.~Le, and E.~Chi.
\newblock Least-to-most prompting enables complex reasoning in large language models.
\newblock \emph{ICLR}, 2023.

\end{thebibliography}
\bibliographystyle{abbrvnat}

\newpage
\appendix

\textbf{\LARGE{SUPPLEMENTARY INFORMATION}}

\section{Primer on Proof Assistants}
\label{app: primer}

Push-button automation in mathematics has long been a dream and has an extensive history~\citep{Bledsoe1977NonResolutionTP, Bundy1988TheUO, newell1956logic, Bundy1983TheCM, Bundy1993RipplingAH, wangMaths, davis1962machine, tarski1969truth, timprover2013, Moura2008Z3AE, Schulz2002EA}. However, the initial goal of specifying problems in a sufficiently expressive logic and solving them routinely with fully automated theorem provers was not realised~\citep{harrison2014history}. This led to a shift in focus towards \textit{interactive} theorem provers (ITPs), or \textit{``proof assistants''}: humans specify the high-level structures of proofs and rely on machines to close out tiny details and weave together components~\citep{felty1993implementing, delahaye2000tactic, paulson2010three}. In this way, humans and machines collaborate to produce mechanically-verifiable proofs. However, adoption in the mathematical community has been slow as ITPs traditionally suffered from two weak points. First, because of their precise nature and relatively weak automation, writing formal proofs in interactive theorem provers is an extremely arduous and expensive task~(e.g., verifying the correctness of Hales's proof of the Kepler conjecture~\citep{Hales2005APO} took a group of mathematicians and computer scientists eleven years~\citep{flyspeck}). Secondly, while ITPs can check whether proofs are correct, they provide little assistance for finding the proofs to truly difficult problems: people usually understand informal proofs before translating them into formal ones instead of directly working out formal proofs. 

With the rise of language models, the role of machines in assisting mathematicians has been reconsidered: can language models also automate high-level mathematical reasoning?
While great strides have been made \citep{dsp, autoform, magnushammer, thor, minerva, polu2020, proofCoTrain, polu2022, mathsSurvey, naturalprover, lemmaReasoning, peano, parsel, lample2022hypertree}, consistent full and correct automation has not yet been met; at present, many language models, on their own, struggle not only on truly ``hard'' graduate level problems \citep{frieder2023mathematical}, but also on simple mathematical concepts such as counting \citep{bubeck2023sparks}. 

However, this does not rule out the possibility that they can be \textit{useful}. The potential role of computerised mathematical \textit{assistants} is being re-imagined -- human-machine partnerships where neither comes up with the proof alone. Yet, to adequately begin to explore these kinds of relationships necessitates studying actual mathematicians' interactions. While the assistance potential of ITPs has been evaluated with humans \citep{sutcliffe2001evaluating, aitken1998interactive, beckert2015usability}, we aim for a platform to facilitate rapid evaluation of LLMs in particular. The space of LLMs is changing rapidly, from new base models~\citep{openai2023gpt4, anil2023palm, touvron2023llama, alpaca} to new ways of linking them together and leveraging the output of these systems, e.g., ~\citep{dohan2022language, kazemi2022lambada, li2023lampp, lipkin2023evaluating}, to new prompting techniques \citep{weichain, zhou2022least, yao2023tree}, and more. As such, there is a need for a reliable scaffold to permit interactive evaluation of these human-machine interaction~\citep{lee2022evaluating, clark+etal:18, cohn2023dialectical, shen2023parachute, lee2022coauthor}. It is this notion -- interactive evaluation of humans with LLMs, specifically in the context of proof assistance -- that we turn to next.

\section{An Example Survey Problem}
\label{example survey problem}
\begin{center}
    \fbox{\begin{minipage}{28em}
    Let $x_1, x_2, \ldots, x_n \in \mathbb{R}$ be real numbers which are all positive.
    
    Let $A_n$ be the arithmetic mean of $x_1, x_2, \ldots, x_n$.

    Let $G_n$ be the geometric mean of $x_1, x_2, \ldots, x_n$.

    Show that
    \[
        A_n \geq G_n
    \]
    with equality holding iff:
    \[
        \forall i,j\in \{1, 2, \ldots, n\}: x_i = x_j,
    \]
    that is, iff all terms are equal.
    \end{minipage}}
\end{center}

\section{An Overly Verbose Response}
The following conversation was between a human and GPT-4, the response by the latter was later considered overly verbose:

\begin{longtable}{p{0.05\linewidth}  p{0.9\linewidth}}
\raisebox{-.62\totalheight}{\includegraphics[width=\linewidth]{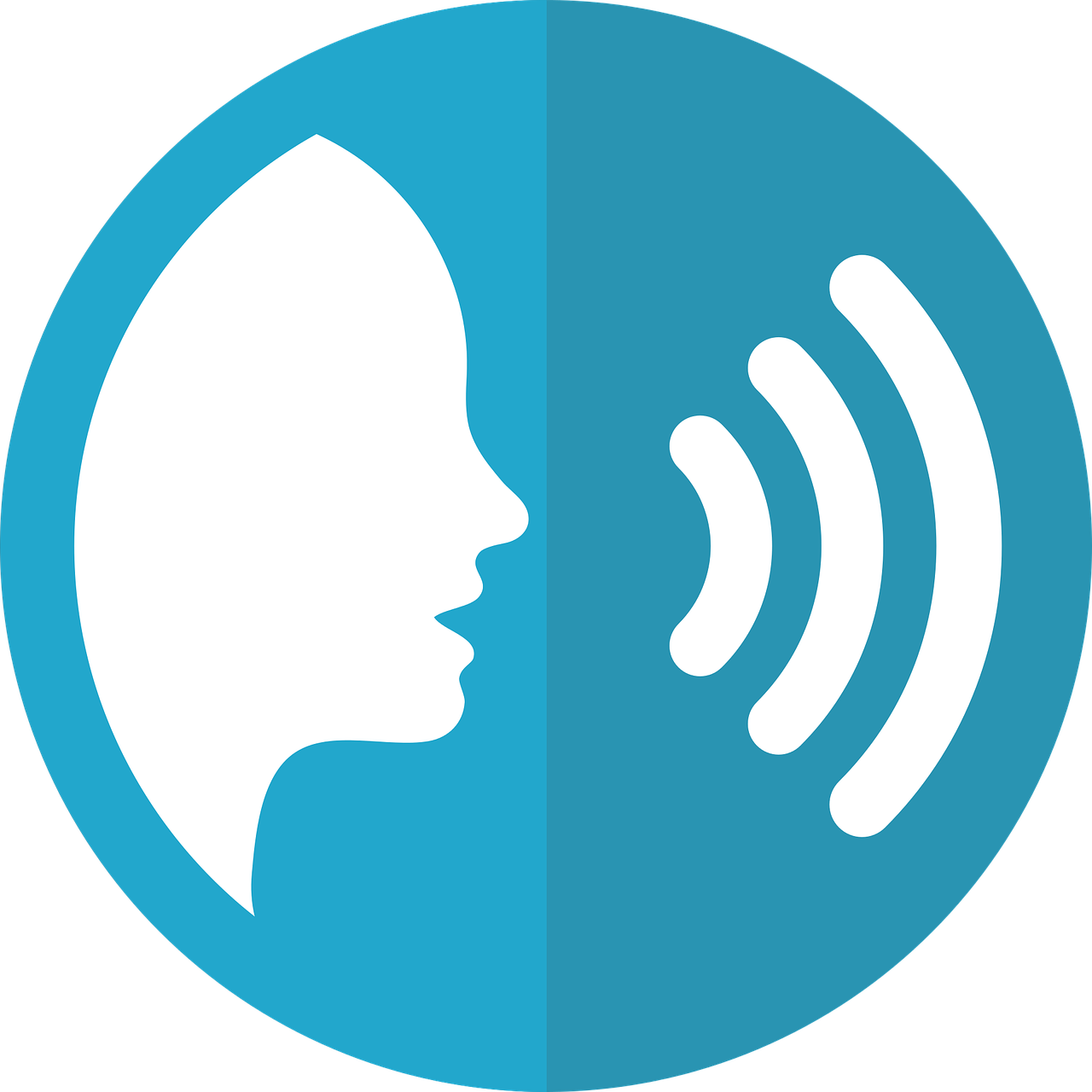}} & How should I go about finding such polynomial? \\\\
\raisebox{-.62\totalheight}{\includegraphics[width=\linewidth]{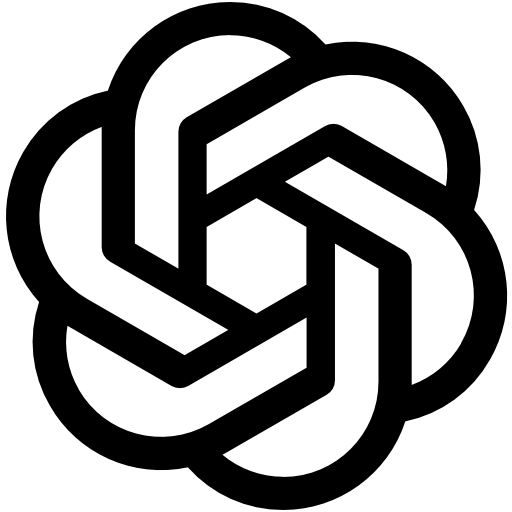}} & To find a polynomial, you need to have some information about its properties, such as its degree, coefficients, roots, or any specific behaviour it exhibits. Here are some general steps to follow: \\\\&
1. Determine the degree of the polynomial: The degree of the polynomial is the highest power of the variable (usually x) in the polynomial. It determines the number of roots the polynomial has and the overall shape of its graph.\\\\&
2. Identify any known roots: If you know any roots of the polynomial (i.e., values of x for which the polynomial equals zero), you can use them to write the polynomial in factored form. For example, if a polynomial has roots at x = 1 and x = -2, it can be written as $P(x) = a(x - 1)(x + 2)$, where a is a constant.\\\\&
3. Use any additional information: If you have any other information about the polynomial, such as its value at a specific point or the behaviour of its graph, use this information to further refine the polynomial. For example, if you know that the polynomial passes through the point (3, 4), you can plug in x = 3 and y = 4 into the polynomial equation and solve for any unknown coefficients.\\\\&
4. Write the polynomial in standard form: Once you have determined the polynomial's factored form and found any unknown coefficients, you can expand the polynomial to write it in standard form (i.e., as a sum of terms with decreasing powers of x).\\\\&
If you can provide more specific information about the polynomial you are trying to find, I can help guide you through the process more precisely.
\end{longtable}

\section{User Guide for \texttt{CheckMate}}
\label{user guide}

We include a user guide for those wishing to extend \texttt{CheckMate} for their own interactive evaluation tasks. Our guide maps directly onto our open-source \href{https://github.com/collinskatie/checkmate}{repository}. 

We describe several hypothetical scenarios in which someone is extending \texttt{CheckMate}. We reiterate that any application of such interaction with real human participants ought to check with the affiliated institutions' ethics review. 

\subsection{Hypothetical extensions}
\label{hypothetical extensions}

\paragraph{New base tasks} If you would like to have different tasks than ProofWiki, you can replace \texttt{data/problems/} with your files. The most straightforward way to do so is to have each new base task as its own file; we encourage files to be associated with a unique task number ID. If you have a grouping for the tasks which you would like a participant to be able to select (e.g., in our case, participants selected a mathematical topic and were only shown problems from that topic), you can code which problems are associated with which topic in \texttt{problems\_per\_topic} in \texttt{experiment.py}. 

\paragraph{New rating scales} If you would like to change the ratings, you can change the instructions in and add different rating endpoints in \texttt{constants.py}, and change the Gradio object type in \texttt{pipeline\_for\_model} in \texttt{experiment.py} (e.g., see that Likert scales like ``ai\_corr\_rating'' which holds user correctness per interaction is coded as a Radio button). That is all that is needed if you keep two ratings per generation; however, if you would like to add or remove ratings, you need to do a bit more programming. You can add new a new rating by going into \texttt{experiment.py} and adding a gradio radio element with the new rating options. You can refer to how \texttt{ai\_corr\_rating} is constructed and stored. Similarly, to delete a rating, the easiest approach is to find how its associated rating options are used in \texttt{experiment.py} and delete all related variables. 
Note, if changing ratings, you will also need to update saving and variable visibility toggling, which is somewhat hard-coded per challenges noted below.

\paragraph{Different models to evaluate} If you would like to vary the models which are evaluated, you can do so by changing the model tags in \texttt{model\_options} in \texttt{constants.py}, and adding associated API calls in \texttt{model\_generate.py}. Note, if you would like to use the OpenAI API, \textit{you will need your own API key}. Additional models can be used if you substitute the OpenAI calls with calls to the additional models. You can play with the prompt used in \texttt{model\_generate.py}, and temperature can be varied if using the OpenAI Completion API in \texttt{constants.py}.

If you would like a different set-up than evaluating the preferences across three models, you can change the number of models before preferences by varying the models passed to \texttt{a\_single\_problem}. 

If you would instead like to remove the final preferences entirely, and just have a rating per model, you can remove the ``Final prefence'' tab  in \texttt{a\_single\_problem} in \texttt{experiment.py}. 

\paragraph{New instructions} It is particularly straightforward to swap out new instructions. To do so, you can go to \texttt{constants.py} and see \texttt{plaintxt\_instructions}. Each entry in the array is a new instruction page. You can track where this information is propogated through \texttt{instruction\_pages} in \texttt{constants.py} and then in \texttt{experiment.py} as you adjust your instructions. Note, as we were circulating the study through volunteer channels, we also included text in a \href{https://sites.google.com/view/interactivemathevals/}{Google Webpage}. This can be made quickly through Google Pages, and requires no coding experience. 

\paragraph{Hosting} 
\texttt{CheckMate} can be adapted to be hosted on a personal server or using a public offering, such as Hugging Face Space~\citep{hfSpace}. Choosing a server that is geographically closer to the target audience will help reduce unwanted interaction lag.

\subsection{Implementation Challenges}

Additionally, we detail several implementation challenges we faced. While the interface can be readily co-opted for new tasks as discussed above, we hope that shedding light on some of the challenges we faced when designing the platform may be of use to those who wish to further customise \texttt{CheckMate} for their own tasks. Note, as the Gradio platform is rapidly evolving, these implementation challenges may become obsolete shortly; we recommend checking the latest capabilities. 

\begin{itemize}
    \item Multiple pages: A common structure in psychological studies is to have multiple pages of instructions and different tasks; the biggest challenge we faced was how to design multiple pages. In particular, the way that Gradio seemed to work -- at least at the time of our construction -- is that variables needed to be shared across pages. So we had to instantiate all key variables and the start and then iteratively show/hide components. It is possible that a more efficient way already was possible, or that one will be developed after the release of this working paper. At present, however, our code does offer a functional starting point for multiple pages.
    \item Saving: Relatedly, we needed to ensure that participants' responses were saved over the course of the study; however, due to state changes, this meant that we needed to be extra careful with saving and deduplicating the final results.
    \item Latency: A common annoyance we heard from participants is that the study took too long to load. There are two core reasons for this: 1) each interaction queried a language model API (see Appendix) which comes with inevitable overhead at present, and 2) as the web server was hosted in western United States, participants seeking to partake from other countries, e.g., France, reported higher latency. Frustrating wait times may help explain the low number of problems each participant took on. Better measuring and addressing latency, as suggested in ~\citep{leeInteractive, shen2023parachute}, are important grounds for future work.
\end{itemize}

\section{Additional Details on Survey Set-Up}
\label{app: additional details}

We hosted the study using Gradio~\citep{abid2019gradio}. We ran the study between April 7, 2023 and April 24, 2023. We circulated a \href{https://sites.google.com/view/interactivemathevals/}{landing page}, which included a link to the actual instance of \texttt{CheckMate}. 

\subsection{Rating Scales}
\label{dims}

We include the labels which were presented to participants for each of the rating dimensions, along with the question. Before participating, users rated their confidence in being able to solve the problem themselves. After interacting with a single model, they rated the correctness and perceived usefulness \textit{of each generation}. And after interacting with the set of three models, they rated overall preference. 

\subsubsection*{Before Generation}

\textbf{``Question: Before interacting with the AI -- how confident are you that \textit{you} could solve this problem \textit{entirely on your own}, with your current knowledge base and no extra assistance?''}
\begin{itemize}
    \item ``(0) Definitely could not solve on my own''
    \item ``(1) Very unlikely to be able to solve on my own''
    \item ``(2) Unlikely to be able to solve on my own'' 
    \item ``(3) May be able to solve on my own''
    \item ``(4) Likely be able to solve on my own''
    \item ``(5) Very likely to be able to solve on my own''
    \item ``(6) Definitely can solve on my own''
\end{itemize}

\subsubsection*{Per Generation Ratings, Per Model} 

\textbf{``Question 1: How correct (i.e., mathematically sound) is the generation?''}
\begin{itemize}
    \item ``(0) N/A - this response does not contain any mathematical information''
    \item ``(1) Completely incorrect or nonsensical''
    \item ``(2) Multiple critical maths errors''
    \item ``(3) At least one critical math error or multiple small errors''
    \item ``(4) One or more minor errors, but otherwise mostly correct''
    \item ``(5) One or two minor errors, but almost entirely correct''
    \item ``(6) Completely correct''
\end{itemize}

\textbf{``Question 2: How helpful would this AI generated response be towards helping someone solve this problem? If you already know how to solve the problem, evaluate this as if you were an undergraduate mathematics student encountering this problem for the first time.''}

\begin{itemize}
    \item ``(0) Actively harmful''
    \item ``(1) Very harmful''
    \item ``(2) Somewhat harmful''
    \item ``(3) Unlikely to help, but unlikely to hurt''
    \item ``(4) Somewhat helpful''
    \item ``(5) Very helpful''
    \item ``(6) Definitely helpful'' 
\end{itemize}

\subsubsection*{Cross-Model Preference}

After interacting blindly with the three models, participants were asked \textbf{``You will now rate which model(s) you prefer as a mathematical assistant. 1 = best, 3 = worst. You can assign the same rating if you think two (or more) models tied''}. Ratings were provided via drop-down options (including 1, 2, 3). 

\subsection{Participant Recruitment and Additional Details} 
\label{recruit}

\begin{figure}[!htb]
    \centering
    \includegraphics[width=0.95\linewidth]{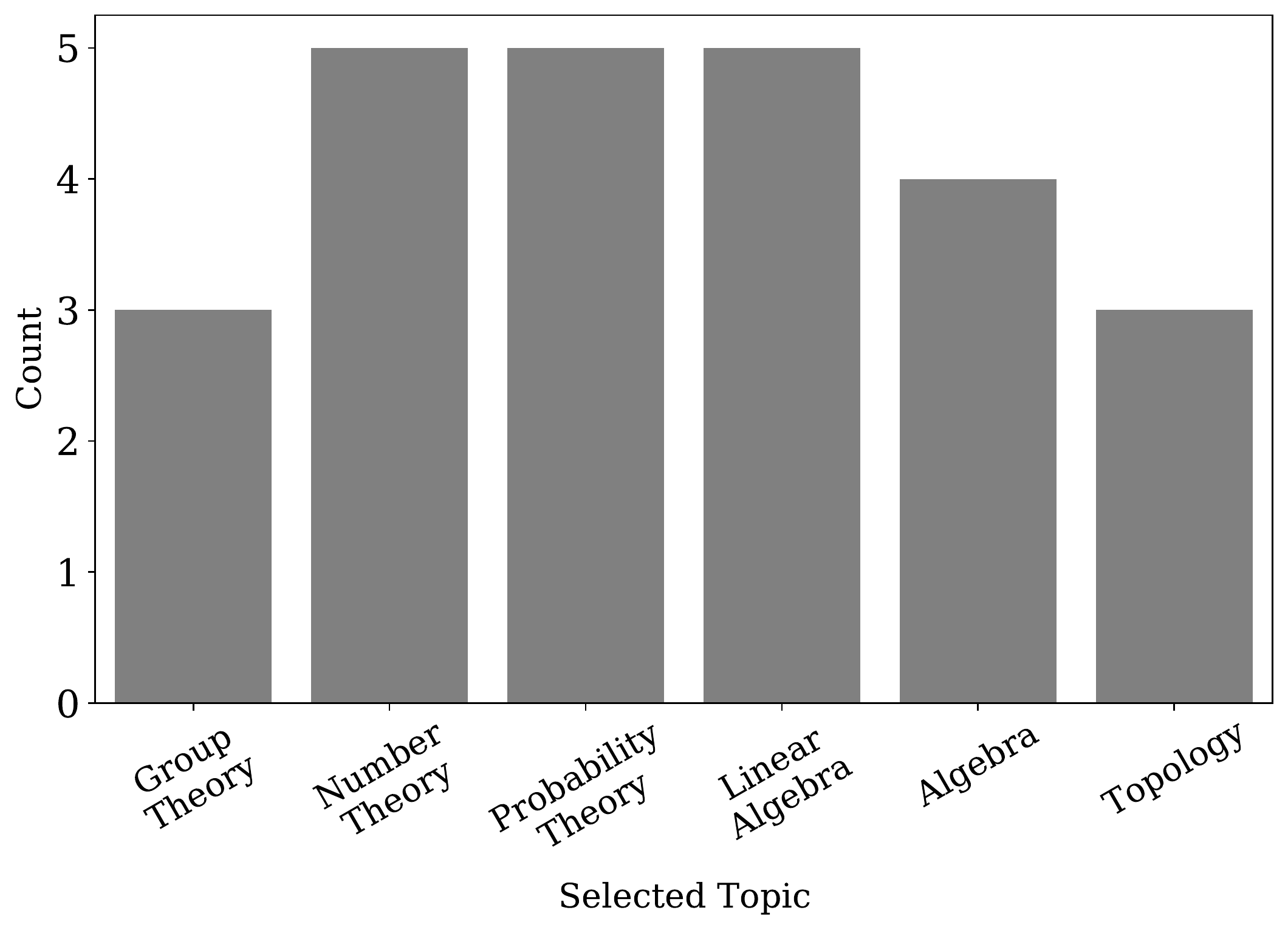}
    \caption{Topics selected across participants; each participant selected a single topic to evaluate models over.}
    \label{fig:selected_topics}
\end{figure}

We provide additional details on our participant recruitment and \texttt{MathConverse} data processing. All participation was unpaid and entirely voluntary. Participants were recruited via authors' connections: We circulated the study through the University of Cambridge Mathematics Department Mailing List, as well as the team channel for the \href{https://wtgowers.github.io/human-style-atp/}{Human-Oriented Automated Theorem Proving project}. The study was also posted on the Lean Zulip channel, where a large community of mathematicians and computer scientists congregate to discuss issues related to formal mathematics. Additionally, the study was circulated amongst mathematics friends at MIT, Oxford, University College London, University of Vienna, École Polytechnique, and Carnegie Mellon University, and elsewhere in Berlin and Paris. We also sent the study to some machine learning students who had mathematics background at the University of Cambridge, MIT, and Princeton; here, participants may not have a formal degree in mathematics but have usually been exposed to mathematics to some degree (i.e., they are not pure amateurs). It is not clear who uptook the survey, as we did not save any personal information, beyond the level of formal mathematics education and experience playing with AI systems, for privacy reasons\footnote{We faced a tradeoff between user tracking for data correlation and privacy. It is possible that a unique participant number that can only be used to correlate participant sessions but does not personally identify participants could be employed in the future without compromising the integrity of our data protection and privacy objectives. Informed consent would be critical, though it is not clear how the perception of tracking, no matter how clearly users are informed of its benign nature, would affect the willingness of users to participate or interact freely with the models. We chose to maximise privacy, participation and freeness of interaction in this study by not tracking participants at all.}. However, we connected with some of the experienced mathematicians who participated for post-survey testimonials about their experience during the study (see the Post-Survey Testimonials below). In the end, we achieved decent coverage across the question topics (see Figure \ref{fig:selected_topics}).

Additionally, of the 25 entries received, 16 resulted in at least one full round of model preferences (i.e., interacting with all three models, and ranking preference); we keep all 25 in the individual interaction analyses as they still provide rich data. Specifically, in cases where preference ratings were not provided, or were provided but a participant had not interacted with all three models (in which case, preference scores are meaningless), we keep the scores for the individual ratings and ignored the preference ratings. In one instance, a participant did not rate all generations from the model; we also discarded this case. 
We emphasise for anyone exploring \texttt{MathConverse} that -- as expected with volunteer-based surveys -- the data can be noisy. For instance, despite informing participants, some did not fully understand that the model was not aware of the problem statement unless the user entered relevant info into the chat window.

On average, we find that our volunteers evaluate 3.1 problems ($\pm 2.2$ problems corresponds to one standard deviation) before stopping, that is, typically just one round of going through each model and providing final preferences. Participants are not informed which model they are evaluating at any time. For any given model, participants interacted for an average of 3.4 queries ($\pm 2.4$; maximum 12 interactions taken).

\subsection{Language Model Prompting}
\label{llm prompting}

We include further details on the prompts used for each language model, and the way in which we call the associated API. The model IDs were as follows: ``text-davinci-003'' (InstructGPT), ``gpt-3.5-turbo'' (ChatGPT), and ``gpt-4'' (GPT-4). Sampling temperature was set to 0.0, and the maximum number of tokens permitted to generate was 512. 

\paragraph{InstructGPT} We interact with Instruct-GPT through the Completion API call. Our instruction prompt starts with ``Help a professional mathematician solve a problem:''.

\paragraph{ChatGPT and GPT-4} We query both models with the ChatCompletion interface. The system was setup with the instruction, ``You are an assistant to a professional mathematician.'' Further generations were elicited through the user-assistant paradigm. 

\subsection{Further Details on Tasks Chosen}
The ProofWiki problems were chosen to be self-contained, and relatively brief. They consist of a mixture of well-known theorems and less famous, exercise-type problems, to provide a spectrum of different problems. The six topics are what typical undergraduate mathematical curricula cover.

\subsection{Example Interface Screens}
\label{interface screenshots}
\begin{figure}[!htb]
\centering
  \includegraphics[width=.95\linewidth]{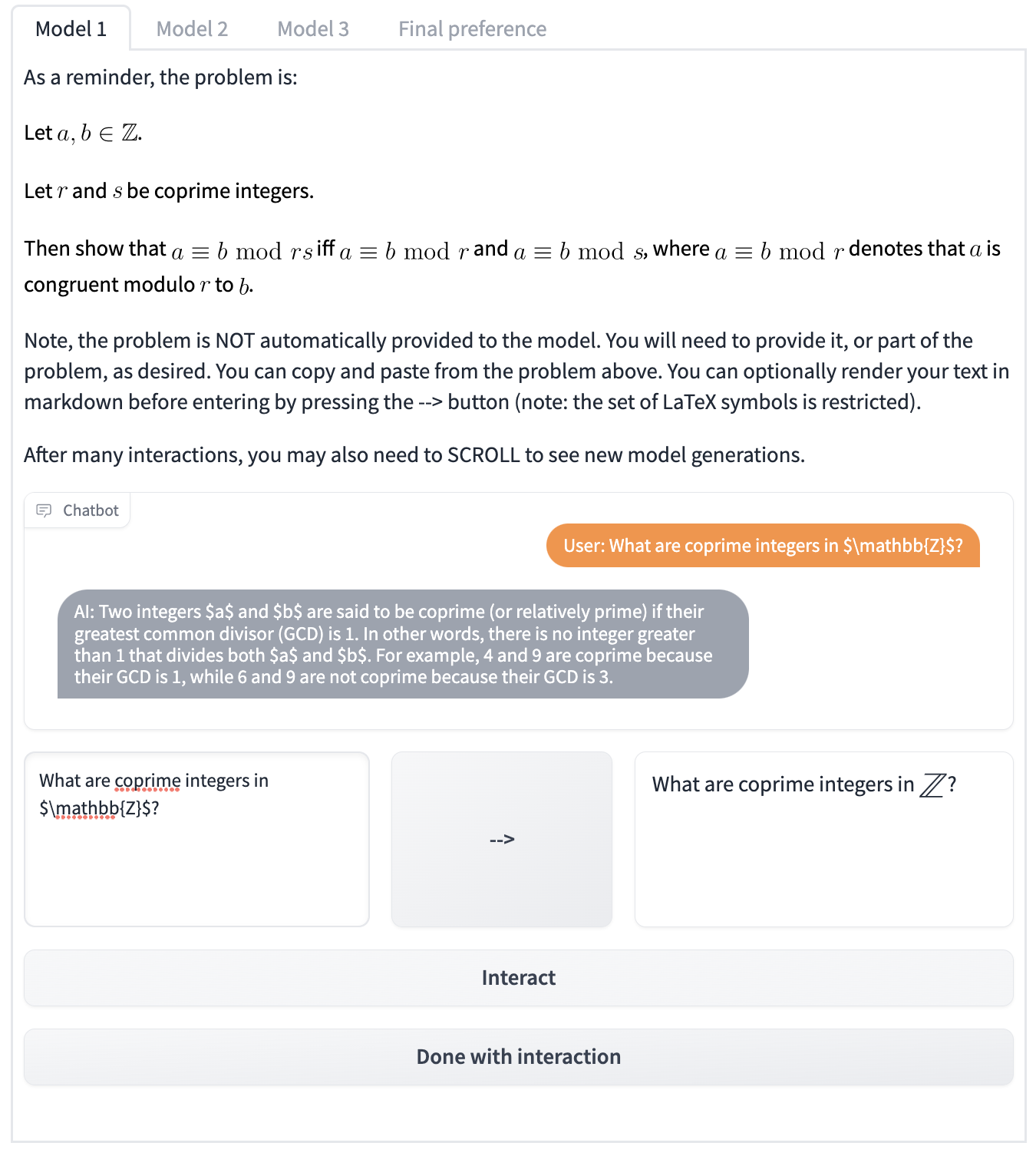}
\caption{Example chat interface to interact with an LLM. The participant is presented with the problem and a reminder of particularly relevant instructions, and enters their chat below (which they can see compiled in LaTeX). Chat history is presented for the user to refer to.}
\label{fig:interface1}
\end{figure}

\begin{figure}[!htb]
  \centering
  \includegraphics[width=.95\linewidth]{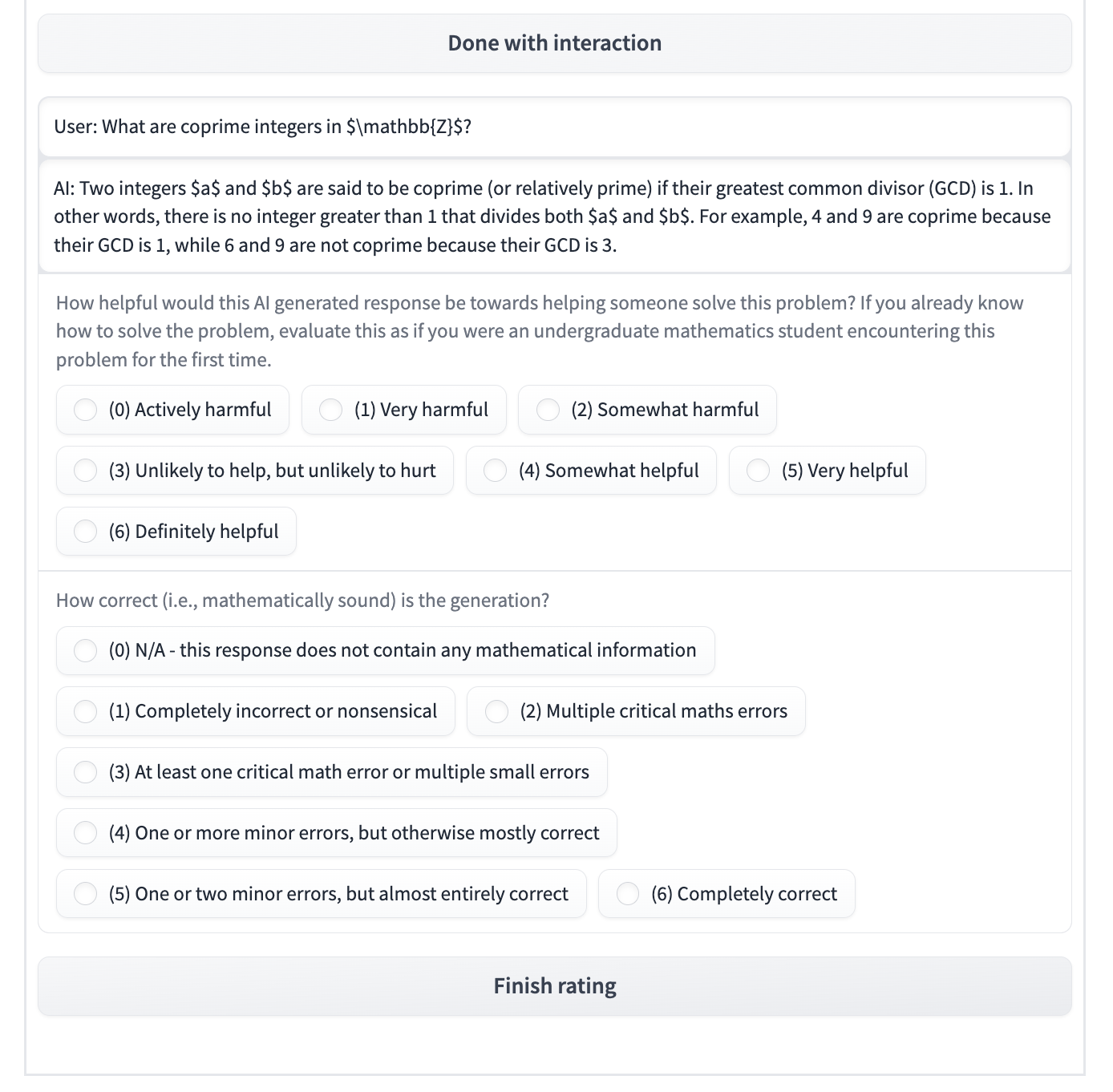}
\caption{Example page of post-interaction ratings of each generation, along multiple dimensions. The user is presented with each query-generation pair and rates according to mathematical correctness and perceived helpfulness.}
\label{fig:interface2}
\end{figure}

We also include example screenshots of our interface in Figures \ref{fig:interface1} and \ref{fig:interface2}. 

\section{Additional Details on Taxonomy Construction}

We include additional details on our taxonomy annotation scheme. The 256 human queries were split into 5 batches of approximately 50 queries each. Four authors participated in the taxonomy annotation; three authors categorised approximated 50 queries each, and one author annotated the rest (i.e., 100...[TODO, check number KMC annotated]). Annotators received a spreadsheet containing the human query per interaction, as well as the original problem-of-interest, the model's response for that interaction step, as well as all prior interactions from between that participant and the AI system on that problem. As noted in the Methods, confusions were discussed amongst annotators. During the process, a new bucket was added -- ``Asking how to do a step of the proof'' -- after an annotator noted that some participants seemed to decompose the problem into a single step, rather than asking for help over the full proof. One annotator who had already categorised their 50 queries went back and reconsidered them with this new bucket in mind. 

We include the annotation guidelines provided to the authors who agreed to partake in the taxonomy annotation:

\texttt{Please classify the user query/interaction (in the ``user\_query'') column into one or more of the annotation buckets.}

\texttt{If a query falls under a bucket, please add a ``y'' for ``yes'' to that column. If you are unsure, but think it may, please mark ``m'' in that column (for maybe). By default, we assume that a query does not fall into a bucket. A query may fall into more than one bucket; though we expect that most fall into only one.}

\texttt{If you think that a query falls into an abstract category that is not represented, please write in the “Other” column.}

\texttt{If you need to look at the original problem statement, see the ``problem\_declaration'' column. The previous human-AI interactions, with the humans’ ratings for the AI outputs, are in ``previous\_interactions''.}

\texttt{We expect that annotating 50 user queries will take approximately 15 minutes. If it's taking significantly longer, please let us know! We appreciate any time you may have to annotate even a few of these!}

\texttt{Note: we will release the annotations later (without names of the annotator). If you aren't comfortable with this, please do not proceed.}
\section{Additional Survey Observations}

We include additional details and analyses into the data we collect and release in \texttt{MathConverse}. 

\subsection{Potential Misleading Generations under Human Uncertainty}
\label{misled}

We include additional examples wherein participants who were uncertain of their abilities to solve the problem on their own (see \textit{Main Text}) rated a generation as completely mathematically correct when in fact it was not. These examples further illuminate the need to heed caution when leveraging these models at present for assistance, particularly in the case where the user may not be able to validate correctness. Below each example, we include commentary from mathematics expert, William Hart, from our author team.


\textit{Commentary}: The model's statement about $g(x)$ having degree $n$ is not correct; it clearly has degree $n+1$. A degree $n$ polynomial cannot be identically zero, so the next statement is also not correct. I don't follow its argument that $f$ and $g$ should be identical polynomials.

\textit{Commentary}: It is obvious only that all the terms with $i > 1$ are divisible by $4$. Therefore the generation is incorrect unless $n$ is even.


\textit{Commentary:} No, the matrix is not necessarily its own transpose. That is incorrect. The rest is poorly explained.

\subsection{Ties in Rank-Order Preferences}
\label{ties}

As noted in the \textit{Main Text}, some participants ranked models with ties (which occurred in 5 of all 15 complete preference ratings). In Table \ref{tab:rating_ties}, we show the full ratings provided by participants in the presence of ties.

\begin{table}[htb!]
    \centering
\begin{tabular}{@{}lll@{}}
\toprule
InstructGPT & ChatGPT & GPT4 \\ \midrule
3 & 3 & 1 \\
3 & 1 & 1 \\
3 & 2 & 2 \\
1 & 1 & 1 \\
2 & 3 & 2 \\ \bottomrule
\end{tabular}
\caption{Instances where a participant rated models with a tie. Lower means more preferable for assistance.}
\label{tab:rating_ties}
\end{table}

\subsection{Relationship Between Web Occurrences and Ratings}
\begin{figure}[!htb]
    \centering
    \includegraphics[width=0.9\textwidth]{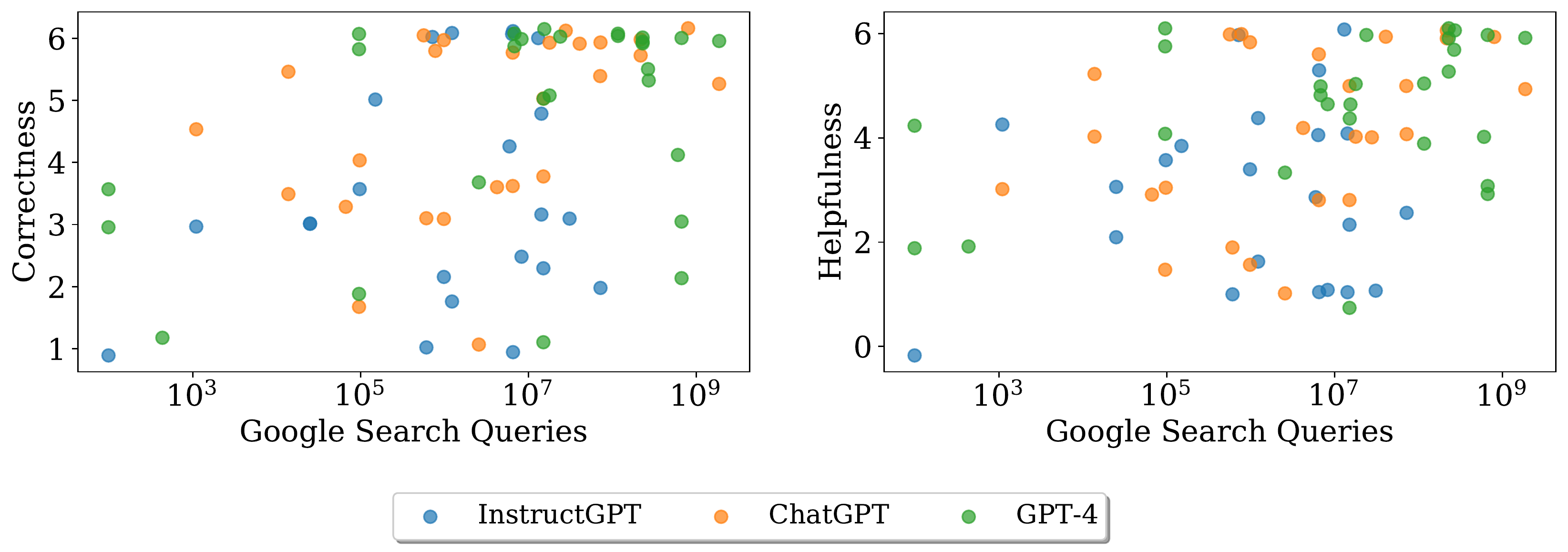}
    \caption{Correctness and perceived helpfulness ratings, broken down by associated model, against the number of google search queries associated with the original problem. One point per problem, depicting the average rating for that problem. Google queries are conducted over the ProofWiki theorem name associated with the problem.}
    \label{fig:google_freq}
\end{figure}

While investigating the data, we wondered whether the models' performance, as measured via the correctness and helpfulness ratings, may be related to the frequency by which the base ProofWiki problem appeared in the training set. While we do not have a transparent measure of dataset prevalence, given these models are largely trained on text from the web~\citep{Radford2019LanguageMA, brown2020language}, we used the number of Google search queries associated with the problem name\footnote{These names can be found in the released data.} as a proxy (see Figure \ref{fig:google_freq}). Intriguingly, we do not observe a reliable correlation between search term and average procured correctness and helpfulness (i.e., the Pearson correlation is not statistically significant for any model-rating pairing, and never more than 0.2). However, we encourage future work to further explore the relationship between prevalence of related problems in a models' training datasets and the models' provided assistance quality.

\subsection{Additional Investigations into the \texttt{MathConverse} Annotated Taxonomy}

\begin{figure}[t!]
    \centering
    \begin{subfigure}{\linewidth}
        \centering
        \includegraphics[width=\linewidth]{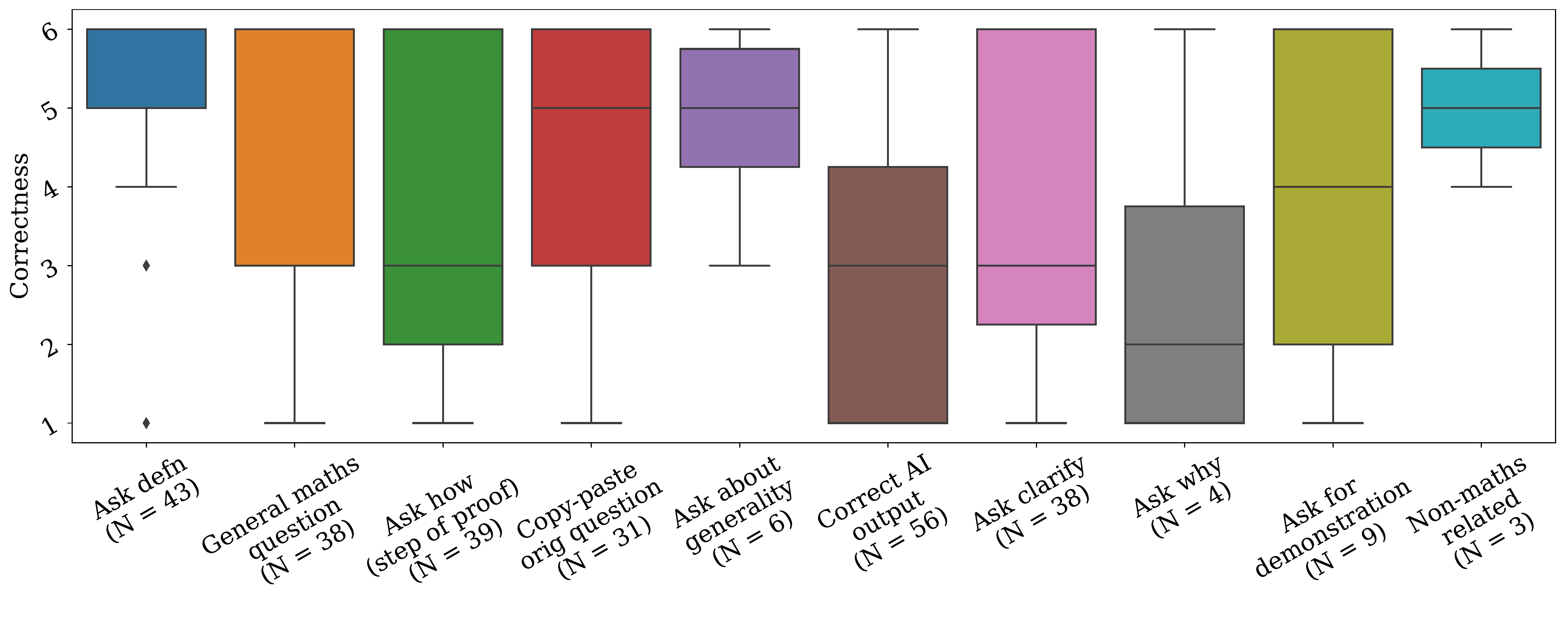}
        \caption{}
        \label{fig:query_type_correctness}
        \vspace{0.1in}
    \end{subfigure}
    
    \begin{subfigure}{\linewidth}
        \centering
        \includegraphics[width=\linewidth]{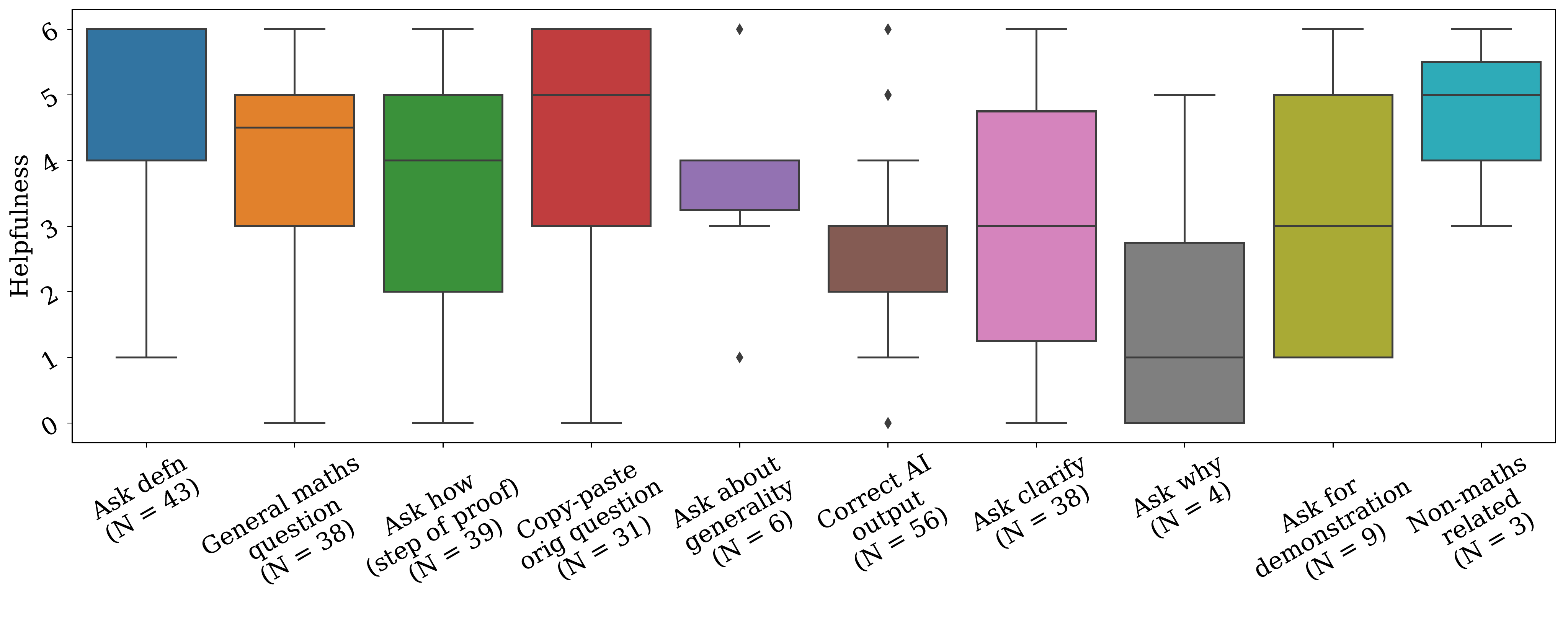}
        \caption{}
        \label{fig:query_type_helpfulness}
    \end{subfigure}
\caption{Rating of model generations along correctness (a) and helpfulness (b), decomposed by the kind of query the user had asked about. Query type is determined from our taxonomy annotations. N specifies the number of interactions that were classified along a particular query, over the full interaction trace.}
\label{fig:query_type_rating}
\end{figure}

We include additional analyses into our annotated taxonomy; in particular, we consider whether users' perceived helpfulness and correctness of the models' responses varies depending on the underlying query type. Such a question is important for both designers of AI systems -- e.g., to tailor model updates to make the system more usable on queries a user may desire -- as well as AI users themselves, who can better learn what kinds of queries the model is best equipped, at present, to provide useful responses for. Indeed, we find in Figure \ref{fig:query_type_rating} that the models' quality of responses do vary as a function of query type. For instance, the models are good at providing definitions (as we noted in the main text), and struggle to uptake corrections. 

\subsection{Handling Errors in Task Specification}

Additionally, after launching the survey, we were informed that two problem statements were incorrect: some assumptions or constraints that are crucial for the correctness of the theorems are missing or too strict. The first incorrectly stated theorem is Hurwitz's. Our description of it is as follows:

\fbox{%
    \parbox{\textwidth}{
    \textbf{Hurwitz's theorem*}
Let $\xi$ be an irrational number.
Then show there are infinitely many relatively prime integers $p, q \in \mathbb{N}_{>0}$ such that:

\[\left| {\xi - \dfrac{p}{q}}\right| < \dfrac {1}{\sqrt{5} q^2}\]
    }
}

The relatively prime integers $p$ and $q$ should be allowed to be any integers instead of positive integers. This is due to a copy error when we reformatted the original ProofWiki statement\footnote{\url{https://proofwiki.org/wiki/Hurwitz\%27s_Theorem_(Number_Theory)}}.

The second such statement is Tamref's Last Theorem:

\fbox{%
    \parbox{\textwidth}{
    \textbf{Tamref's Last Theorem*}
The Diophantine Equation:
\[n^x + n^y = n^z\]
has exactly one form of solutions in integers:

\[2^x + 2^x = 2^{x + 1}\]
for all $x \in \mathbb{Z}$.
    }
}
The theorem was not fully correct as stated on ProofWiki\footnote{\url{https://proofwiki.org/wiki/Tamref\%27s_Last_Theorem}} the day we accessed the content~(2 April 2023), as it did not ask for \textit{non-trivial} solutions. The solution $n=0$ trivially satisfies the Diophantine Equation.

We opted to leave these problems in the survey to act as ``canaries'' -- permitting us to detect whether participants (or models) were able to identify that problems were incorrectly stated in the written form. We find that since these two theorems are relatively well-known or easily understood, our experienced mathematician participants tended to use the definitions they had remembered to override the conditions given in the statements: the incorrect or missing assumptions are amended or added mentally by the users. Hence we find no significant difference in the user behaviours when they deal with these two problems. Sometimes, human users simply state the names of theorems or express them in natural language~(e.g., saying ``Weak Law of Large Numbers'' instead of its full statement), thus ending up not propagating the errors to the LLMs. 

For Tamref's Last Theorem, only GPT-4 made reasonable attempts in our collected responses. When prompted with the incorrect statements, GPT-4 seemed to be aware of the subtlety due to the lack of a non-triviality assumption: it either pointed out that there is a solution at $n=0$ or explicitly said that it was looking for non-trivial solutions. For Hurwitz's Theorem, all LLMs follow the incorrect assumption of the positivity of $p$ and $q$ and fail to come up with good solutions, if they were given the incorrect problem statement.

\subsection{Interaction Ratings by Mathematics Topic}

Recall, participants selected a mathematics topic at the start of the survey and interacted with the models on problems from that topic. We decompose the performance of models by participant topic selection in Tables \ref{tab:instructgpt_topic}, \ref{tab:chatgpt_topic}, and \ref{tab:gpt4_topic}. We depict average correctness and perceived helpfulness ratings for interactions on problems in each topic. Due to limited number of problems seen within each topic, we cannot draw definitive conclusions about differential model performance across these topics; however, we include for completeness.

\begin{table}[!ht]
    \centering
        \caption{InstructGPT scores decomposed by topic.}
\begin{tabular}{llll}
\toprule
 Topic     & \# Problems               & Correctness   & Helpfulness   \\
\midrule
 Algebra & 2            & 5.0$\pm$1.41  & 4.0$\pm$0.0   \\
 Group Theory & 2       & 1.77$\pm$0.97 & 1.85$\pm$0.86 \\
 Linear Algebra & 7     & 3.11$\pm$1.85 & 2.78$\pm$1.71 \\
 Number Theory & 5      & 3.16$\pm$1.9  & 2.21$\pm$2.21 \\
 Probability Theory & 5 & 3.61$\pm$1.83 & 2.94$\pm$1.47 \\
 Topology & 3           & 4.5$\pm$1.41  & 3.75$\pm$1.09 \\
\bottomrule
\end{tabular}

    \label{tab:instructgpt_topic}
\end{table}

\begin{table}[!ht]
    \centering
    \caption{ChatGPT scores decomposed by topic.}
\begin{tabular}{llll}
\toprule
 Topic     & \# Problems            & Correctness   & Helpfulness   \\
\midrule
 Algebra & 4            & 4.56$\pm$1.57 & 3.56$\pm$0.96 \\
 Group Theory & 2       & 4.0$\pm$1.94  & 3.56$\pm$1.26 \\
 Linear Algebra & 6    & 5.29$\pm$1.56 & 5.24$\pm$1.66 \\
 Number Theory & 6      & 3.55$\pm$1.86 & 3.25$\pm$1.81 \\
 Probability Theory & 5 & 4.2$\pm$1.6   & 3.44$\pm$1.77 \\
 Topology &  4          & 4.73$\pm$1.05 & 3.91$\pm$1.24 \\
\bottomrule
\end{tabular}
    
    \label{tab:chatgpt_topic}
\end{table}

\begin{table}[!ht]
    \centering
    \caption{GPT-4 scores decomposed by topic.}
\begin{tabular}{llll}
\toprule
 Topic     & \# Problems               & Correctness   & Helpfulness   \\
\midrule
 Algebra & 3            & 3.71$\pm$1.83 & 3.14$\pm$1.88 \\
 Group Theory & 3       & 3.83$\pm$2.27 & 3.5$\pm$1.89  \\
 Linear Algebra & 7    & 5.65$\pm$0.97 & 5.47$\pm$0.85 \\
 Number Theory & 5      & 3.15$\pm$1.79 & 3.54$\pm$1.65 \\
 Probability Theory & 6  & 4.88$\pm$1.67 & 4.33$\pm$1.65 \\
 Topology & 3           & 5.62$\pm$0.48 & 4.88$\pm$0.33 \\
\bottomrule
\end{tabular}
    
    \label{tab:gpt4_topic}
\end{table}

\section{Post-Survey Testimonials from Participants}
\label{testimonials}

We include additional quotes sourced from participants after completing our survey. Participants were reached out to via connections from the authors. We first include the complete question asked in the testimonial, followed by responses received. It is worth noting that testimonials were curated approximately one month after the interaction; we would encourage future work to consider such a survey immediately after completion of a \texttt{CheckMate} interactive session.

\subsection{Why Stop Interacting?}

 \textit{Full question: ``Why did you stop interacting?''}

\begin{itemize}
    \item \texttt{``I usually stopped when I had a distinct impression that improvement wasn't likely past the current point. That could be very early on, for example after a very poor initial response that demonstrated a profound mathematical shortcoming, or it could be after a longer interaction that convinced me no further improvement of the model's understanding or ability to accept and correct for its mistakes was likely with further prompting''}
    \item \texttt{``In the first experiment, I just tried to be fair, and give each model three queries (also because I thought I could just solve it). In later experiments, simply when it didn’t seem productive anymore, or I felt there was nothing more to ask.''}
    \item \texttt{``I feel like I have already gained enough ideas of the model’s capability. My daily work with mathematical proofs involves more complex manipulations of mathematical objects and interactions with proof assistant, which is still beyond the powers of the existing GPT4 model.''}
    \item \texttt{``Usually the model either gave a good answer after a few prompts or didn’t seem to be able to give one. Sometimes it happened that I did not get any better explanation to additional questions at which point I stopped.''}
\end{itemize}

\subsection{Human Uncertainty in Model Correctness?}

\textit{Full question: ``Were there times where you didn’t know whether the model’s generation was correct? If so, how did you handle this?''}

\begin{itemize}
    \item \texttt{``Yes. I did look up some facts and proofs. The main reason was that if the study is to be scientific it should not rely on my incorrect assertions.''}
    \item \texttt{``Of course, whenever I asked about a definition, or about a theorem I didn’t know (I tried not to look much into external resources). First, I was also often too lazy to check the model’s lengthy response (when it was a proof, I could be punctual with definitions, there is no way to be sure). But even if I check the proof and find a bug there, it doesn’t mean that the theorem doesn’t hold… Once the model really confused me, when I was solving some problem about rational approximation (I don’t remember it exactly), the model suggested continued fractions which sounded reasonable but the standard error estimate for continued fractions was too weak. So I asked whether there exists a closer estimate, and the model provided me one, and just in the final review of the exchange, I figured out that it was false.''}
    \item \texttt{``Yes, my biggest concern with LLMs is the uncertainty of the correctness of the generated answers. One way is, of course, to use the generated answer as a hint and then search with more reliable sources (e.g., textbook and peer-reviewed papers). An alternative way is to keep asking further questions and the same question from a different angle (like having a conversation with a real person). With rounds of Q\&A, I can then gain a better sense of the original question.''}
    \item \texttt{``Most times it was clear, however, sometimes the generations became quite lengthy and therefore difficult to check (especially if there was a mistake at the beginning). I found it similar to grading student homework which is easy when everything is correct (or false), but more difficult when explanations seem to contain valid arguments while still not being entirely correct. Study design feedback: I would have liked to make full text remarks after interacting with a model to be able to judge this more appropriately.''}
\end{itemize}

\subsection{Correct, but Unhelpful?}

\textit{Full question: ``Were there instances where a model’s response was mathematically correct, but you found it unhelpful? If so, why was it unhelpful?''}

\begin{itemize}
    \item \texttt{``Not really, however some of the models state facts that they know on the given topic instead of answering the questions that were asked. This could be helpful in certain situations, but the models don’t seem to discern when this is. It can be unhelpful if the answer to the actual question then gets cut off.''}
    \item \texttt{``Not much but I remember that there was a problem about B-algebras (I think it was called that way), and proofwiki has a very atypical notion of a B-algebra. I didn’t know what a B-algebra is, so I asked the model, and he told me some definition of a B-algebra that was likely mostly correct but it didn’t fit the problem at all. Eventually, this was one of the cases where I just had to go to proofwiki to figure out what the problem means since neither I nor the model had an idea.''}
    \item \texttt{``Yes, there is. The model is really good at retrieving mathematical definitions/statements that I cannot remember exactly. However, the model is yet good at proposing non-trivial proof strategies given those definitions – they can give some sound but not very helpful proof suggestions.''}
    \item \texttt{``It would have been very helpful if answers had been presented more clearly (rendered formulae instead of latex code, more display formulae), but this is not due to the model. Sometimes I tried asking for a hint, which didn’t produce good output. Therefore, I asked for a step-by-step solution which was then more helpful, however, this takes away the opportunity (for a student work with such a model) to actually also think about the question by themselves. ''}
\end{itemize}

\subsection{Using LLM-Based Assistants in Practice?}

\textit{Full question: ``Are there circumstances in which you would use one of the models you interacted with to assist you in mathematics? (or would have used in a particular way earlier in education, if available)''}

\begin{itemize}
    \item \texttt{``I’m going to try and use GPT-4, however I believe I will have to be very creative to do so. There’s no sense in which one can just ask GPT-4 questions and expect reliable, carefully thought out or ingenious responses. For me it is an open question whether it can be genuinely helpful. I’ve already wasted a lot of time chasing up things GPT3/4 hallucinated.''}
    \item \texttt{``With the high uncertainty about when the answer is true, I don’t think it is usable in practice yet. By the way, I just tried another free language model (HuggingChat) about whether there is a reference to a particular theorem which I thought should be known. It gave me three references, and none of them was correct… I don’t expect the models from the experiment to behave much better.\footnote{This response was provided to us on May 19, 2023. It is feasible that HuggingChat has or will improve since then.}''}
    \item \texttt{``I would definitely use them for initial pilot study (e.g., literature review, identifying potential proof routes and related well-known theorems). Nevertheless, given the unreliable nature of LLMs, I may always double-check the answers. Another big usage of LLMs is to polish my writings (even in mathematical writings), which is super helpful for a non-native speaker.''}
    \item \texttt{``I'd consult such models mainly to check for preexisting, relatively well-known results that I might not know about as a complement to classical literature search. And as a tool to improve my writing of non-technical portions of mathematical texts. The current models are definitely to poor in logical reasoning to be employed for finding, checking or rewriting proofs. I’d be quite afraid to develop incorrect understandings by using a model which can present wrong arguments in a very convincing way.''}
\end{itemize}

\section{Interactive Case Studies with Experts}
\label{case studies}

While structured interactive evaluation permits nice quantitative findings, to deeply understand the capability of LLMs -- in the context of mathematics and beyond -- free-form interaction is paramount. As discussed in \citep{burnell2023}, instance-level evaluation can be particularly revealing. In this section then, we offer a second form of interactive evaluation -- working directly with domain expert mathematicians (from our author list) to provide a series of case studies. The first case studies delve deeper into some of the same ProofWiki problems we evaluated with \texttt{CheckMate}. This is followed by a study that expands the scope of the evaluation, attempting to locate the boundary between problems that GPT-4 finds easy and those it finds hard. In our survey, we observed a close relationship between mathematical correctness and perceived usefulness; while correlation is not causation, we further explore the broader \textit{mathematical reasoning capabilities} of these models as a bedrock to inform their utility as proof assistants.

We reiterate that we are not aiming to single out GPT-4 for criticism. Rather, our goal is 1) to offer one of the first real expert mathematician interactive case studies with LLMs to help guide the design of better mathematical assistants and inform their safe, trustworthy use in the present by helping characterise their limitations, 2) to pave the way for further interactive evaluations (of which we still have too few), and 3) to highlight patterns of human-computer interaction not previously known to the community, particularly when the humans interacting are domain-leading experts. We hope the work will be of interest to ML engineers and researchers, cognitive scientists, human-computer interaction specialists, mathematicians, educators, and beyond. 

A complete transcript of interactions for each case study example is included in the Appendix. We maintain the original text of each case study author for authenticity, with only minor edits for precision and coherence.

\paragraph{A Deep Dive into ProofWiki}

First, our recruited experts conduct a deeper dive into some of the problems we explored in our previous evaluation. Specifically, we use the problems as a playground to explore how much the model seems to ``know'' about relevant concepts and further characterise what interactions can yield better (or worse) performance and assistance experience. We focus on GPT-4~(in chat mode) because it has the strongest overall performance in the study above. The experts chose to refer to GPT-4 ``\textit{the assistant}'' in the rest of this section.

The first case studies are provided by Dr. William Hart, a number theorist by training; the second were primarily contributed by Dr. Wenda Li, a formal mathematics expert\footnote{Case studies were conducted between April and early May 2023.}.

\subsubsection*{Number Theory Evaluation}
\label{sec: num thy}

\textit{Contributed by William Hart}

We provide an in-depth analysis of a number of GPT-4 responses to number theoretical questions. Number theory is an area of mathematics where problems are often simply stated, but difficult to solve, involving arbitrarily deep mathematics in their solution. Whilst we didn't interact with the model to work on any problem requiring very deep methods for its solution, such as Fermat's Last Theorem famously proved by Andrew Wiles, we did have a chance to observe the model as it struggled with problems ranging from trivial to moderately challenging.

\vspace{0.1in}
\textbf{General remarks}
Whilst GPT-4 is able to regurgitate some very commonly found elementary number theoretical material and can handle straightforward problems, it has a major difficulty with algebraic manipulation and little or no ability to work on unseen problems that require backtracking, proving intermediate lemmas or extensive planning.

\vspace{0.1in}
\fbox{
    \parbox{.85\linewidth}{
    \textbf{ProofWiki problem 21}
    Show that the equation
\[1 + a^n = 2^m\]
has no solutions in the integers for $n, m > 1$.
    }
}
\vspace{0.1in}

Problem 21 is a simple Diophantine equation. The problem is quite obscure~(a verbatim Google search gives 10 results) and thus not likely to appear in
training material repeatedly. The model took very reasonable steps towards
solving the problem: it started by claiming the proof is by contradiction and proceeded to reason about the assumed solution for $n, m > 1$.

It begins well by reasoning that $a$ must be odd because $1 + a^n$ is even. No explanation is given for this, but an experienced human
wouldn't explain this step either given the routine nature of parity arguments in Number Theory.

The next step is to take an expression $(2k + 1)^n$ which has appeared and expand it using the binomial theorem. However, it does this in
a surprising way, splitting the resulting sum into the first two terms and then a sum for the remaining terms.

$$(2k + 1)^n = \sum_{i=0}^n \binom{n}{i} (2k)^i = 1 + n(2k) + \sum_{i=2}^n \binom{n}{i} (2k)^i$$

This is impressive because GPT-4 is exhibiting some planning. It clearly has in mind to work modulo $4$ and it can see that all of the terms of the final
sum might vanish modulo 4. Indeed this is the very next claim that it makes.

Whilst it didn't explain why every term of the final sum is divisible by 4 it was asked on subsequent generations to explain this step and it correctly did so.

However, things do not go so well from here. It now claims that we can write the original equation $1 + a^n = 2^m$ as 

$$1 + 2kn + 4s = 2^m$$

for some $s$. This is a beguiling step that a human might overlook as correct, but it is not. The expression $1 + 2kn + 4s$ is the expression for $a^n$ not $1 + a^n$.
GPT-4 has made an algebraic error. This sort of thing is unfortunately very common and lets GPT-4 down on many examples.

Asking GPT-4 to self correct did not help it notice and correct its mistake. To see if it could eventually produce a completely correct proof, it was asked numerous times to solve the problem. Whilst its overall strategy was good on each generation, different algebraic mistakes occurred each time so that a correct proof was not eventually reached.

\vspace{0.1in}
\fbox{%
\centering
    \parbox{.9\linewidth}{
    \textbf{ProofWiki problem 28}
    Show that \[3 = \sqrt {1 + 2 \sqrt {1 + 3 \sqrt { 1 + \cdots} } }.\]
    }
}
\vspace{0.1in}

Problem 28 is a more difficult problem and the model is completely unable to deal with it. It admits that problems involving nested radicals can be difficult and actually gives up after standard methods don't make any headway.

A consistent problem here is an inability to write down a correct expression for a recursive relation to describe the nested radical. GPT-4 seems to be convinced that the expression under each square root is the same, so that if we write the initial expression $3 = \sqrt{A}$ then we also have $3 = \sqrt{1 + 2\sqrt{A}}$ and $3 = \sqrt{1 + 2\sqrt{1 + 3\sqrt{A}}}$, etc.

On subsequent attempts additional terms of the initial sequence were provided in the hope that it would pick up on the increasing sequence of constants that the square roots are multiplied by.

Whilst GPT-4 would confirm that it had noticed this pattern, it would always proceed ignoring this fact. On each generation, GPT-4 would finish off by noting it got the wrong answer and that this must be because it didn't take this increasing sequence of constants into account! It's as though GPT-4 only knows one way to handle nested radicals, and knowing that this won't work here, tries it anyway, inevitably getting the wrong answer. 

To probe a little deeper, GPT-4 was instead prompted in a direction that might allow it to make partial progress. The hint was given to try peeling the expression on the right hand side one square root at a time, working backwards from the desired result that the full nested radical should have the value 3 to see if some pattern could be found in the values of the inner nested radicals.

It was easy to prompt it so that it heads in that direction but on every generation it made hopeless algebraic and numerical errors, once again illustrating that very often what holds it back is high school algebra rather than the depth of the mathematics.

As GPT-4 could not be coaxed into returning correct values for the sequence of inner nested radicals, the attempt to solve the problem using GPT-4 was abandoned.

\vspace{0.1in}
\fbox{%
\centering
    \parbox{.9\linewidth}{
    \textbf{ProofWiki problem 24}
Let $\xi$ be an irrational number.
Then show there are infinitely many relatively prime integers $p, q \in \mathbb{N}_{>0}$ such that:

\[\left| {\xi - \dfrac{p}{q}}\right| < \dfrac {1}{\sqrt{5} q^2}\]
    }
}

Finally, Problem 24 is another difficult problem. Its solution on the
ProofWiki website requires a number of lemmas and some subtle reasoning.
Solving a problem of this kind would require some planning capability,
or at the very least the ability to backtrack and experiment with various
ideas. This is something that GPT-4 doesn't appear to possess beyond what
can be `computed' within the model itself.

GPT-4 does make the completely reasonable first step of approaching this problem using a continued fraction expansion of the irrational number $\xi$. Many approximation problems of this kind do indeed proceed this way. Continued fractions yield a sequence of convergents $p_n/q_n$ that converge to the irrational
number $\xi$.

After picking a reasonable theorem from the theory of continued fractions and applying it, GPT-4 has the following expression

$$q_n q_{n+1} > \sqrt{5} q_n^2.$$

At this point it is clear that GPT-4 does not know how to proceed, but knows what it should end up with, so makes the unsubstantiated claim that this inequality is satisfied when $q_{n+1} > \sqrt{5} q_n$.

There is no reason to infer that this should be the case at this point in the problem and if the particular chosen approach is to work out, this would have to be proved. Instead of doing so, GPT-4 just asserts that it is true without attempting to prove it.

When asked directly how to prove this statement GPT-4 clearly has no idea how to do so and makes a completely bogus claim that a sequence with linear growth will eventually outgrow a sequence with exponential growth. It seems to be common for GPT-4 to hallucinate details when things aren't working out or if it doesn't know a reasonable answer.

In other contexts we have observed that GPT-4 can produce better output if asked to stop itself if a
particular mathematical approach does not seem to be working out and to try another
approach. When prompted to do so in this particular case GPT-4 did indeed try numerous reasonable strategies but
unfortunately it was still not ultimately successful. This was in part
due to poor choices along the way and partially due to being plagued by
algebraic errors which ended up misleading it.

For balance we mention that the failed attempts above were not entirely characteristic of GPT-4 which can in some cases produce perfect answers.

For example, Problem 27 is solved completely, although slightly inefficiently and Problem 23 is correct except for a single bogus explanation which was not particularly significant. However, it should be pointed out that Problem 27 is quite trivial, essentially requiring only the binomial theorem and Problem 23 is completely standard in many texts on elementary Number Theory. It is very unlikely that the standard proof would be hit upon at random, and the fact that GPT-4 generates it perfectly is probably an indication of the relative abundance of proofs in training material.

\subsection{Problem Perturbation to Probe Memorisation}
\label{sec: wenda case studies}

\textit{Contributed by Wenda Li}

If a system simply memorizes the answers to problems, its performance can greatly differ depending on whether the problems it is evaluated on are memorized. In this section, we evaluate GPT4's performance on variations of problems from ProofWiki, which are far less likely to be appear in training data since we make novel variations. Concretely, we varied the problems in three different ways to test the model's understanding of the problems: asking for definitions of concepts, loosening assumptions of problems, and instantiating abstract variables with values. Two problem instances and their variations are presented due to them being the most interesting and revealing examples from all that were tried.

\vspace{0.1in}
\textbf{General remarks} 
We found GPT4's performance at variations of several ProofWiki problems quite satisfactory: it can reliably retrieve definitions of concepts used in the problem as well as in its own proof; it can correctly assess whether loosening certain assumptions breaks the proof; it can also instantiate variables quite robustly, given the opportunity of inspection of its own answers.
There have been debates~\citep{stochasticParrots, piantadosi2022meaning} about to what extent shall we say language models ``understand'', given the nature of their stochastic generation. In our study, we find a couple of simple\footnote{Note, ``simple'' here is relative to that of a trained mathematician.} cases where the language-model-based assistant possesses the mathematical understanding of assumptions and variable instantiations beyond mere memorisation.

\vspace{0.1in}
\fbox{
    \parbox{.9\linewidth}{
    \textbf{ProofWiki problem 25}
    Let $a, b \in \mathbb{N}_{>0}$ such that there exists no $m, n \in \mathbb{N}_{>0}$ such that $a^m = b^n$. Prove that $\log_b a$ is irrational.
    }
}
\vspace{0.1in}

Presented the problem statement above, the assistant gave a perfect answer with step-by-step calculations.
To test if the assistant has a true mathematical understanding of the problem, we first asked for definitions of concepts used, and then varied the original problem by loosening some of the assumptions made, and asked the assistant for a proof in the new setting.

\textbf{Asking for definitions}
We found that the assistant gave the correct definitions in the theorem statement as well as in its own proof. Concretely, it gave the right answers for: the definition of logarithm; the range of a logarithm's base; the meaning of the set $\mathbb{N}_{>0}$; and whether $\log_ba=\frac{p}{q}$ can be a negative number~($p$ and $q$ are variables arising from the assistant's own proof).

\textbf{Loosening assumptions}
We started by asking the assistant \textit{whether the proof still holds if we instead have $a, b \in \mathbb{R}_{>0}$?}
The assistant understood the meaning of $\mathbb{R}_{>0}$ and confirmed the derivation still held, so the original lemma/proposition has been generalised (since one of its assumption has been relaxed). Later, we attempted to generalise the proposition further by dropping the assumption $a \in \mathbb{R}_{>0}$ or $b \in \mathbb{R}_{>0}$:

We continued by asking \textit{if dropping the assumption that $b \in \mathbb{R}_{>0}$ or $a \in \mathbb{R}_{>0}$ affects the original proof?}
The assistant knew that these assumptions were necessary to make the log function well-defined, and pointed out that dropping either of the assumptions would invalidate our previous derivation.

These variations, though not impossible, are unlikely to appear together with the problem in the training data of the assistant. We think the assistant does have some understanding of the underlying mathematical concepts and its own proof, in the context of this problem.

\vspace{0.1in}
\noindent\fbox{%
    \parbox{.95\linewidth}{
    \textbf{ProofWiki problem 39}
    Let $X$ be a random variable. Assume $\mathsf{E} (X) = \mu$ for some $\mu \in \mathbb{R}$ and $\mathsf{var} (X) = \sigma^2$ for some $\sigma^2 \in \mathbb{R}_{> 0}$. Show that for all $k > 0$: $\Pr \left({\left| {X - \mu}\right| \geq k \sigma}\right) \leq \dfrac {1}{k^2}.$
    }
}
\vspace{0.1in}

Given this problem statement, the assistant mentioned that we can use the Chebyshev's inequality, and then re-stated the problem in an  almost identical way but with different variable names. This demonstrates a certain level of variable unification, which is an important concept in automatic theorem proving.

\textbf{Variable instantiation}
We then checked whether the assistant knew how to instantiate variables by asking it whether the proof still holds when the following concrete values are assigned to $k$: $2$, $\sqrt{2}, \sqrt{2}-1, \sqrt{2}-2$, and $(\sqrt{2}-2)^2$. Human inspection finds the assistant's behaviour to be correct.
The assistant can clearly handle concrete calculations even when $k$ is a relatively complicated number (e.g., $\sqrt{2} - 1$). The model also knows that the previous derivation cannot be carried out when $k=\sqrt{2} - 2$, a negative number. 

An interesting observation arose when the assistant was not confident of its derivations: we asked: ``\textit{are you sure $(\sqrt{2} - 2)^2 > 0$?}''
The answer should be affirmative, but the assistant started to apologise and revise its previous correct calculation by saying ``\textit{When $k = (\sqrt{2} - 2)^2$, the value of $k$ is indeed non-negative, but it is actually equal to 0, not greater than 0.}'' When we asked again ``\textit{Why do you say your previous statement was incorrect and $k=0$? I don't understand.}'', the assistant corrected its previous mistake and produced the right evaluation.

We found that the assistant is generally quite capable with variable instantiations and evaluating certain complex expressions, with the occasional mistake made with low confidence. We hypothesise that the mistake may be a defect of its reinforcement learning with human feedback (RLHF)\citep{christiano2017deep, ouyang2022training, casper2023open} training: the human feedback is mostly assumed to be right, and when the feedback questions a true fact, the assistant concurs and alters its own (correct) response.

\subsection{Investigating the Boundary between Easy and Hard Problems}
\label{tim case studies}

\textit{Contributed by Timothy Gowers}

If we want to understand how and to what extent large language models can help mathematicians, it is clearly important to understand what they can and cannot do. A range of views have been expressed on this topic, with some saying that they already show glimmerings of AGI~\citep{bubeck2023sparks}, and others dismissing them as mere ``stochastic parrots" \citep{bender2021dangers}. On the latter view, the successes that LLMs undoubtedly have solving mathematical problems are to be understood as very good guesses of what the outward form of a proof would look like, unaccompanied by any genuine understanding of what that proof means, even when it is correct.

A difficulty with evaluating the level of understanding of a language model is its opacity: for any particular answer it gives, we do not know the extent to which it is merely an amalgam of texts it has seen as part of its training data and the extent to which it has actually had to build its answer in a coherent way. One way to try to get round this problem is to ask the LLM questions that are deliberately designed to be ``quirky" and non-standard. A good source of such questions is ones that ask whether mathematical objects of certain kinds exist with various artificial combinations of properties.

Timothy Gowers, a mathematician from our author team, tested GPT-4 on many such questions, attempting to identify a ``boundary" between what it could and could not do. This imposed a further constraint on the questions: that they should ideally be modifiable in various ways, so that one can ``turn the dial" until GPT-4's performance drops.

\textbf{General remarks}
Many of the strengths and weaknesses we observed in GPT-4 are ones that have been commented on several times (in connection with other LLMs as well). For instance, it is not good at calculation, it has a tendency to ignore facts that do not support its main conclusion (even if it itself has generated those facts), and to invent facts that do support it~\citep{azamfirei2023large}. 

When it comes to building examples, it has another weakness, which is that instead of using a process of reasoning to constrain what the example can look like and only then exhibiting the example, it prefers to start by exhibiting the example and then provide the justification that it has the desired properties. If its initial suggestion is correct, then this may be all right (though its justifications are not always correct), but often the example it gives is not correct, and it typically follows it up with a ``justification'' that to a trained mathematician has very obvious flaws. This behaviour supports the stochastic parrot view, since mathematical constructions are typically presented in the literature in the same unhelpful style -- first the example, and then the verification that it works.

One can try to use prompt engineering to discourage GPT-4 from organizing its answers in this way, for example asking the model not to provide an example immediately but to describe a general strategy first, and only then the implementation of the strategy, or to pretend to be a student and ask it to offer hints. While these did improve its output somewhat, they did not make a significant enough difference to affect our general conclusions, so we mainly used more straightforward prompts. 

A further feature that has a negative effect on the experience of interacting with GPT-4 is that even when it gives correct answers, it often describes them and checks their properties in a laborious case-by-case way, and often those checks are not even necessary. For instance, if asked to construct a 3-regular graph with eight vertices, it will typically list all the vertices and edges, and then for each vertex it will say what its neighbours are and comment that there are three of them. Or if it needs the fact that the matrix $\begin{pmatrix}0&1\\0&0\\ \end{pmatrix}$ squares to the zero matrix, instead of just asserting that it does, it will write
\[\begin{pmatrix}0&1\\0&0\\ \end{pmatrix}^2=\begin{pmatrix}0&1\\0&0\\ \end{pmatrix}\begin{pmatrix}0&1\\0&0\\ \end{pmatrix}=\begin{pmatrix}0&0\\0&0\\ \end{pmatrix},\]
which is adding virtually nothing to a bald assertion, since no details are given of the computation of the matrix product (not that one would want them). Similarly, it sometimes ``verifies'' that a matrix is symmetric by writing out that matrix and its transpose and noting that the two matrices it has written out are identical.

In the more positive direction, there were several questions that GPT-4 answered well in a way that is hard to dismiss as mere parroting. To give a simple example, if it is asked to differentiate a function, it will typically do so very competently, suggesting at least some ability to generalize. (A possible explanation for why it is good at differentiation and bad at arithmetic is that there are many more worked examples of differentiation, complete with step-by-step explanations, than there are worked examples of arithmetic.) More examples will be discussed below, with some speculations about why GPT-4 is good at them.

\paragraph{Example 1}
\label{case studies ex1}

Here is a prompt given to GPT-4.

\begin{quote}
Can you find an example of a set A of positive integers that satisfies the following three conditions? (The LaTeX is compiled here but was uncompiled in the prompt.)

1. $|A| = \min(A)$.

2. $A$ consists of prime numbers only.

3. $A$ has cardinality at least 5.
\end{quote}

The response was the following correct and concise answer.

\begin{quote}
Yes, here's an example of a set A of positive integers that satisfies the given conditions:

A = {5, 11, 17, 23, 29}

The cardinality of A is 5, which is also the minimum element in A.

1. All elements in A are prime numbers.

2. The cardinality of A is 5, which is at least 5.

3. Therefore, set A = {5, 11, 17, 23, 29} meets all three conditions.
\end{quote}

Despite the fact that this answer is correct, it has a revealing feature. Why, one might ask, does GPT-4 not give the more obvious answer $\{5,7,11,13,17\}$? That is, why does it not take the five smallest prime numbers greater than or equal to 5?

A clue lies in the fact that the set it does choose, $\{5,11,17,23,29\}$ has the additional interesting property that it forms an arithmetic progression. This property has led over the years to the set being mentioned in many articles about arithmetic progressions that consist of primes, and presumably a significant fraction of those also mention that the number of elements is 5.

What is illustrated here is a preference by GPT-4 to use off-the-shelf examples, a preference that is strong enough that it will often do so even when the examples do not in fact have the required properties. 

Why would most human mathematicians prefer to take the first five primes, starting at 5? It is because that example is easily generalisable: if we change the number 5 to any other number $n$, we simply have to choose a prime $p$ that is at least as big as $n$ and take the first $p$ prime numbers from $p$ onward. 

To understand GPT-4's behaviour better, it is therefore natural to give it the following slightly modified prompt.

\begin{quote}
    Can you find an example of a set A of positive integers that satisfies the following three conditions?

1. $|A| = \min(A)$.

2. $A$ consists of prime numbers only.

3. $A$ has cardinality at least 6.
\end{quote}

The response was

\begin{quote}
    Yes, here is an example of a set A of positive integers that satisfies the given conditions:

A = {7, 11, 13, 17, 19, 23}

1. |A| = 6, and the minimum element in A is 7, so |A| = min(A).

2. All elements in A are prime numbers: 7, 11, 13, 17, 19, and 23.
3. A has cardinality 6, which is at least 6.

The set A satisfies all three conditions.
\end{quote}

Even after its error was pointed out, and a few more rounds of the conversation had taken place, GPT-4 did not manage to eliminate all mistakes and give a correct answer. Indeed, when told that 6 does not equal 7, it at first tried to explain why no such set could exist. 

This example lends some weight to the stochastic-parrot view: the problem is extremely easy for a human, but it is also a sufficiently artificial question that GPT-4 will not be likely to have encountered it, or anything very like it, in the texts on which it has been trained. One might argue that it does at least give a set of consecutive primes this time, but that again is what one would expect, given that sets of consecutive primes will appear much more frequently in the literature than other sets (a tendency that one might guess is overridden when the sets start with 5 and have size 5). 

\paragraph{Example 2}
\label{case studies ex2}

This was a successful attempt to trick GPT-4. It was first asked to evaluate the integral $\int_{-1}^1x^3\cos x\,dx$. It did so by a laborious argument using iterated integration by parts, presumably because the integrand is typical of the integrands in integration-by-parts exercises. After getting everything right until close to the end, it made an elementary mistake, but with a bit of help it arrived at the correct answer, 0 (though still persisting in some of its incorrect calculations). On being asked whether there was a shorter argument, it gave the right response: that the integrand is an odd function and the integral symmetric about 0, so the integral is 0. 

Then it was asked to evaluate the integral $\int_{-1}^1(x^2-1/3)\, dx$. Again it did so correctly, obtaining the answer 0. On being asked whether there was a simple reason for this, it once again pointed out that the integrand was an odd function, and even supplied a bogus proof that the function is odd. 

This behaviour again fits the stochastic-parrot hypothesis quite well: almost always if one is asked for a simple reason that the integral of a function over a symmetric interval is zero, the correct response is that the function is odd. Whether or not it actually is odd is for GPT-4 a secondary consideration.

\paragraph{Example 3}
\label{case studies ex3}

A nice problem that mathematics undergraduates tend to find quite hard is to determine whether there is an order-preserving bijection between the rational numbers and the dyadic rationals. Surprisingly, the answer is yes, and the proof is by what is known as a back-and-forth argument. That is, one enumerates the rationals and the dyadic rationals, and then one alternates between choosing a match for the first unmatched rational and the first unmatched dyadic rational, making sure at each stage that the order is preserved. 

When GPT-4 was asked the problem, it tried to prove that no such bijection could exist. Each time its mistakes were pointed out to it, it replaced its bogus argument by a slightly modified bogus argument. 

Upon being asked whether it knew about back-and-forth arguments, it said that it did, and explained that they could be used to prove a theorem of Cantor, that any two countable dense subsets of the real numbers are order isomorphic. It did not seem to realise that the problem it had been asked was a special case of this theorem. It also tried to explain why one could not use a back-and-forth argument to prove that the rationals and the dyadic rationals are order isomorphic, but after its explanation was criticised, it proceed to give a convincing sketch of how such an argument would indeed work.

\paragraph{Example 4}
\label{case studies ex4}

When GPT-4 was asked whether there is a positive integer $n$ such that $n+k$ is divisible by $k$ for every integer $k$ in the range $\{1,2,\dots,100\}$, it said no, and offered bogus proofs. After being steered towards a positive answer via certain easier questions, it suggested taking $n$ to be $\text{LCM}(1,2,\dots,k)-1$. This was interesting because $\text{LCM}(1,2,\dots,k)$ would have been a correct answer, but the pointless subtraction of 1 ruined it. Furthermore, it gave an argument that would have been correct if the -1 had not been present.

What might explain this act of self sabotage? One idea is that GPT-4 is influenced by Euclid's proof that there are infinitely many primes, which assumes that $p_1,\dots,p_k$ are all the primes and considers the number $p_1p_2\dots p_k+1$. An alternative argument would be to consider the number $\text{LCM}(1,2,\dots,k)-1$, where $k$ is at least as large as the largest prime. 

However, this explanation is rather speculative, and a Google search does not seem to back it up. When GPT-4 was asked why it had subtracted 1, it did not provide a convincing reason either. 

More revealing was its behaviour when its example was criticised on the grounds that, for instance, $\text{LCM}(1,2,\dots,k)-1$ is odd, and therefore not divisible by 2. Instead of adjusting its answer, as a human mathematician might, it decided that no such $n$ existed, and when its arguments for that conclusion were criticised, it went back to the example of $\text{LCM}(1,2,\dots,k)-1$. Even when asked whether $\text{LCM}(1,2,\dots,k)$ would work better, it initially said no. So this was not really a ``boundary'' example, and more just a problem on which GPT-4 got into a rut and could not get out of it. 

\paragraph{Example 5}
\label{case studies ex5}

A better example of a ``boundary'' problem was the following question: does there exist a graph with eight vertices such that every vertex has degree 3? Once again GPT-4 demonstrated its liking for off-the-shelf examples, giving the example of the 3-dimensional discrete cube. (An alternative approach is to take eight vertices joined in a cycle, and to join each vertex in addition to the vertex opposite it in the cycle.) 

When asked whether there was a graph with eight vertices such that every vertex has degree 5, it performed far worse. It did not know of any off-the-shelf examples, and was probably incapable of tricks such as taking the complement of an 8-cycle (which works because in the cycle every vertex has two neighbours, so in the complement of the cycle it has five neighbours). That is, it does not appear to be capable of taking an off-the-shelf example and \emph{modifying} it in a suitable way. Instead, it resorted to listing the vertices as A, B, C, D, E, F, G and H, and for each vertex giving a list of its neighbours. The trouble is that this kind of approach gave it many opportunities to fail as a result of familiar weaknesses such as a propensity to make calculation errors or to write down inconsistent statements. For instance, over its several attempts it would frequently list a vertex $v$ as a neighbour of another vertex $w$, but without listing $w$ as a neighbour of $v$. Eventually, probably with a slice of luck, it came up with an example that turned out to be the complement of the disjoint union of a 3-cycle and a 5-cycle. (Since the complement has to be regular of degree 2, it will always be a disjoint union of cycles.) 

\paragraph{Example 6}
\label{case studies ex6}

It has been noted that GPT-4 likes well-known patterns, and that one way of inducing it to fail is to ask it questions that will tempt it to give answers that fit those patterns. The following attempt to lead it astray in that way was a partial success. It was asked to find integers $a$ and $b$ such that the sequence $(1,3,a,7,9,b,13,15)$ is strictly increasing but not an arithmetic progression. It responded by choosing $a=5$ and $b=11$, thereby falling headlong into the trap. However, it then did a check by calculating all the successive differences. On observing that it obtained the difference sequence $(2,2,2,2,2,2,2)$, it then modified its choice of $b$ to 12, after which it recalculated the difference sequence, obtaining $(2,2,2,2,3,1,2)$ and declaring itself satisfied. 

This was another example where despite arriving at the correct answer, GPT-4 argued in a very non-human way. The main non-human feature was of course that it began by making the one guess that it needed to avoid (out of the strictly increasing possibilities). However, the whole approach of guessing and then verifying is inappropriate for the problem, since it is much more efficient to reason as follows: first, we note that \emph{if} the sequence is to be an arithmetic progression, then it will have to have common difference 2 (since the first two terms differ by 2) so it is sufficient to ensure that $a\ne 5$. This kind of forward planning appears to be beyond the current capabilities of GPT-4, (though maybe it could be induced to some small extent with careful prompt engineering).

We briefly mention its response to a variant of the problem, where it was asked whether it was possible to find integers $a,b$ and $c$ such that the sequence $(1,a,b,c,14)$ is an arithmetic progression. It answered yes, then set $d$ to be the common difference, obtained the equation $14=1+4d$, solved for $d$, discovered that $d$ was not an integer, and answered no, having apparently forgotten that it had previously answered yes. This showed a reluctance to plan in advance even in a situation where it was entirely capable of carrying out the required planning. 

\paragraph{Example 7}
\label{case studies ex7}

GPT-4 was asked to find a colouring of the set $\{1,2,3,4,5,6,7,8,9\}$ using three colours and satisfying the conditions that each colour is used three times, and no integer $n$ has the same colour as $n+1$ or $n+3$. The obvious way to solve this problem is trial and error, which for a typical human will work with a small amount of backtracking. It did indeed choose this approach at first, but failed quite badly because it was unable to check the conditions properly, which caused it to assign colours that were forbidden by the conditions, and, in the other direction, to claim that certain choices were forced when they were not. After a couple of failed attempts, it switched to trying to find a ``systematic" approach. One such attempt was to split the set into even and odd numbers, but that did not help it find a correct colouring. It even tried splitting the numbers into the three sets $\{1,4,7\}, \{2,5,8\}$ and $\{3,6,9\}$ and assigning one colour to each set, which violated the $n+3$ condition many times over. 

In sum, its ability to check simple conditions was too unreliable for it to be able to push through a trial-and-error approach, and for this problem a guess-and-check approach has a very low chance of success. 

\paragraph{Example 8}
\label{case studies ex8}

A somewhat similar question on which it performed badly was to find a sequence of nine distinct integers with no increasing or decreasing subsequence of length 4. Here it once again showed its taste for patterns: the problem was that it did not choose appropriate patterns. An example that was typical of its output was $(4,1,5,2,6,3,7,8,9)$. Interestingly, when, after a few failures, it was given a reason-step-by-step prompt, it produced the same example, this time after talking about interleaving sequences -- an idea which, if used correctly, leads to solutions such as $(1,4,7,2,5,8,3,6,9)$. However, encouraging it to use interleaving just led to further incorrect guesswork, an extreme example of which was when it interleaved the sequences $(1,3,5)$ and $(2,4,6,7)$ to obtain $(1,2,3,4,5,6,7)$, stuck 8 and 9 on the end, and proposed $(1,2,3,4,5,6,7,8,9)$ as a solution (complete with a ``check'' that it worked). 

When given the hint that it might like to start its sequence with $(7,8,9)$, it immediately made the obvious suggestion $(7,8,9,4,5,6,1,2,3)$. When asked for a rigorous proof that this sequence has the desired property, it gave an inadequate answer, stating correctly that the longest increasing subsequences are those that begin with $7, 4$ and 1, and stating incorrectly that the longest decreasing subsequences start with 9, 6 and 3. There was nothing resembling the argument that a decreasing sequence can contain at most one term out of each of the subsequences $(7,8,9), (4,5,6)$ and $(1,2,3)$.

\paragraph{Example 9}
\label{case studies ex9}

The following is a question on which GPT-4's performance was more or less perfect. It was asked to evaluate the sum $\sum_{r=2}^n\log(1-1/r)$. It noticed that $1-1/r$ can be rewritten as $(r-1)/r$, which therefore leads to a telescoping sum and and eventual answer of $\log(1/n)$. It did not give a detailed explanation of why, as it put it, the denominator in each term cancelled with the numerator in the succeeding term, but when asked for one it gave a good one (though it argued for a telescoping product inside the logarithm rather than a telescoping sum, which might have been simpler).

This was interesting because it required putting together two tricks: one to rewrite $1-1/r$ and the second to use the telescoping. However, its performance on a follow-up question strongly suggested that it had seen the question in its training data. It was asked whether it could evaluate the sum $\sum_{m=1}^n\log(1/m(m+1))$. This question was intended to trick it, since the sum $\sum_{m=1}^n 1/m(m+1)$ is famously evaluated by rewriting $1/m(m+1)$ as $1/m-1/(m+1)$ in order to obtain a telescoping sum. However, instead of falling into the intended trap, it stated falsely that it could rewrite the sum as $\sum_{m=1}^n\log((m+1)/m)$, after which it obtained a telescoping sum very similar to that of the previous question. 

When asked a further follow-up question to test how well it understood the principle of telescoping sums, it gave an answer that had good points and bad points, both of which indicated that it was good at capturing the look of a correct answer while not operating at the level of the underlying concepts. The question was to determine the sum $\sum_{n=1}^Ng(n)$, when $g(n)$ had been defined to be $f(n)-2f(n+1)+f(n+2)$. It is instructive to look at its complete answer.

\begin{quote}
Let's work out the sum. We have

\[\sum_{n=1}^Ng(n)=\sum_{n=1}^N(f(n)-2f(n+1)+f(n+2))\]

We can break this into three separate sums:

\[\sum_{n=1}^Nf(n)-2\sum_{n=1}^Nf(n+1)+\sum_{n=1}^Nf(n+2)\]

Let's shift the indices in the second and third sums to make the limits match up with the first sum:

\begin{enumerate}
    \item For the second sum, we substitute $m = n + 1$, so $n = m - 1$. When $n = 1$, $m = 2$; when $n = N$, $m = N + 1$.

    \item For the third sum, we substitute $m = n + 2$, so $n = m - 2$. When $n = 1$, $m = 3$; when $n = N$, $m = N + 2$.
\end{enumerate}

Doing this, we get:

\[\sum_{n=1}^Nf(n)-2\sum_{m=2}^{N+1}f(m)+\sum_{m=3}^{N+2}f(m)\]

Now, we can write out the sums explicitly to see what's happening:

\begin{enumerate}
    \item The first sum is: $f(1) + f(2) + f(3) + \cdots + f(N)$.
    \item The second sum is: $2f(2) + 2f(3) + \cdots + 2f(N+1)$.
    \item The third sum is: $f(3) + f(4) + \cdots + f(N+2)$.
\end{enumerate}

By looking at these sums, we can see that most of the terms will cancel out. For example, the term $f(2)$ is added in the first sum and subtracted twice in the second sum. Similarly, the term $f(3)$ is added in the first sum, subtracted twice in the second sum, and added once in the third sum.

After all the cancellations, we are left with:

\[=f(1)-2f(2)+f(N)-2f(N+1)+f(N+2)\]

So, the sum $\sum_{n=1}^N g(n)$ simplifies to $f(1) - 2f(2) + f(N) - 2f(N+1) + f(N+2)$.
\end{quote}

A first comment is that the final answer here is not just wrong, but clearly wrong to an experienced mathematician, because there is a symmetry in the problem that guarantees that the number of leftover terms at the two ends will be the same, whereas in GPT-4's answer there are two terms at the beginning and three at the end. Given its performance on other problems, it is no surprise that GPT-4 does not spot this anomaly.

On a more positive note, the first two steps -- splitting up the sum into three parts and changing variables -- are exactly the right thing to do, even if the purpose of changing of variables is not to get the limits to match up (in fact, it does the opposite). However, it then abandons this promising beginning in favour of writing out the sums in a more transparent notation, a move that is striking because it is what many human mathematicians would do if they did not immediately see how to proceed (but presumably GPT-4 has seen texts where this style of explanation is given for why telescoping sums work). It then correctly notes various cancellations, but demonstrates that it is not really understanding what it is saying by jumping to a final answer that is both incorrect and easily seen to be inconsistent with what it has just said. For instance, it notes that $f(2)$ is added once and subtracted twice, but its final answer includes a term $-2f(2)$ instead of $-f(2)$.

\subsection{Qualitative Observations from Case Studies}

We next synthesise core insights from the case studies. Similar to \texttt{MathConverse}, we see that the models can be good at providing definitions and interestingly can produce helpful scaffolding for a problem (e.g., the right strategy or first few steps), even if details of the full proof are incorrect. Moreover, even if the model does produce a correct solution, this solution is not always arrived at in a seemingly ``human-like'' way; for instance, the model may follow a guess-and-check approach rather than forward planning (e.g., Examples 1,4, and 7 in Section \ref{tim case studies}). However, guess-and-check cannot work well if one cannot ``check'' solutions; indeed, we see that in general, challenges with algebraic manipulation plague in- and out-of-distribution performance~(e.g., all three examples in Section \ref{sec: num thy}). By probing GPT-4 capabilities on slightly novel problems or those which involve building examples, we notice the model's tendency to over-rely on memorised examples or patterns. The case studies also reiterate potential issues with handling user corrections as well as a tendency towards over-verbosity. Further, one case study illuminated intriguing behaviour when the model was queried about uncertainty; i.e., the model began to apologise despite having been correct (see Section \ref{sec: wenda case studies}).

\pagebreak




\begin{figure}[!htb]
\centering
\begin{subfigure}{\textwidth}
  \centering
  \includegraphics[width=.8\linewidth]{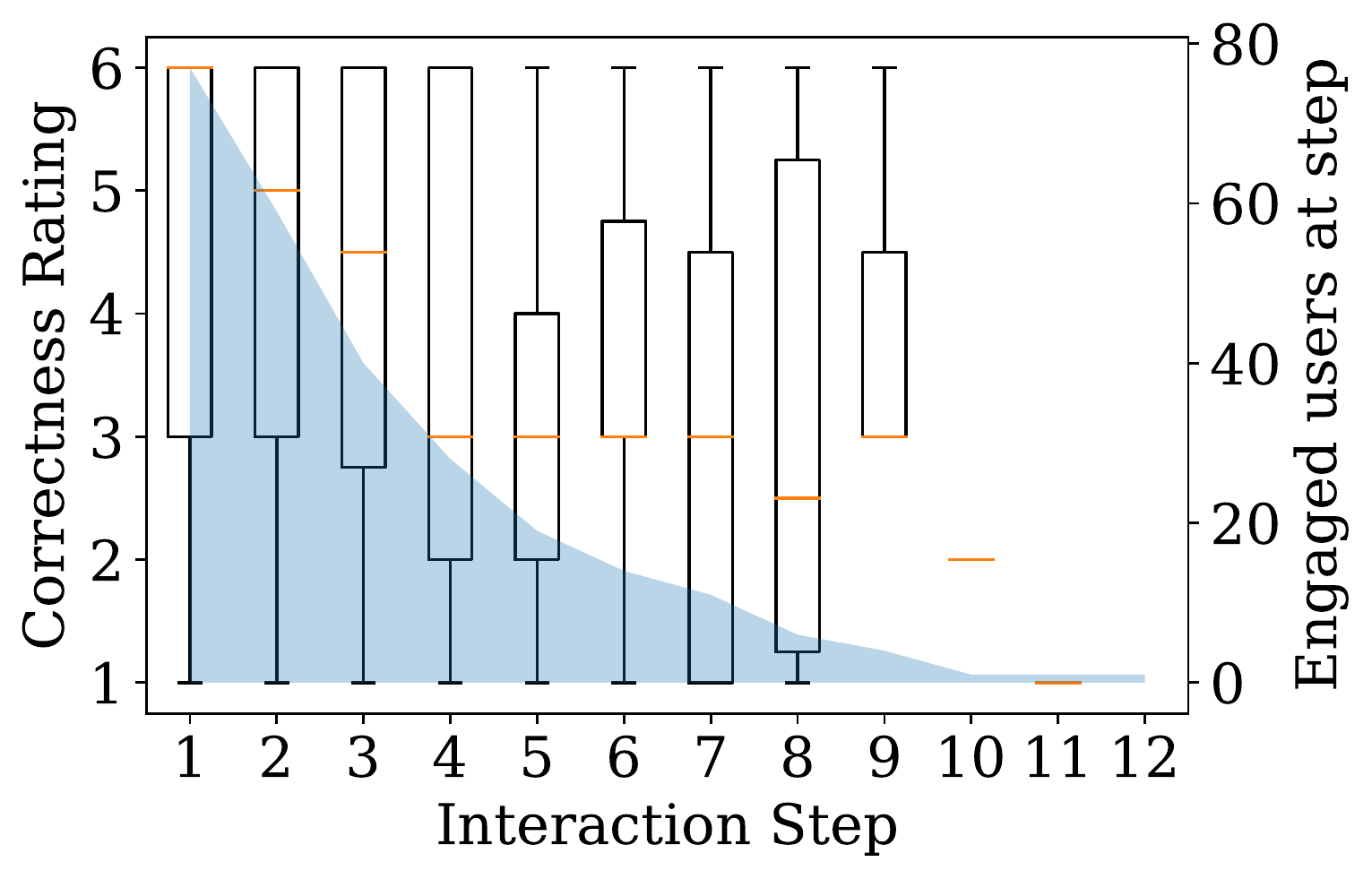}

\end{subfigure}
\begin{subfigure}{\textwidth}
  \centering
  \includegraphics[width=.8\linewidth]{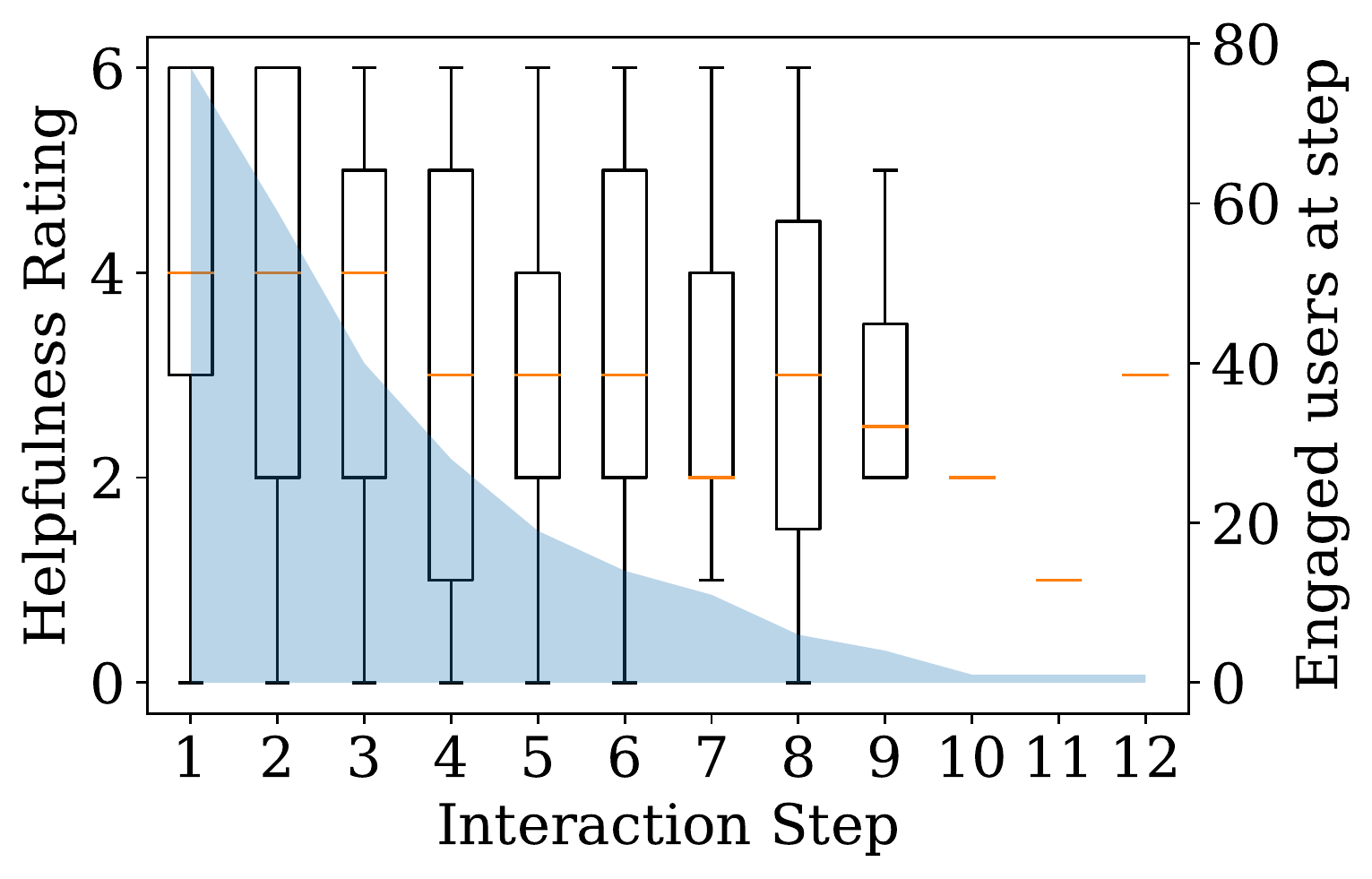}

\end{subfigure}
\caption{Depiction of how the users' correctness and helpfulness ratings progress with time. On the horizontal axis is the index of the interaction step. The box plot represents the correctness/helpfulness rating distributions given by users at each step~(vertical axis on the left), while the blue shade indicates how many users are still active at that given step~(vertical axis on the right). Note, few users undertook more than 9 interactions (3 people undertook 9 interactions, and only one person did 12), hence the boxplot collapses to a line with a single user. }
\label{fig:ratings_progress_with_time}
\end{figure}

\section{Full Interaction Traces from Expert Case Studies}

\subsection{Full Interaction Traces from ``Number Theory'' Case Studies} 

\textbf{Problem 21}

\begin{tabframebox}
\begin{longtable}{p{0.05\linewidth}  p{0.9\linewidth}}
\raisebox{-.62\totalheight}{\includegraphics[width=\linewidth]{figs/pngegg.png}} &  How would I show that the equation $1 + a^n = 2^m$ has no solutions in the integers for $m, n > 1$?\\\\
\raisebox{-.62\totalheight}{\includegraphics[width=\linewidth]{figs/openai-icon.png}} & To prove that the equation $1 + a^n = 2^m$ has no solutions in the integers for $m, n > 1$, we can proceed by contradiction. We'll assume there are solutions in integers, and then show that
this assumption leads to a contradiction.\\\\&

Let $1 + a^n = 2^m$ for some integers $a, m, n$ with $m, n > 1$.\\\\&

First, notice that $a$ must be odd since $1 + a^n$ is even. Let $a = 2k + 1$ for some integer $k$.\\\\&

Now, the equation becomes:\\&

$$1 + (2k + 1)^n = 2^m$$

Expand the left side using the binomial theorem:\\&

$$(2k + 1)^n = \sum_{i=0}^n \binom(n, i) (2k)^i = 1 + n(2k) + \sum_{i=2}^n \binom(n, i) (2k)^i$$

Now, notice that each term in the sum $\sum_{i=2}^n {n \choose i} (2k)^i$ is divisible by
$2^2 = 4$.\\\\&

Therefore, we can write the sum as $4s$ for some integer $s$. Then, we have:\\&

$$1 + 2kn + 4s = 2^m$$

$$2kn = 2^m - 1 - 4s$$\\

\raisebox{-.62\totalheight}{\includegraphics[width=\linewidth]{figs/pngegg.png}} & Can you check your work and see if there are any errors in this solution\\\\

\raisebox{-.62\totalheight}{\includegraphics[width=\linewidth]{figs/openai-icon.png}} & Apologies for the confusion in my previous response. You're right; there are errors in the solution. Let's revise the proof.\\\\&

We want to show that the equation $1 + a^n = 2^m$ has no solutions in the integers for
$m, n > 1$.\\\\&

Notice that $a$ must be odd since $1 + a^n$ is even. Let $a = 2k + 1$ for some integer $k$. Then,
the equation becomes:\\&

$$1 + (2k + 1)^n = 2^m$$

Expanding the left side using the binomial theorem:\\&

$$(2k + 1)^n = \sum_{i=0}^n {n \choose i} (2k)^i = 1 + n(2k) + \sum_{i=2}^n {n \choose i} (2k)^i$$

Now, observe that each term in the sum $\sum_{i=2}^n {n \choose i} (2k)^i$ is divisible by
$2^2 = 4$, as well as the term $n(2k)$.\\\\&

Therefore, we can write:\\&

$$1 + 2^m = 1 + 4s$$

for some integer $s$.\\\\&

From this, we have $2^m = 4s$, which implies $2^{m-1} = s$. Since $m > 1$, we have that
$m - 1 \ge 1$. So, $s$ must be even. Let $s = 2t$ for some integer $t$. Now we have:\\&

$$1 + a^n = 1 + 8t$$

Subtracting 1 from both sides, we get:\\&

$$a^n = 8t$$

Since the left side of the equation, $a^n$, is odd (because $a$ is odd), the right side must also
be odd. However, the right side is even (because $8t$ is a multiple of 8), which is a
contradiction. Therefore, there can be no solutions in integers for $m, n > 1$.

\end{longtable}
\end{tabframebox}


\textbf{Problem 23}
\begin{tabframebox}
\begin{longtable}{p{0.05\linewidth}  p{0.9\linewidth}}
\raisebox{-.62\totalheight}{\includegraphics[width=\linewidth]{figs/pngegg.png}} &  Can you prove that if $a$ and $m$ are coprime integers that $a^\phi(m) \equiv 1 \pmod{m}$ where $\phi$ is Euler's totient function?\\\\

\raisebox{-.62\totalheight}{\includegraphics[width=\linewidth]{figs/openai-icon.png}} & Yes, I can prove that if $a$ and $m$ are coprime integers, then $a^{\phi(m)} \equiv 1 \pmod{m}$. This result is known as Euler's theorem. I will provide a proof using the concept of modular arithmetic.\\\\&

1. Let's consider the set of integers less than $m$ that are coprime to $m$, denoted by
   $R_m = \{r_1, r_2, \ldots, r_{\phi(m)}\}$. The cardinality of this set is $\phi(m)$.\\\\&

2. Since $a$ is coprime to $m$, multiplying each element of $R_m$ by $a$ will yield another set
   of integers that are also coprime to $m$. We'll denote this new set as\\&
   $$S_m = \{ar_1, ar_2, \ldots, ar_{\phi(m)}\},$$

3. Notice that $S_m$ is a permutation of $R_m$, since each element in $S_m$ is congruent to a
   unique element in $R_m$ modulo $m$.\\\\&

4. Now, let's consider the product of all elements in $R_m$ and $S_m$. We can write this as:\\&

   $$\prod_{i=1}^{\phi(m)} r_i \equiv \prod_{i=1}^{\phi(m)} (ar_i) \pmod{m}$$

5. We can factor out the $a^{\phi(m)}$ term from the product on the right-hand side:\\&

   $$\prod_{i=1}^{\phi(m)} r_i \equiv a^{\phi(m)} \prod_{i=1}^{\phi(m)} r_i \pmod{m}$$

6. Since $\prod_{i=1}^{\phi(m)} r_i$ is coprime to $m$, we can safely cancel out this term from both sides:\\&

   $$a^{\phi(m)} \equiv 1 \pmod{m}$$\\

\raisebox{-.62\totalheight}{\includegraphics[width=\linewidth]{figs/pngegg.png}} &  Can you explain more fully why $R_m$ is a permutation of $S_m$ in step 3?\\\\

\raisebox{-.62\totalheight}{\includegraphics[width=\linewidth]{figs/openai-icon.png}} & Certainly! In step 3, we claim that $S_m$ is a permutation of $R_m$. To explain this more fully, let's revisit the definitions of $R_m$ and $S_m$.\\\\&

$R_m = {r_1, r_2, \ldots, r_{\phi(m)}}$ is the set of integers less than $m$ that are coprime to
$m$, and $S_m = {ar_1, ar_2, \ldots, ar_{\phi(m)}}$ is the set obtained by multiplying each
element of $R_m$ by $a$, where $a$ is an integer coprime to $m$.\\\\&

Now, we want to show that $S_m$ is a permutation of $R_m$, which means that the elements in
$S_m$ are just a rearrangement of the elements in $R_m$. To show this, we need to prove two
things:\\\\&

1. Every element of $S_m$ is coprime to $m$.\\\\&

2. The elements of $S_m$ are distinct modulo $m$.\\\\&

3. To prove that every element of $S_m$ is coprime to $m$, let's take an arbitrary element
$ar_i \in S_m$. Since $a$ is coprime to $m$ and $r_i$ is coprime to $m$ (by the definition of
$R_m$), their product $ar_i$ must also be coprime to $m$. This is due to the property that the
product of two coprime integers is coprime to their common divisor.\\\\&

Therefore, every element of $S_m$ is coprime to $m$.\\\\&

4. To prove that the elements of $S_m$ are distinct modulo $m$, let's assume that there are two
distinct elements $ar_i$ and $ar_j$ in $S_m$ such that $ar_i \equiv ar_j \pmod{m}$. Since $a$ is
coprime to $m$, we can multiply both sides of the congruence by $a^{-1}$ (the modular inverse of
$a$ modulo $m$).\\\\&

This gives us $r_i \equiv r_j \pmod{m}$, which contradicts the fact that $r_i$
and $r_j$ are distinct elements of $R_m$. Thus, the elements of $S_m$ must be distinct modulo
$m$.\\\\&

Since every element of $S_m$ is coprime to $m$ and the elements of $S_m$ are distinct modulo $m$,
we can conclude that $S_m$ is a permutation of $R_m$.\\\\

\raisebox{-.62\totalheight}{\includegraphics[width=\linewidth]{figs/pngegg.png}} & In your proof of 2, can you explain why $a$ must have a modular inverse?\\\\

\raisebox{-.62\totalheight}{\includegraphics[width=\linewidth]{figs/openai-icon.png}} & Apologies for not clarifying that part. In the proof of 2, we assume that $a$ has a modular inverse modulo $m$. This is indeed the case, and I'll explain why.\\\\&

Recall that $a$ and $m$ are coprime integers, which means their greatest common divisor (GCD) is
1. A number $a$ has a modular inverse modulo $m$ if and only if the GCD of $a$ and $m$ is 1. In
other words, a modular inverse of $a$ modulo $m$ exists if and only if $a$ and $m$ are coprime.\\\\&

Since $a$ and $m$ are coprime by assumption, there must exist integers $x$ and $y$ such that:\\&

$$ax + my = 1$$

This is a consequence of Bézout's identity, which states that the GCD of two integers can be
expressed as a linear combination of those integers. In our case, the GCD of $a$ and $m$ is 1.\\\\&

Now, if we take this equation modulo $m$, we get:\\&

$$ax + my \equiv 1 \pmod{m}$$

Since $my \equiv 0 \pmod{m}$, we have:\\\\&

$$ax \equiv 1 \pmod{m}$$

This means that $x$ is the modular inverse of $a$ modulo $m$. So, a modular inverse of $a$
exists, which is essential for our proof.

\end{longtable}
\end{tabframebox}


\textbf{Problem 24}
\begin{tabframebox}
\begin{longtable}{p{0.05\linewidth}  p{0.9\linewidth}}
\raisebox{-.62\totalheight}{\includegraphics[width=\linewidth]{figs/pngegg.png}} &  Let $\zeta$ be an irrational number. Show that there are infinitely many relatively prime integers $p$ and $q$ such that $|\zeta - p/q| < \frac{1}{\sqrt{5}q^2}$.\\\\

\raisebox{-.62\totalheight}{\includegraphics[width=\linewidth]{figs/openai-icon.png}} & To prove that there are infinitely many relatively prime integers $p$ and $q$ such that $|\zeta - p/q| < \frac{1}{\sqrt{5}q^2}$, we can use the theory of continued fractions.\\\\&

Every irrational number $\zeta$ can be represented by an infinite simple continued fraction:\\&

$$\zeta = a_0 + \cfrac{1}{a_1 + \cfrac{1}{a_2 + \cfrac{1}{a_3 + \cdots}}}$$\\&

where $a_0, a_1, a_2, \ldots$ are integers, and $a_i > 0$ for $i > 0$. The convergents of this
continued fraction are a sequence of rational numbers $\frac{p_n}{q_n}$ that can be computed
recursively as follows:\\&

$$p_n = a_n p_{n-1} + p_{n-2} \quad \text{and} \quad q_n = a_n q_{n-1} + q_{n-2}$$

with initial values $p_{-1} = 1$, $p_{-2} = 0$, $q_{-1} = 0$, and $q_{-2} = 1$. It is well known
that the convergents are reduced fractions, meaning that $p_n$ and $q_n$ are relatively prime.\\\\&

A result from the theory of continued fractions states that for any irrational number $\zeta$ and
its convergent $\frac{p_n}{q_n}$, we have:\\&

$$\left|\zeta - \frac{p_n}{q_n}\right| < \frac{1}{q_n q_{n+1}}$$

Now, let's consider the sequence of convergents of $\zeta$. Since there are infinitely many
convergents, we can find a convergent $\frac{p_n}{q_n}$ such that:\\&

$$q_n q_{n+1} > \sqrt{5} q_n^2$$

This inequality is satisfied when $q_{n+1} > \sqrt{5} q_n$.\\\\&

By the inequality above, we have:\\&

$$\left|\zeta - \frac{p_n}{q_n}\right| < \frac{1}{q_n q_{n+1}} < \frac{1}{\sqrt{5} q_n^2}$$

Thus, we have shown that there are infinitely many relatively prime integers $p$ and $q$ such
that $\left|\zeta - \frac{p}{q}\right| < \frac{1}{\sqrt{5} q^2}$.\\\\

\raisebox{-.62\totalheight}{\includegraphics[width=\linewidth]{figs/pngegg.png}} &  How can we prove that there will always be an $n$ for which $q_{n+1} > \sqrt{5} q_n$?\\\\

\raisebox{-.62\totalheight}{\includegraphics[width=\linewidth]{figs/openai-icon.png}} & To show that there will always be an $n$ for which $q_{n+1} > \sqrt{5} q_n$, we can exploit the fact that the continued fraction expansion contains arbitrarily large partial quotients.\\\\&

Let's assume the contrary: Suppose that for all $n$, we have $q_{n+1} \leq \sqrt{5} q_n$. Then,
we would have:\\&

$$q_{n+1} \leq \sqrt{5} q_n \leq 5 q_{n-1} \leq \cdots \leq 5^{n+1} q_0$$

This inequality implies that the denominators $q_n$ are bounded by an exponential function, which
contradicts the fact that the denominators of the convergents of an irrational number grow at
least linearly.\\\\&

The reason why the denominators grow at least linearly can be seen by looking at the recursive
formula for $q_n$:\\&

$$q_n = a_n q_{n-1} + q_{n-2}$$

For $a_n \geq 1$, we have:\\&

$$q_n \geq q_{n-1} + q_{n-2} \geq q_{n-1}$$

Thus, the sequence of denominators is non-decreasing. Moreover, if $a_{n'} \geq 2$ for some $n'$,
we have:\\&

$$q_{n'+1} = a_{n'+1} q_{n'} + q_{n'-1} \geq q_{n'} + q_{n'-1} \geq 2q_{n'}$$

This shows that the denominators grow at least linearly with $n$. So, we must have an $n$ for
which $q_{n+1} > \sqrt{5} q_n$, since otherwise, we would have a contradiction with the growth of
the denominators.

\end{longtable}
\end{tabframebox}

\textbf{Problem 27}
\begin{tabframebox}
\begin{longtable}{p{0.05\linewidth}  p{0.9\linewidth}}
\raisebox{-.62\totalheight}{\includegraphics[width=\linewidth]{figs/pngegg.png}} &  Suppose that $p$ is a prime number. Can you show that
$(a + b)^p \equiv a^p  + b^p \pmod{p}$ for all integers $a$ and $b$?\\\\

\raisebox{-.62\totalheight}{\includegraphics[width=\linewidth]{figs/openai-icon.png}} & Yes, we can show that $(a + b)^p \equiv a^p + b^p \pmod{p}$ for all integers $a$ and $b$ if $p$ is a prime number. We will use Fermat's Little Theorem and the binomial theorem to prove this.\\\\&

Fermat's Little Theorem states that if $p$ is a prime number and $a$ is an integer not divisible
by $p$, then $a^{p-1} \equiv 1 \pmod{p}$. Thus, $a^p \equiv a \pmod{p}$.\\\\&

Now, let's expand $(a+b)^p$ using the binomial theorem:\\&

$$(a + b)^p = \sum_{k=0}^{p} \binom{p}{k} a^{p-k} b^k$$

We know that for $0 < k < p$, the binomial coefficient $\binom{p}{k} = \frac{p!}{k!(p-k)!}$ is
an integer. Since $p$ is a prime, it only has two divisors: $1$ and $p$. Thus, if $0 < k < p$,
then $p$ must be a divisor of $\binom{p}{k}$, meaning that the binomial coefficient is divisible
by $p$. We can rewrite this as $\binom{p}{k} \equiv 0 \pmod{p}$ for $0 < k < p$.\\\\&

So, when we consider the expansion of $(a+b)^p$ modulo $p$, all the terms in the sum with
$0 < k < p$ will be equivalent to $0$ modulo $p$. Therefore, we are left with only two terms:\\&

$$(a + b)^p \equiv \binom{p}{0} a^p b^0 + \binom{p}{p} a^0 b^p \pmod{p}$$

Since $\binom{p}{0} = \binom{p}{p} = 1$, this simplifies to:\\&

$$(a + b)^p \equiv a^p + b^p \pmod{p}$$

And that completes the proof.

\end{longtable}
\end{tabframebox}

\textbf{Problem 28}
\begin{tabframebox}
\begin{longtable}{p{0.05\linewidth}  p{0.9\linewidth}}
\raisebox{-.62\totalheight}{\includegraphics[width=\linewidth]{figs/pngegg.png}} &  Can you show that\\&
$$3 = 1\times\sqrt{1 +2\times\sqrt{1 + 3\times\sqrt{\ldots}}}$$\\

\raisebox{-.62\totalheight}{\includegraphics[width=\linewidth]{figs/openai-icon.png}} & To show that\\&
$$3 = 1\times\sqrt{1 +2\times\sqrt{1 + 3\times\sqrt{\ldots}}},$$

we can start by defining the expression as a series with the following recursion relation:\\\\&

Let $a_n = \sqrt{1 + n \times a_{n+1}}$, where\\&

$$a_1 = 1\times\sqrt{1 +2\times\sqrt{1 + 3\times\sqrt{\ldots}}}$$

Our goal is to show that $a_1 = 3$.\\\\&

We can proceed using mathematical induction to prove this statement. First, let's consider the
base case, $n = 1$:\\&

$$a_1 = \sqrt{1 + 1 \times a_2}.$$

Now, let's assume the induction hypothesis holds for some $n = k$:\\&

$$a_k = \sqrt{1 + k \times a_{k+1}}$$

We want to show that this also holds for $n = k+1$:\\&

$$a_{k+1} = \sqrt{1 + (k+1) \times a_{k+2}}$$

From the induction hypothesis, we can substitute $a_{k+1}$:\\&

$$\sqrt{1 + k \times a_{k+1}} = \sqrt{1 + k \times \sqrt{1 + (k+1) \times a_{k+2}}}.$$

Now, let's square both sides of the equation:\\&

$$1 + k \times a_{k+1} = 1 + k \times \sqrt{1 + (k+1) \times a_{k+2}}$$

Next, subtract 1 from both sides:\\&

$$k \times a_{k+1} = k \times \sqrt{1 + (k+1) \times a_{k+2}}$$

Now, divide both sides by $k$:\\&

$$a_{k+1} = \sqrt{1 + (k+1) \times a_{k+2}}$$

So, the induction step holds for $n = k+1$.\\\\&

Now, let's see if we can find a pattern for the first few values of $a_n$:\\&

$$a_1 = \sqrt{1 + 1 \times a_2}$$
$$a_2 = \sqrt{1 + 2 \times a_3}$$
$$a_3 = \sqrt{1 + 3 \times a_4}$$

We can substitute the expression for $a_3$ into the expression for $a_2$:\\&

$$a_2 = \sqrt{1 + 2 \times \sqrt{1 + 3 \times a_4}}.$$

Now substitute the expression for $a_2$ into the expression for $a_1$:\\&

$$a_1 = \sqrt{1 + 1 \times \sqrt{1 + 2 \times \sqrt{1 + 3 \times a_4}}}.$$

Unfortunately, there is no simple way to close the recursion and prove that $a_1 = 3$ without
making additional assumptions or using more advanced mathematical techniques. This problem is
known as a nested radical, and finding a closed-form expression for such problems can be
challenging.\\\\

\raisebox{-.62\totalheight}{\includegraphics[width=\linewidth]{figs/pngegg.png}} &  Can you use the known value of $a_1$ to figure out what the value of $a_2$ would have to be and so on, and perhaps find a pattern?\\\\

\raisebox{-.62\totalheight}{\includegraphics[width=\linewidth]{figs/openai-icon.png}} & Sure! Given that $a_1 = 3$, let's find the value of $a_2$ and subsequent values using the recursive relation.\\&

$$a_1 = 3 = \sqrt{1 + 1 \times a_2}$$
$$a_2 = \sqrt{1 + 2 \times a_3}$$
$$a_3 = \sqrt{1 + 3 \times a_4}$$
$$a_4 = \sqrt{1 + 4 \times a_5}$$

We know the value of $a_1$, so let's find the value of $a_2$:\\&

$$3 = \sqrt{1 + 1 \times a_2}$$

Square both sides:\\&

$$9 = 1 + a_2$$

Solve for $a_2$:\\&

$$a_2 = 8$$

Now we can find the value of $a_3$:\\&

$$a_2 = 8 = \sqrt{1 + 2 \times a_3}$$

Square both sides:\\&

$$64 = 1 + 2 \times a_3$$

Subtract 1:\\&

$$63 = 2 \times a_3$$

Divide by 2:\\&

$$a_3 = 31.5$$

Now, find the value of $a_4$:\\&

$$a_3 = 31.5 = \sqrt{1 + 3 \times a_4}$$

Square both sides:\\&

$$992.25 = 1 + 3 \times a_4$$

Subtract 1:\\&

$$991.25 = 3 \times a_4$$

Divide by 3:\\&

$$a_4 = 330.416\overline{6}$$

So far, the sequence of values is:\\&

$$a_1 = 3, a_2 = 8, a_3 = 31.5, a_4 = 330.416\overline{6}$$

Unfortunately, there doesn't seem to be a simple pattern in the sequence. Nested radicals like
this one can often lead to complex and difficult-to-predict sequences.\\\\

\raisebox{-.62\totalheight}{\includegraphics[width=\linewidth]{figs/pngegg.png}} &  I believe you have the relations between the various $a_i$ incorrect. Can you go back to the original question and correct these and try again.\\\\

\raisebox{-.62\totalheight}{\includegraphics[width=\linewidth]{figs/openai-icon.png}} & Apologies for the confusion. Let's go back and redefine the recursive relation correctly:\\\\&

Let $a_n = \sqrt{1 + n \times a_{n-1}}$, with $a_0 = 0$.\\\\&

We want to show that\\&
$$a_1 = 1\times\sqrt{1 +2\times\sqrt{1 + 3\times\sqrt{\ldots}}} = 3.$$

Let's find the first few values of $a_n$:\\&

$$a_1 = \sqrt{1 + 1 \times a_0} = \sqrt{1 + 0} = \sqrt{1} = 1$$
$$a_2 = \sqrt{1 + 2 \times a_1} = \sqrt{1 + 2 \times 1} = \sqrt{3}$$
$$a_3 = \sqrt{1 + 3 \times a_2} = \sqrt{1 + 3 \times \sqrt{3}}$$
$$a_4 = \sqrt{1 + 4 \times a_3} = \sqrt{1 + 4 \times \sqrt{1 + 3 \times \sqrt{3}}}$$

Now, let's substitute the expressions for $a_n$ into the original equation:\\&

$$a_1 = 1\times\sqrt{1 +2\times\sqrt{1 + 3\times\sqrt{\ldots}}} = 1 \times \sqrt{1 + 2 \times a_2}$$

Since $a_2 = \sqrt{3}$, we have:\\&

$$a_1 = 1 \times \sqrt{1 + 2 \times \sqrt{3}}$$

Unfortunately, this does not simplify to $a_1 = 3$. Instead, we have a nested radical expression
for $a_1$, and we can't find a closed-form expression for this expression without making
additional assumptions or using more advanced mathematical techniques.\\\\&

It seems that the original statement, $3 = 1\times\sqrt(1 +2\times\sqrt(1 + 3\times\sqrt(\ldots)))$,
cannot be directly proven with the methods I have attempted so far.

\end{longtable}
\end{tabframebox}

\pagebreak

\subsection{Full Interaction Traces from ``Problem Perturbation to Probe Memorisation'' Case Studies}

\textbf{Problem 25}
\begin{tabframebox}
\begin{longtable}{p{0.05\linewidth}  p{0.9\linewidth}}
\raisebox{-.62\totalheight}{\includegraphics[width=\linewidth]{figs/pngegg.png}} &  Let $a, b \in \mathbb{N}_{>0}$ such that there exists no $m, n \in \mathbb{N}_{>0}$ such that $a^m = b^n$.\\&

Prove that $\log_b a$ is irrational.\\\\
\raisebox{-.62\totalheight}{\includegraphics[width=\linewidth]{figs/openai-icon.png}} & 
    We will prove this by contradiction. Suppose that $\log_b a$ is rational. Then we can write $\log_b a = \frac{p}{q}$, where $p$ and $q$ are coprime integers and $q > 1$.\\&

    By definition of logarithm, we have:\\&

    $$b^{\frac{p}{q}} = a$$

    Raise both sides to the power of $q$:\\&

    $$(b^{\frac{p}{q}})^q = a^q$$
    $$b^p = a^q$$

    Since $a, b \in \mathbb{N}_{>0}$, it follows that $a^q$ and $b^p$ are both positive integers. But this contradicts the initial assumption that there exists no $m, n \in \mathbb{N}_{>0}$ such that $a^m = b^n$. In this case, we found that $a^q = b^p$ where $q, p \in \mathbb{N}_{>0}$.\\\\&

    Therefore, our assumption that $\log_b a$ is rational must be false, which implies that $\log_b a$ is irrational.\\\\
\raisebox{-.62\totalheight}{\includegraphics[width=\linewidth]{figs/pngegg.png}} & What is the definition of logarithm here?\\\\
\raisebox{-.62\totalheight}{\includegraphics[width=\linewidth]{figs/openai-icon.png}} & The logarithm is a mathematical function that helps us find the exponent to which a given base must be raised to produce a specific number. In general, the logarithm with base $b$ of a positive number $a$ is denoted as $\log_b a$ and is defined as follows:\\\\&

    Given a positive real number $a$, and a positive base $b \neq 1$, the logarithm of $a$ with base $b$, denoted as $\log_b a$, is the exponent to which $b$ must be raised to obtain the number $a$. In other words, if:\\&

    $$b^x = a$$

    Then:\\&

    $$x = \log_b a$$

    The logarithm function has various properties and is widely used in many branches of mathematics, science, and engineering. The most common logarithm bases are base 10 (common logarithm) and base $e$ (natural logarithm), denoted as $\log a$ and $\ln a$, respectively.\\\\
\raisebox{-.62\totalheight}{\includegraphics[width=\linewidth]{figs/pngegg.png}} & Can a logarithm take a base less than 1?\\\\
\raisebox{-.62\totalheight}{\includegraphics[width=\linewidth]{figs/openai-icon.png}} & Yes, a logarithm can have a base between 0 and 1, but not equal to 0 or 1. In general, the logarithm function with base $b$ is well-defined for any positive real number $b \neq 1$. \\\\&

    When the base $b$ is between 0 and 1 (i.e., $0 < b < 1$), the logarithm function has a different behaviour compared to when the base is greater than 1. Specifically, the logarithm function is decreasing (i.e., its value decreases as the input increases) when the base is between 0 and 1. This is because a number between 0 and 1 raised to a higher power becomes smaller, while a number between 0 and 1 raised to a smaller power becomes larger.\\\\& 

    For example, consider the logarithm with base $\frac{1}{2}$:\\&

    $$\log_{\frac{1}{2}} a$$

    If $a > 1$, then we need a negative exponent to make the base $\frac{1}{2}$ equal to $a$. In other words, the logarithm will be negative for values of $a > 1$. As $a$ increases, the magnitude of the negative exponent required to make $\frac{1}{2}$ equal to $a$ also increases, so the logarithm function is decreasing in this case.\\\\
\raisebox{-.62\totalheight}{\includegraphics[width=\linewidth]{figs/pngegg.png}} & What is the meaning of $\mathbb{N}_{>0}$ in the previous proof?\\\\
\raisebox{-.62\totalheight}{\includegraphics[width=\linewidth]{figs/openai-icon.png}} & In the previous proof, the notation $\mathbb{N}_{>0}$ denotes the set of natural numbers greater than 0. The set of natural numbers $\mathbb{N}$ typically includes all non-negative integers, starting from 0:\\&

    $$\mathbb{N} = \{0, 1, 2, 3, 4, ...\}$$

    However, sometimes we want to consider only positive integers (excluding 0). In this case, we use the notation $\mathbb{N}_{>0}$ to represent the set of all positive integers:\\&

    $$\mathbb{N}_{>0} = \{1, 2, 3, 4, ...\}$$

    In the context of the previous proof, $a, b \in \mathbb{N}_{>0}$ means that both $a$ and $b$ belong to the set of positive integers.\\\\
\raisebox{-.62\totalheight}{\includegraphics[width=\linewidth]{figs/pngegg.png}} & Can we still finish the proof if we instead have $a, b \in \mathbb{R}_{>0}$?\\\\
\raisebox{-.62\totalheight}{\includegraphics[width=\linewidth]{figs/openai-icon.png}} & Yes, we can still complete the proof for $a, b \in \mathbb{R}_{>0}$, i.e., $a$ and $b$ are positive real numbers. We will use a similar approach, proving by contradiction. Suppose that $\log_b a$ is rational. Then we can write $\log_b a = \frac{p}{q}$, where $p$ and $q$ are coprime integers and $q > 1$.\\\\&

    By definition of logarithm, we have:\\&

    $$b^{\frac{p}{q}} = a$$

    Raise both sides to the power of $q$:\\&

    $$(b^{\frac{p}{q}})^q = a^q$$
    $$b^p = a^q$$

    Since $a, b \in \mathbb{R}_{>0}$, it follows that $a^q$ and $b^p$ are both positive real numbers. However, this contradicts the initial assumption that there exists no $m, n \in \mathbb{N}_{>0}$ such that $a^m = b^n$. In this case, we found that $a^q = b^p$ where $q, p \in \mathbb{N}_{>0}$.\\\\&

    Therefore, our assumption that $\log_b a$ is rational must be false, which implies that $\log_b a$ is irrational. Note that this proof holds for both positive integers and positive real numbers.\\\\
\raisebox{-.62\totalheight}{\includegraphics[width=\linewidth]{figs/pngegg.png}} & Could $\log_b a = \frac{p}{q}$ be a negative number?\\\\
\raisebox{-.62\totalheight}{\includegraphics[width=\linewidth]{figs/openai-icon.png}} & Yes, $\log_b a = \frac{p}{q}$ can be a negative number, depending on the values of $a$ and $b$. Recall that the logarithm of a number $a$ with base $b$ is the exponent to which $b$ must be raised to obtain the number $a$. In other words, if $b^x = a$, then $x = \log_b a$.\\\\&

    If $a$ is a number between 0 and 1 (i.e., $0 < a < 1$) and $b > 1$, then the logarithm $\log_b a$ will be negative. This is because, to obtain a number between 0 and 1 when the base $b$ is greater than 1, we need to raise $b$ to a negative exponent. In this case, $x$ (or $\log_b a$) will be negative.\\\\&

    For example, consider the following logarithm:\\&

    $$\log_2 \frac{1}{4}$$

    We know that $2^{-2} = \frac{1}{4}$, so in this case, $\log_2 \frac{1}{4} = -2$. Here, $\log_b a$ is a negative number.\\\\&

    In general, $\log_b a$ can be positive, negative, or even zero depending on the values of $a$ and $b$. It's important to note that when the base $b$ is between 0 and 1, the relationship between $a$ and the sign of $\log_b a$ is reversed; that is, when $a > 1$, the logarithm will be negative, and when $0 < a < 1$, the logarithm will be positive.\\\\
\raisebox{-.62\totalheight}{\includegraphics[width=\linewidth]{figs/pngegg.png}} &  How about if we drop the assumption that $b \in \mathbb{R}_{>0}$? Does this affect the original proof?\\\\
\raisebox{-.62\totalheight}{\includegraphics[width=\linewidth]{figs/openai-icon.png}} & If we drop the assumption that $b \in \mathbb{R}_{>0}$, then $b$ could be a non-positive real number or not a real number at all. In either case, the original proof will be affected, as the logarithm function is not well-defined for bases that are not positive real numbers different from 1.\\&

\begin{enumerate}
    \item If $b \le 0$, the logarithm function is not defined for negative or zero bases. In this case, the original proof would not hold, as the logarithm function cannot be used with such bases.
    \item If $b$ is not a real number (e.g., a complex number), the original proof would also not hold. The logarithm function for complex numbers, called the complex logarithm, behaves differently from the real logarithm function. The properties of real logarithms, like the monotonic behaviour and properties used in the proof, may not apply to complex logarithms.
\end{enumerate}

    So, in order to use the original proof, it is essential to maintain the assumption that $b \in \mathbb{R}_{>0}$ and $b \neq 1$.\\\\
\raisebox{-.62\totalheight}{\includegraphics[width=\linewidth]{figs/pngegg.png}} & How about the assumption that $c \in \mathbb{R}_{>0}$? Can we still proceed with the original proof if we don't have this assumption?\\\\
\raisebox{-.62\totalheight}{\includegraphics[width=\linewidth]{figs/openai-icon.png}} & It seems that you're introducing a new variable $c$. To clarify the context of your question, let's restate the original problem: We have $a, b \in \mathbb{R}_{>0}$ (or $a, b \in \mathbb{N}_{>0}$ in the original question) such that there exists no $m, n \in \mathbb{N}_{>0}$ such that $a^m = b^n$. The goal is to prove that $\log_b a$ is irrational.\\\\&

    If you meant to ask about either $a$ or $b$ instead of introducing a new variable $c$, then please refer to my previous responses addressing the effects of dropping the assumptions on $a$ and $b$.\\\\&

    If you indeed intended to introduce a new variable $c$ and are asking about its effect on the original proof without specifying its role, it is not clear how $c$ would be related to the problem or the proof. Please provide more context or clarify the role of $c$ in the problem, and I'd be happy to help.\\\\
    
\raisebox{-.62\totalheight}{\includegraphics[width=\linewidth]{figs/pngegg.png}} & Sorry I made a mistake. What I meant was to drop the assumption that $a \in \mathbb{R}_{>0}$. Can we still finish the proof?\\\\
\raisebox{-.62\totalheight}{\includegraphics[width=\linewidth]{figs/openai-icon.png}} & If we drop the assumption that $a \in \mathbb{R}_{>0}$, then $a$ could be a non-positive real number or not a real number at all. In either case, the original proof will be affected, as the logarithm function is not well-defined for non-positive input values.\\&
\begin{enumerate}
    \item If $a \le 0$, the logarithm function is not defined for negative or zero input values. In this case, the original proof would not hold, as the logarithm function cannot be used with such input values.
    \item If $a$ is not a real number (e.g., a complex number), the original proof would also not hold. The logarithm function for complex numbers, called the complex logarithm, behaves differently from the real logarithm function. The properties of real logarithms, like the properties used in the proof, may not apply to complex logarithms.
\end{enumerate}
    So, in order to use the original proof, it is essential to maintain the assumption that $a \in \mathbb{R}_{>0}$.
\end{longtable}
\end{tabframebox}

\textbf{Problem 39 (Chebyshev's inequality)}
\begin{tabframebox}
\begin{longtable}{p{0.05\linewidth}  p{0.9\linewidth}}
\raisebox{-.62\totalheight}{\includegraphics[width=\linewidth]{figs/pngegg.png}} & Let $X$ be a random variable. Assume $\mathsf{E} (X) = \mu$ for some $\mu \in \mathbb{R}$ and $\mathsf{var} (X) = \sigma^2$ for some $\sigma^2 \in \mathbb{R}_{> 0}$.\\\\&

    Show that for all $k > 0$:\\&

    $$\Pr \left({\left| {X - \mu}\right| \geq k \sigma}\right) \leq \dfrac {1}{k^2}.$$\\\\
\raisebox{-.62\totalheight}{\includegraphics[width=\linewidth]{figs/openai-icon.png}} & To show this, we will use the Chebyshev's inequality. Chebyshev's inequality states that for any random variable $Y$ with expected value $\mathsf{E}(Y) = m$ and variance $\mathsf{var}(Y) = v^2$:\\&

    $$\Pr\left({\left| {Y - m}\right| \geq k v}\right) \leq \dfrac{1}{k^2}$$

    for all $k > 0$.\\\\&

    Now, let's apply Chebyshev's inequality to the given random variable $X$. We are given that $\mathsf{E}(X) = \mu$ and $\mathsf{var}(X) = \sigma^2$. Thus, we have:\\&

    $$\Pr\left({\left| {X - \mu}\right| \geq k \sigma}\right) \leq \dfrac{1}{k^2}$$

    This proves the desired result.\\\\
\raisebox{-.62\totalheight}{\includegraphics[width=\linewidth]{figs/pngegg.png}} & Will the proof still hold when $k=2$?\\\\
\raisebox{-.62\totalheight}{\includegraphics[width=\linewidth]{figs/openai-icon.png}} & Yes, the proof still holds when $k=2$. Chebyshev's inequality is valid for all $k > 0$, which includes $k=2$. When $k=2$, the inequality states:\\&

    $$\Pr\left({\left| {X - \mu}\right| \geq 2 \sigma}\right) \leq \dfrac{1}{2^2} = \dfrac{1}{4}$$

    This means that the probability that the random variable $X$ deviates from its mean $\mu$ by more than $2\sigma$ (twice the standard deviation) is at most $\dfrac{1}{4}$, or 25\%. In other words, there is at least a 75\% chance that $X$ lies within the interval $(\mu - 2\sigma, \mu + 2\sigma)$.\\\\
\raisebox{-.62\totalheight}{\includegraphics[width=\linewidth]{figs/pngegg.png}} & How about $k=\sqrt{2}$? Will this affect the derivation?\\\\
\raisebox{-.62\totalheight}{\includegraphics[width=\linewidth]{figs/openai-icon.png}} & Yes, the derivation still holds for $k=\sqrt{2}$. Chebyshev's inequality is valid for any positive value of $k$, including $k=\sqrt{2}$. When $k=\sqrt{2}$, the inequality states:\\&

    $$\Pr\left({\left| {X - \mu}\right| \geq \sqrt{2} \sigma}\right) \leq \dfrac{1}{(\sqrt{2})^2} = \dfrac{1}{2}$$

    This means that the probability that the random variable $X$ deviates from its mean $\mu$ by more than $\sqrt{2}\sigma$ (approximately $1.414$ times the standard deviation) is at most $\dfrac{1}{2}$, or 50\%. In other words, there is at least a 50\% chance that $X$ lies within the interval $(\mu - \sqrt{2}\sigma, \mu + \sqrt{2}\sigma)$.\\\\&

    So, the derivation remains valid for $k=\sqrt{2}$ and any other positive value of $k$.\\\\
    
\raisebox{-.62\totalheight}{\includegraphics[width=\linewidth]{figs/pngegg.png}} & What if $k=\sqrt{2} - 1$?\\\\
\raisebox{-.62\totalheight}{\includegraphics[width=\linewidth]{figs/openai-icon.png}} & The derivation still holds for $k=\sqrt{2}-1$. Chebyshev's inequality is valid for any positive value of $k$, which includes $k=\sqrt{2}-1$. When $k=\sqrt{2}-1$, the inequality states:\\&

    $$\Pr\left({\left| {X - \mu}\right| \geq (\sqrt{2} - 1) \sigma}\right) \leq \dfrac{1}{(\sqrt{2}-1)^2}$$

    This means that the probability that the random variable $X$ deviates from its mean $\mu$ by more than $(\sqrt{2}-1)\sigma$ (approximately $0.414$ times the standard deviation) is at most $\dfrac{1}{(\sqrt{2}-1)^2}$.\\\\&

    In other words, there is at least a $1-\dfrac{1}{(\sqrt{2}-1)^2}$ chance that $X$ lies within the interval $(\mu - (\sqrt{2}-1)\sigma, \mu + (\sqrt{2}-1)\sigma)$.\\\\&

    So, the derivation remains valid for $k=\sqrt{2}-1$ and any other positive value of $k$.\\\\
    
\raisebox{-.62\totalheight}{\includegraphics[width=\linewidth]{figs/pngegg.png}} & What if $k=\sqrt{2} - 2$?\\\\
\raisebox{-.62\totalheight}{\includegraphics[width=\linewidth]{figs/openai-icon.png}} & When $k = \sqrt{2} - 2$, the value of $k$ becomes negative because $\sqrt{2} < 2$. Since Chebyshev's inequality is only valid for $k > 0$, it does not apply when $k = \sqrt{2} - 2$.\\\\& 

    In this case, you cannot use Chebyshev's inequality to derive a bound on the probability of the random variable $X$ deviating from its mean by a factor of $(\sqrt{2} - 2) \sigma$.\\\\
    
\raisebox{-.62\totalheight}{\includegraphics[width=\linewidth]{figs/pngegg.png}} & What if $k=(\sqrt{2} -2)^2$?\\\\
\raisebox{-.62\totalheight}{\includegraphics[width=\linewidth]{figs/openai-icon.png}} & When $k=(\sqrt{2} - 2)^2$, the value of $k$ is positive because $(\sqrt{2} - 2)^2 > 0$. Since Chebyshev's inequality is valid for $k > 0$, we can apply it in this case. When $k=(\sqrt{2}-2)^2$, the inequality states:\\& 

    $$\Pr\left({\left| {X - \mu}\right| \geq (\sqrt{2} - 2)^2 \sigma}\right) \leq \dfrac{1}{(\sqrt{2}-2)^4}$$

    This means that the probability that the random variable $X$ deviates from its mean $\mu$ by more than $(\sqrt{2}-2)^2\sigma$ (approximately $0.1716$ times the standard deviation) is at most $\dfrac{1}{(\sqrt{2}-2)^4}$.\\\\& 

    In other words, there is at least a $1-\dfrac{1}{(\sqrt{2}-2)^4}$ chance that $X$ lies within the interval $(\mu - (\sqrt{2}-2)^2\sigma, \mu + (\sqrt{2}-2)^2\sigma)$.\\\\&

    So, the derivation remains valid for $k=(\sqrt{2}-2)^2$ and any other positive value of $k$.\\\\
    
\raisebox{-.62\totalheight}{\includegraphics[width=\linewidth]{figs/pngegg.png}} & Are you sure $(\sqrt{2} - 2)^2 > 0$?\\\\
\raisebox{-.62\totalheight}{\includegraphics[width=\linewidth]{figs/openai-icon.png}} & Apologies for the confusion in my previous response. You are right; my statement was incorrect. \\\\&

    When $k = (\sqrt{2} - 2)^2$, the value of $k$ is indeed non-negative, but it is actually equal to 0, not greater than 0. Here is the correct calculation:\\&

    $$(\sqrt{2} - 2)^2 = (1.414 - 2)^2 = (-0.586)^2 = 0.343396$$

    Since $0.343396 > 0$, the value of $k$ is positive in this case, and Chebyshev's inequality can be applied. \\\\&

    When $k = (\sqrt{2} - 2)^2$, the inequality states:\\&

    $$\Pr\left({\left| {X - \mu}\right| \geq (\sqrt{2} - 2)^2 \sigma}\right) \leq \dfrac{1}{(\sqrt{2} - 2)^4}$$

    This means that the probability that the random variable $X$ deviates from its mean $\mu$ by more than $(\sqrt{2} - 2)^2\sigma$ (approximately $0.3434$ times the standard deviation) is at most $\dfrac{1}{(\sqrt{2} - 2)^4}$.\\\\&

    In other words, there is at least a $1-\dfrac{1}{(\sqrt{2} - 2)^4}$ chance that $X$ lies within the interval $(\mu - (\sqrt{2} - 2)^2\sigma, \mu + (\sqrt{2} - 2)^2\sigma)$.\\\\&

    So, the derivation remains valid for $k = (\sqrt{2} - 2)^2$ and any other positive value of $k$.\\\\

\raisebox{-.62\totalheight}{\includegraphics[width=\linewidth]{figs/pngegg.png}} & Why do you say your previous statement was incorrect and $k=0$?  I don't understand.\\\\
\raisebox{-.62\totalheight}{\includegraphics[width=\linewidth]{figs/openai-icon.png}} & I apologize for any confusion my previous responses may have caused. I made a mistake in my explanation. Let me clarify the situation:\\\\&

    When $k = (\sqrt{2} - 2)^2$, the value of $k$ is indeed positive, as I mentioned in my previous response. Here is the correct calculation:\\&

    $$(\sqrt{2} - 2)^2 = (1.414 - 2)^2 = (-0.586)^2 = 0.343396$$

    Since $0.343396 > 0$, the value of $k$ is positive in this case, and Chebyshev's inequality can be applied.\\\\&

    When $k = (\sqrt{2} - 2)^2$, the inequality states:\\&

    $$\Pr\left({\left| {X - \mu}\right| \geq (\sqrt{2} - 2)^2 \sigma}\right) \leq \dfrac{1}{(\sqrt{2} - 2)^4}$$

    This means that the probability that the random variable $X$ deviates from its mean $\mu$ by more than $(\sqrt{2} - 2)^2\sigma$ (approximately $0.3434$ times the standard deviation) is at most $\dfrac{1}{(\sqrt{2} - 2)^4}$.\\\\&

    In other words, there is at least a $1-\dfrac{1}{(\sqrt{2} - 2)^4}$ chance that $X$ lies within the interval $(\mu - (\sqrt{2} - 2)^2\sigma, \mu + (\sqrt{2} - 2)^2\sigma)$.\\\\&

    So, the derivation remains valid for $k = (\sqrt{2} - 2)^2$ and any other positive value of $k$.
\end{longtable}
\end{tabframebox}
\pagebreak

\subsection{Full Interaction Traces from ``Boundary of Easy and Hard'' Case Studies}

We include the full interaction traces from case studies designed and undertaken by Timothy Gowers in Section \ref{tim case studies}.  

\textbf{Example 1 (from section \ref{case studies ex1})}

\begin{tabframebox}
\begin{longtable}{p{0.05\linewidth}  p{0.9\linewidth}}
\raisebox{-.62\totalheight}{\includegraphics[width=\linewidth]{figs/pngegg.png}} & Can you find an example of a set A of positive integers that satisfies the following three conditions?
\begin{enumerate}
    \item $|A| = \min(A)$.
    \item $A$ consists of prime numbers only.
    \item $A$ has cardinality at least 5.
\end{enumerate}\\
\raisebox{-.62\totalheight}{\includegraphics[width=\linewidth]{figs/openai-icon.png}} & Yes, here's an example of a set A of positive integers that satisfies the given conditions:

A = \{5, 11, 17, 23, 29\}
\begin{enumerate}
    \item The cardinality of A is 5, which is also the minimum element in A.
    \item All elements in A are prime numbers.
    \item The cardinality of A is 5, which is at least 5.
\end{enumerate}

Therefore, set A = \{5, 11, 17, 23, 29\} meets all three conditions.\\\\
\raisebox{-.62\totalheight}{\includegraphics[width=\linewidth]{figs/pngegg.png}} & Can you find an example of a set A of positive integers that satisfies the following three conditions?
\begin{enumerate}
    \item $|A| = \min(A)$.
    \item $A$ consists of prime numbers only.
    \item $A$ has cardinality at least 6.
\end{enumerate}\\
\raisebox{-.62\totalheight}{\includegraphics[width=\linewidth]{figs/openai-icon.png}} & Yes, here is an example of a set A of positive integers that satisfies the given conditions:

A = {7, 11, 13, 17, 19, 23}

\begin{enumerate}
    \item |A| = 6, and the minimum element in A is 7, so |A| = min(A).
    \item All elements in A are prime numbers: 7, 11, 13, 17, 19, and 23.
    \item A has cardinality 6, which is at least 6.
\end{enumerate}

The set A satisfies all three conditions.\\\\
\raisebox{-.62\totalheight}{\includegraphics[width=\linewidth]{figs/pngegg.png}} & 6 does not equal 7.\\\\
\raisebox{-.62\totalheight}{\includegraphics[width=\linewidth]{figs/openai-icon.png}} &
You are correct, my previous response contained an error. I apologize for the confusion. It is not possible to construct a set A of positive integers that satisfies all three conditions simultaneously.

To see why, let's consider the conditions:

\begin{enumerate}
    \item |A| = min(A): This implies that the minimum element of the set A is equal to the size of the set A.
    \item A consists of prime numbers only: This implies that every element of A is a prime number.
    \item A has cardinality at least 6: This implies that A has 6 or more elements.
\end{enumerate}\\
&
The smallest prime number is 2. If the set A consists of prime numbers only, the minimum element of A must be greater than or equal to 2. However, if |A| = min(A) and A has at least 6 elements, the minimum element in A would need to be at least 6, which is not prime. This creates a contradiction, making it impossible to construct a set A that satisfies all three conditions simultaneously.
\end{longtable}
\end{tabframebox}

\textbf{Example 2 (from section \ref{case studies ex2})}

\begin{tabframebox}
\begin{longtable}{p{0.05\linewidth}  p{0.9\linewidth}}
\raisebox{-.62\totalheight}{\includegraphics[width=\linewidth]{figs/pngegg.png}} & Can you work out the integral $\int_{-1}^1 x^3\cos x dx$?\\\\
\raisebox{-.62\totalheight}{\includegraphics[width=\linewidth]{figs/openai-icon.png}} & To solve the integral, we can use integration by parts. Integration by parts states that for two differentiable functions $u$ and $v$:

$$\int u\,dv = uv - \int v\,du.$$

Let's choose $u = x^3$ and $dv = \cos x\,dx$. Then, we compute $du$ and $v$:

$$du = \frac{d}{dx}(x^3) dx = 3x^2 dx, \quad v = \int \cos x\,dx = \sin x.$$

Now, using integration by parts:

$$\int_{-1}^1 x^3\cos x\,dx = \left[x^3\sin x\right]_{-1}^1 - \int_{-1}^1 \sin x(3x^2) dx.$$

Let's evaluate the first term:

$$\left[x^3\sin x\right]_{-1}^1 = (1^3\sin(1) - (-1)^3\sin(-1)) = \sin(1) + \sin(-1).$$

For the remaining integral, we apply integration by parts again. This time, let $u = 3x^2$ and $dv = \sin x\,dx$. Then, we compute $du$ and $v$:

$$du = \frac{d}{dx}(3x^2) dx = 6x dx, \quad v = \int \sin x\,dx = -\cos x.$$

Now, using integration by parts:

$$\int_{-1}^1 3x^2\sin x\,dx = \left[-3x^2\cos x\right]_{-1}^1 - \int_{-1}^1 -6x\cos x\,dx.$$\\
&

Evaluating the first term:

$$\left[-3x^2\cos x\right]_{-1}^1 = -3(1^2)\cos(1) - (-3(-1)^2\cos(-1)) = -3\cos(1) + 3\cos(-1).$$

For the remaining integral, notice that the integrand $-6x\cos x$ is an odd function. The integral of an odd function over a symmetric interval around zero is always zero:

$$\int_{-1}^1 -6x\cos x\,dx = 0.$$

So, our final result is:

$$\int_{-1}^1 x^3\cos x\,dx = \sin(1) + \sin(-1) - 3\cos(1) + 3\cos(-1) + 0 = 2\sin(1) - 6\cos(1).$$\\

\raisebox{-.62\totalheight}{\includegraphics[width=\linewidth]{figs/pngegg.png}} & What is $\sin(1) + \sin(-1)$?\\\\

\raisebox{-.62\totalheight}{\includegraphics[width=\linewidth]{figs/openai-icon.png}} &
To find the value of $\sin(1) + \sin(-1)$, we can use the property of the sine function that states that $\sin(-x) = -\sin(x)$ for any real number $x$. 

So, $\sin(-1) = -\sin(1)$. 

Therefore, 

$$\sin(1) + \sin(-1) = \sin(1) - \sin(1) = 0.$$\\

\raisebox{-.62\totalheight}{\includegraphics[width=\linewidth]{figs/pngegg.png}} & How about $-3\cos(1)+3\cos(-1)$?\\\\

\raisebox{-.62\totalheight}{\includegraphics[width=\linewidth]{figs/openai-icon.png}} & To find the value of $-3\cos(1) + 3\cos(-1)$, we can use the property of the cosine function that states that $\cos(-x) = \cos(x)$ for any real number $x$. 

So, $\cos(-1) = \cos(1)$. 

Therefore, 

$$-3\cos(1) + 3\cos(-1) = -3\cos(1) + 3\cos(1) = 0.$$\\

\raisebox{-.62\totalheight}{\includegraphics[width=\linewidth]{figs/pngegg.png}} & 
In the light of that, would you like to revise your final answer above? (Please do not repeat the entire calculation -- the rest was fine and just the last step was a problem.)\\\\

\raisebox{-.62\totalheight}{\includegraphics[width=\linewidth]{figs/openai-icon.png}} &
Apologies for the confusion in my previous response. Given that $\sin(1) + \sin(-1) = 0$ and $-3\cos(1) + 3\cos(-1) = 0$, the final answer for the integral should be:

$$\int_{-1}^1 x^3\cos x\,dx = 2\sin(1) - 6\cos(1) + 0 = 2\sin(1) - 6\cos(1).$$

However, as we just found out, the correct values for these sums are both 0. Therefore, the correct answer is:

$$\int_{-1}^1 x^3\cos x\,dx = 0.$$\\

\raisebox{-.62\totalheight}{\includegraphics[width=\linewidth]{figs/pngegg.png}} & 
Can you establish that in an easier way?\\\\

\raisebox{-.62\totalheight}{\includegraphics[width=\linewidth]{figs/openai-icon.png}} & Yes, there is an easier way to establish that the integral is zero by noticing that the integrand is an odd function.\\\\&

An odd function is a function that satisfies the property $f(-x) = -f(x)$ for all $x$. The product of an odd function and an even function is an odd function. In this case, $x^3$ is an odd function and $\cos x$ is an even function. Therefore, their product $x^3 \cos x$ is an odd function.\\\\&

Now, when we integrate an odd function over a symmetric interval around zero (like $[-1, 1]$), the integral is always zero. This is because the positive and negative areas under the curve cancel each other out.\\\\&

So, for the given integral:

$$\int_{-1}^1 x^3\cos x\,dx = 0$$

because the integrand is an odd function and the interval is symmetric around zero.\\\\

\raisebox{-.62\totalheight}{\includegraphics[width=\linewidth]{figs/pngegg.png}} & 
Can you calculate the integral $\int_{-1}^1(x^2-1/3)dx$?\\\\

\raisebox{-.62\totalheight}{\includegraphics[width=\linewidth]{figs/openai-icon.png}} & Certainly! Let's evaluate the given integral:\\&

$$\int_{-1}^1 (x^2 - \frac{1}{3})\,dx.$$\\&

The integral of a sum (or difference) of functions is equal to the sum (or difference) of their integrals. So we can split the integral into two parts:\\&

$$\int_{-1}^1 x^2\,dx - \int_{-1}^1 \frac{1}{3}\,dx.$$\\&

To find the antiderivative of $x^2$, we can use the power rule for integration:\\&

$$\int x^n\,dx = \frac{x^{n+1}}{n+1} + C.$$\\&

In this case, $n = 2$. So the antiderivative of $x^2$ is:\\&

$$\frac{x^{2+1}}{2+1} = \frac{x^3}{3}.$$\\&

Now, let's evaluate the first integral:\\&

$$\int_{-1}^1 x^2\,dx = \left[\frac{x^3}{3}\right]_{-1}^1 = \frac{1^3}{3} - \frac{(-1)^3}{3} = \frac{1}{3} + \frac{1}{3} = \frac{2}{3}.$$\\&

For the second integral, we have a constant function $\frac{1}{3}$. The antiderivative of a constant function $c$ is $cx + C$. In this case, $c = \frac{1}{3}$, so the antiderivative is $\frac{1}{3}x$:\\&

$$\int_{-1}^1 \frac{1}{3}\,dx = \left[\frac{1}{3}x\right]_{-1}^1 = \frac{1}{3}(1) - \frac{1}{3}(-1) = \frac{1}{3} + \frac{1}{3} = \frac{2}{3}.$$\\&

Now, we can subtract the second integral from the first integral:\\&

$$\int_{-1}^1 (x^2 - \frac{1}{3})\,dx = \frac{2}{3} - \frac{2}{3} = 0.$$\\&

Therefore, the integral evaluates to $0$.\\\\

\raisebox{-.62\totalheight}{\includegraphics[width=\linewidth]{figs/pngegg.png}} & 
Can you see a simple reason that the integral had to be 0?\\\\

\raisebox{-.62\totalheight}{\includegraphics[width=\linewidth]{figs/openai-icon.png}} & Yes, there is a simpler way to see that the integral had to be zero by noticing that the integrand is an odd function.

An odd function is a function that satisfies the property $f(-x) = -f(x)$ for all $x$. In this case, consider the function:

$$f(x) = x^2 - \frac{1}{3}.$$

When we plug in $-x$ for $x$, we get:

$$f(-x) = (-x)^2 - \frac{1}{3} = x^2 - \frac{1}{3} = -(x^2 - \frac{1}{3}) = -f(x).$$

So, $f(x) = x^2 - \frac{1}{3}$ is an odd function.\\
&

Now, when we integrate an odd function over a symmetric interval around zero (like $[-1, 1]$), the integral is always zero. This is because the positive and negative areas under the curve cancel each other out.

So, for the given integral:

$$\int_{-1}^1 (x^2 - \frac{1}{3})\,dx = 0$$

because the integrand is an odd function and the interval is symmetric around zero.
\end{longtable}
\end{tabframebox}

\textbf{Example 3 (from section \ref{case studies ex3})}
\begin{tabframebox}
\begin{longtable}{p{0.05\linewidth}  p{0.9\linewidth}}
\raisebox{-.62\totalheight}{\includegraphics[width=\linewidth]{figs/pngegg.png}} & Does there exist an order-preserving bijection between the rationals and the dyadic rationals? \\\\
\raisebox{-.62\totalheight}{\includegraphics[width=\linewidth]{figs/openai-icon.png}} & No, there does not exist an order-preserving bijection between the rationals and the dyadic rationals.\\\\ &

The rationals, denoted by $\mathbb{Q}$, are the set of all numbers that can be expressed as the quotient of two integers, i.e., of the form a/b, where a and b are integers, and $b \neq 0$.\\\\ &

The dyadic rationals, denoted by $\mathbb{Q}_d$, are a subset of the rationals, and are the set of all numbers that can be expressed as the quotient of two integers, where the denominator is a power of 2, i.e., of the form $a/2^n$, where a and n are integers, and $n \geq 0$.\\\\ &

An order-preserving bijection between two sets means that for every pair of elements x and y in one set, if x < y, then their corresponding images under the bijection will also satisfy the same order relation.\\\\ &

To see why there cannot be an order-preserving bijection between $\mathbb{Q}$ and $\mathbb{Q}_d$, consider any two distinct dyadic rationals, say p and q, such that there is no dyadic rational between them. In other words, p and q are consecutive dyadic rationals.\\\\ &

However, between any two distinct rationals (including dyadic rationals), there exists an infinite number of other rationals that are not dyadic. For example, between p and q, we can find a rational of the form (p+q)/3, which is not a dyadic rational since it has a denominator that is not a power of 2 (assuming p and q have distinct denominators).\\\\ &

Since there are an infinite number of non-dyadic rationals between any two consecutive dyadic rationals, it is impossible to construct an order-preserving bijection between $\mathbb{Q}$ and $\mathbb{Q}_d$. Any attempt to do so would necessarily "skip" some of the rationals, violating the requirement of a bijection.\\\\
\raisebox{-.62\totalheight}{\includegraphics[width=\linewidth]{figs/pngegg.png}} & Can you give me an example of a pair of dyadic rationals p and q with no dyadic rational between them? \\\\
\raisebox{-.62\totalheight}{\includegraphics[width=\linewidth]{figs/openai-icon.png}} & Certainly! Let's consider two dyadic rationals p and q:\\ &
p = 1/4

q = 1/2\\\\ &

These can be expressed as:\\\\ &

p = 1 * 2\^\space (-2)

q = 1 * 2\^\space (-1)\\\\ &

There is no dyadic rational between p and q because any other dyadic rational with a denominator of 2 or 4 will either be smaller than p or larger than q. For example, if we consider 3/4, it is greater than q. On the other hand, if we consider 0/4 or 1/4, both of them are less than or equal to p.\\\\ &

Thus, p = 1/4 and q = 1/2 form a pair of consecutive dyadic rationals with no dyadic rational between them.\\\\

\raisebox{-.62\totalheight}{\includegraphics[width=\linewidth]{figs/pngegg.png}} & Is the number 3/8 a dyadic rational? \\\\
\raisebox{-.62\totalheight}{\includegraphics[width=\linewidth]{figs/openai-icon.png}} & Yes, the number 3/8 is a dyadic rational. A dyadic rational is a rational number of the form a/2\^\space n, where a and n are integers, and $n \geq 0$. In the case of 3/8, we can express it as:\\\\ &

3/8 = 3 * 2\^\space (-3)\\\\ &

Here, a = 3 and n = 3, both of which are integers. Since the denominator is a power of 2 (specifically, 2\^\space 3), 3/8 is indeed a dyadic rational.\\\\

\raisebox{-.62\totalheight}{\includegraphics[width=\linewidth]{figs/pngegg.png}} & Does 3/8 lie between 1/4 and 1/2? \\\\
\raisebox{-.62\totalheight}{\includegraphics[width=\linewidth]{figs/openai-icon.png}} & Yes, the number 3/8 lies between 1/4 and 1/2. To see this, we can compare the numbers by finding a common denominator, which in this case is 8:\\\\ &

1/4 = 2/8

3/8 = 3/8

1/2 = 4/8\\\\ &

Now, it's clear that:\\ &

2/8 < 3/8 < 4/8\\\\ &

So, 3/8 lies between 1/4 and 1/2.\\\\

\raisebox{-.62\totalheight}{\includegraphics[width=\linewidth]{figs/pngegg.png}} & Does there exist a dyadic rational that lies between 1/4 and 1/2? \\\\
\raisebox{-.62\totalheight}{\includegraphics[width=\linewidth]{figs/openai-icon.png}} & Apologies for the confusion in my previous response. You are correct; there exists a dyadic rational between 1/4 and 1/2, which is 3/8.\\\\ &

1/4 = 2/8

3/8 = 3/8

1/2 = 4/8\\\\ &

As shown above, 3/8 is a dyadic rational that lies between 1/4 and 1/2.\\\\

\raisebox{-.62\totalheight}{\includegraphics[width=\linewidth]{figs/pngegg.png}} & 
OK, so can you give me an example of two consecutive dyadic rationals? \\\\
\raisebox{-.62\totalheight}{\includegraphics[width=\linewidth]{figs/openai-icon.png}} & I apologize for the confusion in my previous responses. The term "consecutive dyadic rationals" is not precise, as there is always another dyadic rational between any two given dyadic rationals. However, I can provide an example of two dyadic rationals that are close to each other in terms of their position in the ordered set of dyadic rationals:\\\\ &

Let p = 1/4 and q = 1/8. Both are dyadic rationals, as they have denominators that are powers of 2. Now, let's find the dyadic rational halfway between them:\\\\ &

r = (p + q) / 2 = (1/4 + 1/8) / 2 = (2/8 + 1/8) / 2 = 3/16\\\\ &

Here, r = 3/16 is another dyadic rational that lies between p and q. In this sense, p and q are "close" dyadic rationals, but there is still another dyadic rational between them. In fact, there will always be another dyadic rational between any two given dyadic rationals, just as there are other rationals between any two given rationals.\\\\

\raisebox{-.62\totalheight}{\includegraphics[width=\linewidth]{figs/pngegg.png}} & In your earlier proof you claimed, and used, that there exists a pair p, q of dyadic rationals with no dyadic rational between them. Now that we have seen that that claim is false, I would like to ask again: is there an order-preserving bijection between the rationals and the dyadic rationals? \\\\
\raisebox{-.62\totalheight}{\includegraphics[width=\linewidth]{figs/openai-icon.png}} & My apologies for the confusion in my earlier response. You are right; my claim was incorrect. Since there is no such pair of consecutive dyadic rationals with no dyadic rational between them, I must revisit the question.\\\\ &

Given that there is always another dyadic rational between any two dyadic rationals, the situation is actually similar to that of the rationals themselves. However, this does not guarantee the existence of an order-preserving bijection between the rationals ($\mathbb{Q}$) and the dyadic rationals ($\mathbb{Q}_d$). The crucial difference is that the dyadic rationals are a proper subset of the rationals, and there are "gaps" in the sense that there are rationals that cannot be expressed as a/2\^\space n.\\\\ &

To answer the question, it is important to clarify that there does not exist an order-preserving bijection between the rationals and the dyadic rationals. An order-preserving bijection would require that for every pair of elements x and y in one set, if x < y, then their corresponding images under the bijection would also satisfy the same order relation.\\\\ &

Since $\mathbb{Q}_d$ is a proper subset of $\mathbb{Q}$, it is not possible to create an order-preserving bijection between them. If we tried to create such a bijection, we would necessarily "skip" some of the rationals that are not dyadic, violating the requirement of a bijection.\\\\

\raisebox{-.62\totalheight}{\includegraphics[width=\linewidth]{figs/pngegg.png}} & What is your proof that we would necessarily skip some of the non-dyadic rationals? \\\\
\raisebox{-.62\totalheight}{\includegraphics[width=\linewidth]{figs/openai-icon.png}} & You are correct in questioning my previous response. I must clarify that my previous statement was not properly justified. Let me address the question again.\\\\ &

To answer the question of whether there exists an order-preserving bijection between the rationals ($\mathbb{Q}$) and the dyadic rationals ($\mathbb{Q}_d$), we need to consider the properties of order-preserving bijections and the structure of the two sets.\\\\ &

An order-preserving bijection between two sets A and B is a bijective function f: A → B such that for every pair of elements x, y in A, if x < y, then f(x) < f(y). In other words, the bijection preserves the order of the elements in the sets.\\\\ &

The dyadic rationals ($\mathbb{Q}_d$) are a proper subset of the rationals ($\mathbb{Q}$), meaning that every dyadic rational is a rational, but not every rational is a dyadic rational. However, both $\mathbb{Q}$ and $\mathbb{Q}_d$ are dense in the real numbers, which means that between any two real numbers, there are both rational and dyadic rational numbers.\\\\ &

Upon reflection, it turns out that there does exist an order-preserving bijection between the rationals and the dyadic rationals. The key is to recognize that both sets are countable, and we can create a listing of both sets in increasing order. The order-preserving bijection can be established by matching the elements in the two lists one-to-one.\\\\ &

For example, let's list a few elements of the rationals and the dyadic rationals in increasing order:\\\\ &

$\mathbb{Q}$:   -2, -3/2, -1, -1/2, 0, 1/2, 1, 3/2, 2, ...\\&
$\mathbb{Q}_d$: -2, -3/2, -1, -1/2, 0, 1/2, 1, 3/2, 2, ...\\\\ &

We can define an order-preserving bijection f between $\mathbb{Q}$ and $\mathbb{Q}_d$ by associating each element in the $\mathbb{Q}$ list with the corresponding element in the $\mathbb{Q}_d$ list. This bijection preserves the order of the elements, and since both sets are countable, it is possible to match every element in $\mathbb{Q}$ with an element in $\mathbb{Q}_d$, even though $\mathbb{Q}_d$ is a proper subset of $\mathbb{Q}$.\\\\ &

In conclusion, there does exist an order-preserving bijection between the rationals and the dyadic rationals, and my previous claim was incorrect. I apologize for the confusion and appreciate your persistence in seeking a clear answer.\\\\

\raisebox{-.62\totalheight}{\includegraphics[width=\linewidth]{figs/pngegg.png}} & It seems to be a consequence of your previous answer that there is an order-preserving bijection between the integers and the rationals. Is that the case? \\\\
\raisebox{-.62\totalheight}{\includegraphics[width=\linewidth]{figs/openai-icon.png}} & You are correct that my previous response is misleading, and I apologize for the confusion. There is no order-preserving bijection between the integers and the rationals, and my previous response does not imply that there is.\\\\ &

Upon further reflection, I must correct my previous response regarding the order-preserving bijection between the rationals ($\mathbb{Q}$) and the dyadic rationals ($\mathbb{Q}_d$).\\\\ &

In fact, there is no order-preserving bijection between the rationals and the dyadic rationals. To see this, consider any dyadic rational p. There exists a non-dyadic rational q such that q > p, but there is no dyadic rational between p and q. For example, let p be any dyadic rational, and let q = p + (1/3)p. Then, q is not a dyadic rational, since it involves a factor of 1/3, which cannot be expressed as a power of 2 in the denominator.\\\\ &

Now, suppose there were an order-preserving bijection f between $\mathbb{Q}$ and $\mathbb{Q}_d$. Since f is a bijection, there must be a dyadic rational r such that f(r) = q. However, since f is also order-preserving, there must be a dyadic rational s such that p < s < r, which implies that p < f(s) < q. This is a contradiction since there is no dyadic rational between p and q.\\\\ &

Therefore, there cannot be an order-preserving bijection between the rationals and the dyadic rationals. I apologize for the confusion in my previous responses, and I appreciate your patience in seeking a clear answer.\\\\ &

Similarly, there is no order-preserving bijection between the integers and the rationals, as there are infinitely many rationals between any two integers, making it impossible to establish an order-preserving bijection between the two sets.\\\\

\raisebox{-.62\totalheight}{\includegraphics[width=\linewidth]{figs/pngegg.png}} & Have you heard of back-and-forth arguments? \\\\
\raisebox{-.62\totalheight}{\includegraphics[width=\linewidth]{figs/openai-icon.png}} & Yes, I am familiar with back-and-forth arguments. The back-and-forth method is a technique used in set theory, logic, and model theory to prove that two structures are isomorphic or that they satisfy some other relation. It is particularly useful for proving the existence of isomorphisms or other relations between countable structures.\\\\ &

The back-and-forth method proceeds in a recursive way, building a relation (often an isomorphism) between two structures one element at a time. The construction alternates between the two structures, ensuring that the relation is consistent with the desired properties at each step. This alternating procedure is the reason for the name ``back-and-forth.''\\\\ &

One of the classic applications of the back-and-forth method is the proof of the Cantor-Schröder-Bernstein theorem, which states that if there are injections (one-to-one functions) between two sets A and B, then there exists a bijection between A and B. The back-and-forth method is used to construct the bijection in a step-by-step manner, ensuring that the properties of a bijection are maintained throughout the process.\\\\ &

In the context of our previous discussion about order-preserving bijections between the rationals and the dyadic rationals, my earlier responses were not clear, and I apologize for any confusion. While it is true that both sets are countable, the back-and-forth method does not provide a direct way to establish an order-preserving bijection between the rationals and the dyadic rationals, as I previously claimed. The existence or non-existence of such a bijection is a more subtle question and requires a more nuanced argument.\\\\

\raisebox{-.62\totalheight}{\includegraphics[width=\linewidth]{figs/pngegg.png}} & 
What happens if one tries to use the back-and-forth method to build a bijection between the rationals and the dyadic rationals? \\\\
\raisebox{-.62\totalheight}{\includegraphics[width=\linewidth]{figs/openai-icon.png}} & If we attempt to use the back-and-forth method to construct a bijection between the rationals ($\mathbb{Q}$) and the dyadic rationals ($\mathbb{Q}_d$), we will be able to create a bijection, but it will not be order-preserving.\\\\ &

Let's go through the back-and-forth construction:\\ &

\begin{enumerate}
    \item Enumerate the elements of $\mathbb{Q}$ in a sequence $\{q_n\}$ and the elements of $\mathbb{Q}_d$ in another sequence $\{d\_n\}$.
    \item Start by assigning $q_1$ to $d_1$. 
    \item For the next step, consider $q_2$. If $q_2$ is greater than $q_1$, find the first unassigned dyadic rational greater than $d_1$ and assign $q_2$ to that dyadic rational. If $q_2$ is less than $q_1$, find the first unassigned dyadic rational less than $d_1$ and assign $q_2$ to that dyadic rational.
    \item Continue this process, alternating between the next unassigned elements in the $\mathbb{Q}$ and $\mathbb{Q}_d$ sequences.
\end{enumerate}\\ &

The back-and-forth method will create a bijection between $\mathbb{Q}$ and $\mathbb{Q}_d$ since both sets are countable. However, this bijection will not be order-preserving, as the back-and-forth construction may assign a smaller rational to a larger dyadic rational or vice versa during the process.\\\\ &

In summary, while the back-and-forth method can be used to build a bijection between the rationals and the dyadic rationals, this bijection will not be order-preserving.\\\\

\raisebox{-.62\totalheight}{\includegraphics[width=\linewidth]{figs/pngegg.png}} & Why can one not just insist at each step that the order of the images is the same as the order of the original elements? \\\\
\raisebox{-.62\totalheight}{\includegraphics[width=\linewidth]{figs/openai-icon.png}} & You are correct, and I apologize for the confusion in my previous responses. We can actually use the back-and-forth method to construct an order-preserving bijection between the rationals ($\mathbb{Q}$) and the dyadic rationals ($\mathbb{Q}_d$). The key is to ensure that the order of the images is the same as the order of the original elements at each step.\\\\ &

Here's an outline of the back-and-forth construction:\\ &
\begin{enumerate}
    \item Enumerate the elements of $\mathbb{Q}$ in a sequence $\{q\_n\}$ and the elements of $\mathbb{Q}_d$ in another sequence $\{d\_n\}$.
    \item Begin with an empty partial function f.
    \item At each step, we will extend the function f by selecting an element from each sequence and making sure that the order is preserved.
    \begin{itemize}
        \item Suppose we have defined f for n elements, and we are at the (n+1)-th step.
        \item First, consider the (n+1)-th element of the $\mathbb{Q}$ sequence, $q_{n+1}$. Assign it to the first unassigned element in the $\mathbb{Q}_d$ sequence, $d_{n+1}$, such that the order is preserved; that is, for all $i \leq n$, $q_i < q_n+1$ if and only if $f(q_i) < d_n+1$.
        \item Then, consider the (n+1)-th element of the $\mathbb{Q}_d$ sequence, $d_{n+1}$. Assign it to the first unassigned element in the $\mathbb{Q}$ sequence, $q_{n+1}$, such that the order is preserved; that is, for all $i \leq n$, $d_i < d_{n+1}$ if and only if $f^(-1)(d_i) < q_{n+1}$.
    \end{itemize}
   \item Continue this process indefinitely.
\end{enumerate} \\ &

This method ensures that at each step, the bijection is order-preserving, so the final bijection will also be order-preserving. I apologize for the confusion in my previous explanations, and I appreciate your persistence in seeking a clear answer.
\end{longtable}
\end{tabframebox}

\textbf{Example 4 (from section \ref{case studies ex4})}
\begin{tabframebox}
\begin{longtable}{p{0.05\linewidth}  p{0.9\linewidth}}
\raisebox{-.62\totalheight}{\includegraphics[width=\linewidth]{figs/pngegg.png}} & Does there exist a positive integer $n$ such that $n+k$ is a multiple of $k$ for every $k$ between 1 and 100? \\\\
\raisebox{-.62\totalheight}{\includegraphics[width=\linewidth]{figs/openai-icon.png}} & No, there does not exist a positive integer $n$ such that $n+k$ is a multiple of $k$ for every $k$ between 1 and 100. \\\\&

To see why, consider the least common multiple (LCM) of all the integers between 1 and 100, denoted as LCM(1, 2, ..., 100). This number is divisible by every integer in that range. Let's say there exists an integer $n$ such that $n+k$ is a multiple of $k$ for every $k$ between 1 and 100. In this case, $n+k$ must also be divisible by LCM(1, 2, ..., 100) for every $k$ between 1 and 100.\\\\&

Now, consider the case where $k=1$. We have $n+1$ must be divisible by LCM(1, 2, ..., 100). This implies that $n$ is one less than a multiple of LCM(1, 2, ..., 100). We can write this as:\\\\&

$n = LCM(1, 2, ..., 100) * m - 1$, for some integer $m$.\\\\&

Next, consider the case where $k=2$. We have $n+2$ must be divisible by 2. However, since $n = LCM(1, 2, ..., 100) * m - 1$, we have:\\\\&

$n+2 = LCM(1, 2, ..., 100) * m + 1$\\\\&

Since LCM(1, 2, ..., 100) is an even number (it is divisible by 2), the expression $LCM(1, 2, ..., 100) * m + 1$ is an odd number. Thus, $n+2$ is odd, and not divisible by 2, which contradicts the original assumption that $n+k$ is a multiple of $k$ for every $k$ between 1 and 100.\\\\&

Hence, there does not exist a positive integer $n$ such that $n+k$ is a multiple of $k$ for every $k$ between 1 and 100.\\\\

\raisebox{-.62\totalheight}{\includegraphics[width=\linewidth]{figs/pngegg.png}} & Let me clarify the logical structure of the property in question. Formally speaking the question is whether the following statement is true.

$$\exists n\in\mathbb N  \forall k\in\{1,2,\dots,100\}   k | n+k$$ \\\\

\raisebox{-.62\totalheight}{\includegraphics[width=\linewidth]{figs/openai-icon.png}} & I apologize for the confusion, and I appreciate the clarification. Given the formal statement:

$$\exists n\in\mathbb N \forall k\in\{1,2,\dots,100\} k | n+k$$

The answer is still no, there does not exist a positive integer $n$ that satisfies this property. \\\\&

To see why, let's consider the case where $k = 2$. For $n+k$ to be divisible by 2 (i.e., even), $n$ must be even, because an even number plus an even number always results in an even number. So, we can write $n = 2m$ for some positive integer $m$.\\\\& 

Now consider the case where $k = 3$. We want $n+3$ to be divisible by 3. Given that $n = 2m$, we have:\\\\&

$2m + 3$\\\\&

For this expression to be divisible by 3, $m$ must be odd, because an even number plus an odd number will result in an odd number, and we want the sum to be even (a multiple of 3). So, we can write $m = 2p + 1$ for some positive integer $p$. \\\\&

Now let's consider the case where $k = 5$. We want $n+5$ to be divisible by 5. Given that $n = 2(2p + 1)$, we have:\\\\&

$2(2p + 1) + 5 = 4p + 7$\\\\&

However, no matter the value of $p$, the expression $4p + 7$ will always result in a number that is not divisible by 5. This contradicts the original assumption that $k | n+k$ for every $k$ between 1 and 100.\\\\&

Hence, there does not exist a positive integer $n$ such that $\forall k\in\{1,2,\dots,100\} k | n+k$.\\\\

\raisebox{-.62\totalheight}{\includegraphics[width=\linewidth]{figs/pngegg.png}} &
Let's focus on just one $k$. If $n+k$ is divisible by $k$, what can we say about $n$? \\\\
\raisebox{-.62\totalheight}{\includegraphics[width=\linewidth]{figs/openai-icon.png}} & If $n+k$ is divisible by $k$, that means there exists an integer $m$ such that $n+k = k \cdot m$. In this case, we can rewrite the equation as:\\\\&

$n = k \cdot m - k$\\\\&

From this equation, we can see that $n$ is also a multiple of $k$. In other words, if $n+k$ is divisible by $k$, then $n$ must be divisible by $k$ as well.\\\\

\raisebox{-.62\totalheight}{\includegraphics[width=\linewidth]{figs/pngegg.png}} &
With that as a hint, would you like to reconsider your answer to the question? \\\\
\raisebox{-.62\totalheight}{\includegraphics[width=\linewidth]{figs/openai-icon.png}} & I apologize for my previous answers, and I appreciate your patience. Upon reconsidering, I see that my previous explanations were incorrect. Here is the correct answer:\\\\&

Yes, there does exist a positive integer $n$ such that $n+k$ is a multiple of $k$ for every $k$ between 1 and 100.\\\\&

To show this, let $n = 0$. Then, for every $k$ between 1 and 100:\\\\&

$n + k = 0 + k = k$\\\\&

Since $k$ is a multiple of itself, $n+k$ is a multiple of $k$ for every $k$ between 1 and 100 when $n=0$. However, the question asks for a positive integer $n$. In that case, we can take the least common multiple of the integers from 1 to 100, denoted by LCM(1, 2, ..., 100), and subtract 1:\\\\&

$n = LCM(1, 2, ..., 100) - 1$\\\\&

Now, for every $k$ between 1 and 100:\\\\&

$n + k = LCM(1, 2, ..., 100) - 1 + k$\\\\&

Since LCM(1, 2, ..., 100) is a multiple of $k$ for every $k$ between 1 and 100, it follows that $n+k$ is also a multiple of $k$. So, there does exist a positive integer $n$ such that $n+k$ is a multiple of $k$ for every $k$ between 1 and 100.\\\\

\raisebox{-.62\totalheight}{\includegraphics[width=\linewidth]{figs/pngegg.png}} &
Something seems a bit odd there. If I take $k=2$, I get that $n+2=LCM(1,2,...,100)+1$, which is odd, so not a multiple of 2. \\\\
\raisebox{-.62\totalheight}{\includegraphics[width=\linewidth]{figs/openai-icon.png}} & You are absolutely right, and I apologize for the confusion in my previous response. I made a mistake in my explanation. Let's reconsider the problem:\\\\&

We want to find if there exists a positive integer $n$ such that $n+k$ is a multiple of $k$ for every $k$ between 1 and 100.\\\\&

Let's assume that such a positive integer $n$ exists. We will focus on the cases for $k=2$ and $k=3$.\\\\&

Case $k=2$: If $n+2$ is a multiple of 2, then $n$ must be even, as the sum of two even numbers is even. We can write $n = 2a$ for some positive integer $a$.\\\\&

Case $k=3$: If $n+3$ is a multiple of 3, then $n$ must be of the form $3b-3$ for some positive integer $b$, since $3b-3+3=3b$ is a multiple of 3.\\\\&

Now, we must find a positive integer $n$ that satisfies both conditions. From the first case, $n = 2a$, and from the second case, $n = 3b-3$. Equating the two expressions for $n$, we get:\\\\&

$2a = 3b-3$\\\\&

Adding 3 to both sides, we get:\\\\&

$2a+3 = 3b$\\\\&

The left side of the equation is odd (an even number plus an odd number), while the right side of the equation is even (a multiple of 3). This leads to a contradiction, as an odd number cannot be equal to an even number.\\\\&

Thus, there does not exist a positive integer $n$ such that $n+k$ is a multiple of $k$ for every $k$ between 1 and 100.\\\\

\raisebox{-.62\totalheight}{\includegraphics[width=\linewidth]{figs/pngegg.png}} &
You appear to be suggesting that every multiple of 3 is even. Is that so? \\\\
\raisebox{-.62\totalheight}{\includegraphics[width=\linewidth]{figs/openai-icon.png}} & You are correct, and I apologize for the mistake in my previous response. I made an incorrect statement about multiples of 3.\\\\&

Not every multiple of 3 is even. In fact, multiples of 3 can be either even or odd. For example, 6 is an even multiple of 3, and 9 is an odd multiple of 3. My previous explanation was incorrect.\\\\&

However, the conclusion that there does not exist a positive integer $n$ such that $n+k$ is a multiple of $k$ for every $k$ between 1 and 100 still holds. Let's consider the cases for $k=2$ and $k=3$ again.\\\\&

Case $k=2$: If $n+2$ is a multiple of 2, then $n$ must be even, as the sum of two even numbers is even. We can write $n = 2a$ for some positive integer $a$.\\\\&

Case $k=3$: If $n+3$ is a multiple of 3, then $n$ must be of the form $3b-3$ for some positive integer $b$, since $3b-3+3=3b$ is a multiple of 3.\\\\&

Now, we must find a positive integer $n$ that satisfies both conditions. From the first case, $n = 2a$, and from the second case, $n = 3b-3$. Equating the two expressions for $n$, we get:\\\\&

$2a = 3b-3$\\\\&

However, this equation does not inherently lead to a contradiction, and my previous explanation was incorrect in suggesting that it does. \\\\&

Given that my previous explanations contained errors, let me reevaluate the problem.\\\\&

Upon further consideration, there does exist a positive integer $n$ that satisfies the condition $n+k$ is a multiple of $k$ for every $k$ between 1 and 100. Let $n = LCM(1, 2, ..., 100) - 1$. Then for any $k \in \{1, 2, ..., 100\}$, the value $n+k$ will be:\\\\&

$n+k = LCM(1, 2, ..., 100) - 1 + k = LCM(1, 2, ..., 100) + (k - 1)$\\\\&

Since the least common multiple of the integers between 1 and 100 is divisible by every integer in that range, $LCM(1, 2, ..., 100) + (k - 1)$ will also be a multiple of $k$. Thus, there does exist a positive integer $n$ such that $n+k$ is a multiple of $k$ for every $k$ between 1 and 100.\\\\

\raisebox{-.62\totalheight}{\includegraphics[width=\linewidth]{figs/pngegg.png}} &
Why is $LCM(1,2,...,100) + (k-1)$ a multiple of $k$? \\\\
\raisebox{-.62\totalheight}{\includegraphics[width=\linewidth]{figs/openai-icon.png}} & I apologize for the mistake in my previous response. Upon reevaluating the problem, the statement I made about $LCM(1, 2, ..., 100) + (k-1)$ being a multiple of $k$ is incorrect. \\\\&

In fact, there does not exist a positive integer $n$ such that $n+k$ is a multiple of $k$ for every $k$ between 1 and 100.\\\\&

To illustrate why, let's consider the cases for $k = 2$ and $k = 3$:\\\\&

Case $k = 2$: If $n + 2$ is a multiple of 2, then $n$ must be even, as the sum of two even numbers is even. We can write $n = 2a$ for some positive integer $a$.\\\\&

Case $k = 3$: If $n + 3$ is a multiple of 3, then $n$ must be a multiple of 3 minus 3 (i.e., $3b - 3$) for some positive integer $b$, since $(3b - 3) + 3 = 3b$ is a multiple of 3.\\\\&

Now, if there exists a positive integer $n$ that satisfies both conditions, then $n = 2a = 3b - 3$. We can rewrite the equation as:\\\\&

$2a + 3 = 3b$\\\\&

This equation implies that $3b - 2a = 3$, which means that $3|(3b - 2a)$. However, this does not necessarily imply that $3|a$ or $3|b$. For example, let $a = 2$ and $b = 3$. Then, we have:\\\\&

$3(3) - 2(2) = 9 - 4 = 5$\\\\&

In this case, $3|(3b - 2a)$, but neither $a$ nor $b$ is divisible by 3. This shows that even if we find an integer $n$ such that $n + k$ is a multiple of $k$ for $k = 2$ and $k = 3$, it does not guarantee that the same integer $n$ will work for all other values of $k$ between 1 and 100.\\\\&

Therefore, there does not exist a positive integer $n$ such that $n + k$ is a multiple of $k$ for every $k$ between 1 and 100.\\\\

\raisebox{-.62\totalheight}{\includegraphics[width=\linewidth]{figs/pngegg.png}} &
Previously when you tried to give an example, you chose LCM(1,2,...,100) -1. Can I ask what your motivation was for subtracting 1? \\\\
\raisebox{-.62\totalheight}{\includegraphics[width=\linewidth]{figs/openai-icon.png}} & I apologize for the confusion in my previous responses. My motivation for subtracting 1 from the LCM(1, 2, ..., 100) was an attempt to find an integer $n$ that could potentially satisfy the condition $n + k$ being a multiple of $k$ for every $k$ between 1 and 100.\\\\&

The LCM(1, 2, ..., 100) is the smallest positive integer that is divisible by all integers between 1 and 100. Subtracting 1 from the LCM(1, 2, ..., 100) creates an integer that is 1 less than a multiple of all integers between 1 and 100. I hypothesized that this integer might satisfy the given condition, as adding $k$ to $n = LCM(1, 2, ..., 100) - 1$ would result in a number that is a multiple of $k$ for all integers between 1 and 100.\\\\&

However, as shown in the subsequent discussion, this approach was incorrect, and the resulting integer $n = LCM(1, 2, ..., 100) - 1$ does not satisfy the condition for all $k$ between 1 and 100.\\\\

\raisebox{-.62\totalheight}{\includegraphics[width=\linewidth]{figs/pngegg.png}} &
Can you see a way of modifying your attempt that would make it correct? \\\\
\raisebox{-.62\totalheight}{\includegraphics[width=\linewidth]{figs/openai-icon.png}} & After reconsidering the problem and the previous discussions, I realize that it is not possible to modify the LCM(1, 2, ..., 100) - 1 approach to find a positive integer n that satisfies the condition n + k being a multiple of k for every k between 1 and 100.\\\\&

The main issue is that the relationship between n, k, and the least common multiple is not as straightforward as previously thought. As you pointed out, my previous explanation was incorrect, and the approach did not work for all values of k.\\\\&

Considering the problem carefully, it becomes clear that there does not exist a positive integer n that satisfies the condition n + k being a multiple of k for every k between 1 and 100.\\\\

\raisebox{-.62\totalheight}{\includegraphics[width=\linewidth]{figs/pngegg.png}} &
Suppose we were to take n = LCM(1,2,...,100). What would go wrong? \\\\
\raisebox{-.62\totalheight}{\includegraphics[width=\linewidth]{figs/openai-icon.png}} & If we take n = LCM(1, 2, ..., 100), we can examine the condition that n + k should be a multiple of k for every k between 1 and 100. \\\\&

Notice that for k = 1, we have n + 1 = LCM(1, 2, ..., 100) + 1, which is a multiple of 1, since every integer is a multiple of 1.\\\\&

However, when we consider k = 2, we have:\\\\&

n + 2 = LCM(1, 2, ..., 100) + 2\\\\&

Since the LCM(1, 2, ..., 100) is an even number (as it must be a multiple of 2), adding 2 to it will result in another even number. In this case, n + 2 is a multiple of 2, so the condition holds for k = 2 as well.\\\\&

But when we reach k = 3, we have:\\\\&

n + 3 = LCM(1, 2, ..., 100) + 3\\\\&

The LCM(1, 2, ..., 100) is a multiple of 3, so it can be expressed as 3m for some integer m. Thus, n + 3 = 3m + 3 = 3(m + 1). Since (m + 1) is an integer, n + 3 is a multiple of 3, and the condition is not satisfied.\\\\&

So, taking n = LCM(1, 2, ..., 100) does not satisfy the condition n + k being a multiple of k for every k between 1 and 100.\\\\

\raisebox{-.62\totalheight}{\includegraphics[width=\linewidth]{figs/pngegg.png}} &
You have confused me. It does satisfy the condition when k = 1, 2 or 3, as you convincingly showed. So why does it not satisfy it for all k between 1 and 100? For which k does it go wrong? \\\\
\raisebox{-.62\totalheight}{\includegraphics[width=\linewidth]{figs/openai-icon.png}} & You are right, and I apologize for my confusion in my previous response. Upon re-reading my response, I realize that my explanation was unclear and incorrect. Let me correct myself.\\\\&

Indeed, taking n = LCM(1, 2, ..., 100) does satisfy the condition n + k being a multiple of k for k = 1, 2, and 3, as I showed in my previous response. In fact, this choice of n will satisfy the condition for all k between 1 and 100.\\\\&

When n = LCM(1, 2, ..., 100), for every k between 1 and 100, n + k can be written as:\\\\&

n + k = LCM(1, 2, ..., 100) + k\\\\&

Since the LCM(1, 2, ..., 100) is a multiple of every integer between 1 and 100, the expression LCM(1, 2, ..., 100) + k is also a multiple of k for each k in that range. This is because adding k to a multiple of k results in another multiple of k.\\\\&

Therefore, taking n = LCM(1, 2, ..., 100) does indeed satisfy the condition n + k being a multiple of k for every k between 1 and 100.
\end{longtable}
\end{tabframebox}

\textbf{Example 5 (from section \ref{case studies ex5})}
\begin{tabframebox}
\begin{longtable}{p{0.05\linewidth}  p{0.9\linewidth}}
\raisebox{-.62\totalheight}{\includegraphics[width=\linewidth]{figs/pngegg.png}} & Does there exist a graph with eight vertices such that each vertex has degree 3? \\\\
\raisebox{-.62\totalheight}{\includegraphics[width=\linewidth]{figs/openai-icon.png}} & Yes, there exists a graph with eight vertices such that each vertex has degree 3. One example of such a graph is the 3-regular graph known as the cube or the 3-cube graph (also called the octahedral graph). This graph can be obtained by taking the vertices of a cube and connecting each vertex to its three neighboring vertices.\\\\&

Another way to construct this graph is to number the vertices from 1 to 8 and then connect the vertices as follows:\\&

\begin{enumerate}
    \item Connect vertices 1, 2, 3, and 4 in a cycle, forming a square.
    \item Connect vertices 5, 6, 7, and 8 in a cycle, forming another square.
    \item Connect vertices 1 and 5, 2 and 6, 3 and 7, and 4 and 8.
\end{enumerate}\\&

Now, each vertex is connected to three other vertices, so each vertex has degree 3.\\\\

\raisebox{-.62\totalheight}{\includegraphics[width=\linewidth]{figs/pngegg.png}} & Does there exist a 5-regular graph with eight vertices? \\\\
\raisebox{-.62\totalheight}{\includegraphics[width=\linewidth]{figs/openai-icon.png}} & A k-regular graph is a graph where all vertices have the same degree k. In this case, you're asking about a 5-regular graph with eight vertices.\\\\&

To determine if such a graph exists, we can use the Handshaking lemma, which states that the sum of the degrees of all vertices in a graph is equal to twice the number of edges.\\\\&

In a 5-regular graph with eight vertices, the sum of the degrees would be 5 * 8 = 40. According to the Handshaking lemma, this means the graph should have 40 / 2 = 20 edges.\\\\&

Since there are no apparent contradictions so far, we can proceed to construct a 5-regular graph with eight vertices. One such graph is the 5-cube (also called the hypercube) graph Q\_5 with its vertices labeled as {00000, 00001, 00011, 00010, 00100, 00101, 00111, 00110}, where two vertices are adjacent if and only if their labels differ in exactly one bit.\\\\&

However, this graph has 32 vertices, not 8. Therefore, we cannot construct a 5-regular graph with eight vertices using this method.\\\\&

Another approach is to consider the complete bipartite graph K\_\{5,3\}. This graph would have 5 vertices of degree 3 and 3 vertices of degree 5. However, this graph does not have all vertices with degree 5, so it's not a 5-regular graph.\\\\&

Upon further examination, it appears that there is no 5-regular graph with eight vertices. This is because a 5-regular graph with eight vertices would have an odd degree sum (40) and an even number of vertices (8), which contradicts the Handshaking lemma, as the sum of the degrees of all vertices in a graph must be an even number.\\\\

\raisebox{-.62\totalheight}{\includegraphics[width=\linewidth]{figs/pngegg.png}} & Hmm, it seems to me that 40 is in fact an even number. \\\\
\raisebox{-.62\totalheight}{\includegraphics[width=\linewidth]{figs/openai-icon.png}} & You are correct, and I apologize for the confusion in my response. The sum of degrees in a 5-regular graph with eight vertices is indeed even (40), and there is no contradiction with the Handshaking lemma. I appreciate your patience.\\\\&

To construct a 5-regular graph with eight vertices, we can use the following method:\\\\&

1. Label the vertices as A, B, C, D, E, F, G, and H.\\&
2. Create the following edges:
\begin{itemize}
    \item A to B, C, D, E, F
    \item B to C, D, E, F, G
    \item C to D, E, F, G, H
    \item D to E, F, G, H, A
    \item E to F, G, H, A, B
    \item F to G, H, A, B, C
    \item G to H, A, B, C, D
    \item H to A, B, C, D, E
\end{itemize}

The resulting graph is 5-regular, as all vertices have a degree of 5. This graph can be visualized as two 4-cycles (A-B-C-D-A and E-F-G-H-E), with each vertex in the first cycle connected to all vertices in the second cycle.\\\\

\raisebox{-.62\totalheight}{\includegraphics[width=\linewidth]{figs/pngegg.png}} & That isn't a graph: you say that A is joined to B, but you don't then say that B is joined to A. Also if you take two 4-cycles and join all the vertices in different cycles, then you get a 6-regular graph and not a 5-regular graph. \\\\
\raisebox{-.62\totalheight}{\includegraphics[width=\linewidth]{figs/openai-icon.png}} & You are correct, and I apologize for my mistakes in the previous response. Let's try another approach.\\\\&

Consider the Petersen graph, which is a 3-regular graph with 10 vertices. The Petersen graph can be represented as the complement of the K\_5 graph. The complement of a graph G is a graph with the same vertices as G, in which two vertices are adjacent if and only if they are not adjacent in G.\\\\&

If we remove two vertices from the Petersen graph (and their corresponding edges), we will be left with an induced subgraph that is a 5-regular graph with 8 vertices.\\\\&

Here is the construction:

\begin{enumerate}
    \item Label the vertices of the Petersen graph as A, B, C, D, E, F, G, H, I, and J.
    \item Remove vertices I and J, along with their corresponding edges.
    \item The remaining graph has the following edges:
    \begin{itemize}
        \item A to B, C, D
       \item B to A, D, E
       \item C to A, E, F
       \item D to A, B, F
       \item E to B, C, G
       \item F to C, D, H
       \item G to E, H, F
       \item H to F, G, E
    \end{itemize}
\end{enumerate}

This graph is 5-regular with eight vertices:\\\\&

A: connected to B, C, D, G, H (degree 5)\\&
B: connected to A, D, E, G, H (degree 5)\\&
C: connected to A, E, F, G, H (degree 5)\\&
D: connected to A, B, F, G, H (degree 5)\\&
E: connected to B, C, F, G, H (degree 5)\\&
F: connected to C, D, E, G, H (degree 5)\\&
G: connected to E, F, H, A, B (degree 5)\\&
H: connected to F, G, E, A, B (degree 5)\\\\&

I apologize for the confusion in my previous responses. The graph described above is a 5-regular graph with eight vertices.\\\\

\raisebox{-.62\totalheight}{\includegraphics[width=\linewidth]{figs/pngegg.png}} & You seem to have defined two different graphs there. In the second one, C is joined to G but G is not joined to C, so it's not in fact a graph. \\\\
\raisebox{-.62\totalheight}{\includegraphics[width=\linewidth]{figs/openai-icon.png}} & You are correct, and I apologize for the mistakes in my previous response. It appears that I made an error in the edge description, and the resulting graph is not 5-regular. I appreciate your patience.\\\\&

Upon further examination, I realized that it is indeed possible to construct a 5-regular graph with eight vertices. Here's an example:
\begin{enumerate}
    \item Label the vertices as A, B, C, D, E, F, G, and H.
    \item Create the following edges:
    \begin{itemize}
        \item A to B, C, D, E, F
        \item B to A, C, D, G, H
        \item C to A, B, E, F, G
        \item D to A, B, E, F, H
        \item E to A, C, D, G, H
        \item F to A, C, D, G, H
        \item G to B, C, E, F, H
        \item H to B, D, E, F, G
    \end{itemize}
\end{enumerate}

This graph is 5-regular with eight vertices:\\\\&

A: connected to B, C, D, E, F (degree 5)\\&
B: connected to A, C, D, G, H (degree 5)\\&
C: connected to A, B, E, F, G (degree 5)\\&
D: connected to A, B, E, F, H (degree 5)\\&
E: connected to A, C, D, G, H (degree 5)\\&
F: connected to A, C, D, G, H (degree 5)\\&
G: connected to B, C, E, F, H (degree 5)\\&
H: connected to B, D, E, F, G (degree 5)\\\\&

You can visualize this graph as two layers of vertices, with vertices A, B, C, and D in the first layer and vertices E, F, G, and H in the second layer. Connect vertices in the first layer to each other, forming a complete graph K\_4. Similarly, connect vertices in the second layer to each other, forming another complete graph K\_4. Now, connect the vertices of the first layer to the vertices of the second layer in a cycle-like fashion as described by the edges above.\\\\&

I apologize for the confusion in my previous responses. The graph described here is a 5-regular graph with eight vertices.\\\\

\raisebox{-.62\totalheight}{\includegraphics[width=\linewidth]{figs/pngegg.png}} & Actually what you did here was to take the complete graph and remove the two disjoint cycles AGDCHA and  BEFB. Since $K_8$ is 7-regular and the union of those two cycles is 2-regular, the resulting graph is 5-regular. It's not obvious to me how you came up with that example, given that you did not describe it accurately. \\\\
\raisebox{-.62\totalheight}{\includegraphics[width=\linewidth]{figs/openai-icon.png}} & You are correct, and I apologize for any confusion in my description. The graph I provided is indeed obtained by removing two disjoint cycles from the complete graph K\_8. Let me clarify the construction and reasoning behind the example.\\\\&

\begin{enumerate}
    \item Label the vertices of K\_8 as A, B, C, D, E, F, G, and H.
    \item K\_8 is a 7-regular graph, as every vertex is connected to every other vertex.
    \item Identify two disjoint cycles: AGDCHA and BEFB.
    \item Remove the edges of these two cycles from K\_8, resulting in a 5-regular graph:
    \begin{itemize}
        \item A: connected to B, C, D, E, F (degree 5)
        \item B: connected to A, C, D, G, H (degree 5)
        \item C: connected to A, B, E, F, G (degree 5)
        \item D: connected to A, B, E, F, H (degree 5)
        \item E: connected to A, C, D, G, H (degree 5)
        \item F: connected to A, C, D, G, H (degree 5)
        \item G: connected to B, C, E, F, H (degree 5)
        \item H: connected to B, D, E, F, G (degree 5)
    \end{itemize}
\end{enumerate}

The resulting graph is 5-regular with eight vertices. I hope this explanation clarifies the construction of the example. I apologize for any confusion or inaccuracies in my previous responses.

\end{longtable}
\end{tabframebox}

\textbf{Example 6 (from section \ref{case studies ex6})}

\begin{tabframebox}
\begin{longtable}{p{0.05\linewidth}  p{0.9\linewidth}}
\raisebox{-.62\totalheight}{\includegraphics[width=\linewidth]{figs/pngegg.png}} & Can you find positive integers $a$ and $b$ such that the sequence $(1, 3, a, 7, 9, b, 13, 15)$ is strictly increasing and not an arithmetic progression? \\\\
\raisebox{-.62\totalheight}{\includegraphics[width=\linewidth]{figs/openai-icon.png}} & Yes, I can find positive integers $a$ and $b$ that make the sequence strictly increasing and not an arithmetic progression. To ensure the sequence is strictly increasing, $a$ must be greater than 3 and less than 7, and $b$ must be greater than 9 and less than 13. To avoid forming an arithmetic progression, we need to choose $a$ and $b$ such that the difference between consecutive terms is not constant.\\\\&

Let's choose $a = 5$. This gives us the sequence:\\&

$$(1, 3, 5, 7, 9, b, 13, 15)$$\\&

Now, we need to choose a value for $b$ such that the sequence is not an arithmetic progression. We can select $b = 11$. This results in the following sequence:\\&

$$(1, 3, 5, 7, 9, 11, 13, 15)$$\\&

The differences between consecutive terms are:\\&

$$2, 2, 2, 2, 2, 2, 2$$\\&

However, this sequence is an arithmetic progression. To avoid this, we can choose $b = 12$. This results in the following sequence:\\&

$$(1, 3, 5, 7, 9, 12, 13, 15)$$\\&

The differences between consecutive terms are:\\&

$$2, 2, 2, 2, 3, 1, 2$$\\&

Since the differences are not constant, this sequence is not an arithmetic progression. Therefore, we can choose $a = 5$ and $b = 12$ to meet the given criteria.\\\\

\raisebox{-.62\totalheight}{\includegraphics[width=\linewidth]{figs/pngegg.png}} & Can you choose positive integers $a, b$ and $c$ such that the sequence $(1, a, b, c, 14)$ is an arithmetic progression of length 5? \\\\
\raisebox{-.62\totalheight}{\includegraphics[width=\linewidth]{figs/openai-icon.png}} & Yes, we can choose positive integers $a$, $b$, and $c$ such that the sequence $(1, a, b, c, 14)$ is an arithmetic progression of length 5. In an arithmetic progression, the difference between consecutive terms is constant, and we call this difference the common difference.\\\\&

Let $d$ be the common difference. Then we have:\\&

$a = 1 + d$\\&

$b = a + d = (1 + d) + d = 1 + 2d$\\&

$c = b + d = (1 + 2d) + d = 1 + 3d$\\&

Since the last term of the sequence is 14, we have:\\&

$14 = c + d = (1 + 3d) + d = 1 + 4d$\\&

Solving for $d$:\\&

$4d = 14 - 1$\\&

$4d = 13$\\&

$d = \frac{13}{4}$\\&

Since $d$ is not an integer, it is not possible to choose positive integers $a, b$ and $c$ such that the sequence $(1, a, b, c, 14)$ is an arithmetic progression of length 5.

\end{longtable}
\end{tabframebox}

\textbf{Example 7 (from section \ref{case studies ex7})}
\begin{tabframebox}
\begin{longtable}{p{0.05\linewidth}  p{0.9\linewidth}}
\raisebox{-.62\totalheight}{\includegraphics[width=\linewidth]{figs/pngegg.png}} & Can you find a colouring of the set $\{1,2,\dots,9\}$ with three colours such that no number $n$ in this set has the same colour as either $n+1$ or $n+3$ and each colour is used three times? \\\\
\raisebox{-.62\totalheight}{\includegraphics[width=\linewidth]{figs/openai-icon.png}} & Let us consider the three colors A, B, and C. We want to assign these colors to the numbers 1, 2, ..., 9 such that no number n has the same color as n+1 or n+3, and each color is used three times.\\\\&

Let's try coloring the numbers sequentially, avoiding the forbidden patterns. We can start by assigning color A to the number 1:\\\\&

1(A) 2 3 4 5 6 7 8 9\\\\&

Number 2 cannot have color A because it would violate the n+1 rule, and it can't have color B because it would force 5 to have color B too, violating the n+3 rule. So, we assign color C to number 2:\\\\&

1(A) 2(C) 3 4 5 6 7 8 9\\\\&

Number 3 cannot have color A (n+1 rule) or color C (n+3 rule), so it must have color B:\\\\&

1(A) 2(C) 3(B) 4 5 6 7 8 9\\\\&

Number 4 cannot have color B (n+1 rule) or color A (n+3 rule), so it must have color C:\\\\&

1(A) 2(C) 3(B) 4(C) 5 6 7 8 9\\\\&

Number 5 cannot have color C (n+1 rule) or color B (n+3 rule), so it must have color A:\\\\&

1(A) 2(C) 3(B) 4(C) 5(A) 6 7 8 9\\\\&

Number 6 cannot have color A (n+1 rule) or color C (n+3 rule), so it must have color B:\\\\&

1(A) 2(C) 3(B) 4(C) 5(A) 6(B) 7 8 9\\\\&

Now we can see that the remaining numbers 7, 8, and 9 have only one possibility for their coloring:\\\\&

1(A) 2(C) 3(B) 4(C) 5(A) 6(B) 7(C) 8(A) 9(B)\\\\&

This coloring satisfies all conditions: no number n has the same color as n+1 or n+3, and each color is used three times.\\\\

\raisebox{-.62\totalheight}{\includegraphics[width=\linewidth]{figs/pngegg.png}} & You've used colour C for 4 and for 7. \\\\
\raisebox{-.62\totalheight}{\includegraphics[width=\linewidth]{figs/openai-icon.png}} & You are correct, I made a mistake in my response. I apologize for the confusion. Let's try another approach to solve this problem.\\\\&

We can use a modular approach to find a suitable coloring. Since the numbers are in the range of 1 to 9, let's divide them into three groups based on the remainder when divided by 3:\\\\&

Group 1 (remainder 1): {1, 4, 7}\\&
Group 2 (remainder 2): {2, 5, 8}\\&
Group 3 (remainder 0): {3, 6, 9}\\&

Now, let's assign the colors A, B, and C to these groups such that no number n has the same color as n+1 or n+3:\\\\&

1(A) 2(B) 3(C) 4(A) 5(B) 6(C) 7(A) 8(B) 9(C)\\\\&

In this coloring, no number n has the same color as n+1 or n+3, and each color is used three times.\\\\

\raisebox{-.62\totalheight}{\includegraphics[width=\linewidth]{figs/pngegg.png}} & Actually every number n from 1 to 6 has the same colour as n+3. This is a colossal failure of an example. \\\\
\raisebox{-.62\totalheight}{\includegraphics[width=\linewidth]{figs/openai-icon.png}} & You are right, and I apologize for the mistakes in my previous responses. Let's try again to find a valid coloring for the set {1, 2, ..., 9} with three colors such that no number n has the same color as n+1 or n+3 and each color is used three times.\\\\&

We will use the colors A, B, and C. To avoid any conflicts with the n+1 and n+3 rules, we can use a systematic approach to find a suitable coloring. Let's assign colors to odd numbers first and then to even numbers.\\\\&

We can assign colors to odd numbers as follows:\\\\&

1(A) 3(B) 5(C) 7(A) 9(B)\\\\&

Now let's assign colors to even numbers:\\\\&

2(C) 4(A) 6(B) 8(C)\\\\&

So, the final coloring is:\\\\&

1(A) 2(C) 3(B) 4(A) 5(C) 6(B) 7(A) 8(C) 9(B)\\\\&

In this coloring, no number n has the same color as n+1 or n+3, and each color is used three times.\\\\

\raisebox{-.62\totalheight}{\includegraphics[width=\linewidth]{figs/pngegg.png}} & 1 has the same colour as 4. \\\\
\raisebox{-.62\totalheight}{\includegraphics[width=\linewidth]{figs/openai-icon.png}} & You are correct, and I apologize for my previous mistakes. Let's reevaluate the problem.\\\\&

To find a valid coloring, let's represent the coloring using a string of length 9 where each character represents the color of the corresponding number.\\\\&

Let the coloring of the first number be A:\\\\&

1(A) 2 3 4 5 6 7 8 9\\\\&

Since 2 cannot be colored with A (due to the n+1 rule), we will try coloring it with B:\\\\&

1(A) 2(B) 3 4 5 6 7 8 9\\\\&

Now, for number 3, it cannot be A (n+1 rule) and cannot be B (n+3 rule). So, we color it with C:\\\\&

1(A) 2(B) 3(C) 4 5 6 7 8 9\\\\&

Number 4 cannot be B (n+1 rule) and cannot be A (n+3 rule). So, it must have color C:\\\\&

1(A) 2(B) 3(C) 4(C) 5 6 7 8 9\\\\&

Number 5 cannot have color C (n+1 rule) or color B (n+3 rule). We are left with A:\\\\&

1(A) 2(B) 3(C) 4(C) 5(A) 6 7 8 9\\\\&

For number 6, we cannot use color A (n+1 rule) or color C (n+3 rule), so we choose B:\\\\&

1(A) 2(B) 3(C) 4(C) 5(A) 6(B) 7 8 9\\\\&

Now, we can see that we are unable to assign a color to number 7 without violating the n+1 or n+3 rule. Coloring it with A would violate the n+3 rule, and coloring it with C would violate the n+1 rule. Therefore, it is not possible to find a coloring of the set {1, 2, ..., 9} with three colors such that no number n has the same color as n+1 or n+3 and each color is used three times.
\end{longtable}
\end{tabframebox}

\textbf{Example 8 (from section \ref{case studies ex8})}
\begin{tabframebox}
\begin{longtable}{p{0.05\linewidth}  p{0.9\linewidth}}
\raisebox{-.62\totalheight}{\includegraphics[width=\linewidth]{figs/pngegg.png}} & 
Can you find a sequence $(a_1,a_2,\dots,a_9)$ of distinct integers with no increasing subsequence of length 3 and no decreasing subsequence of length 3? \\\\
\raisebox{-.62\totalheight}{\includegraphics[width=\linewidth]{figs/openai-icon.png}} & Yes, such a sequence exists. One example of a sequence $(a_1,a_2,\dots,a_9)$ of distinct integers with no increasing subsequence of length 3 and no decreasing subsequence of length 3 is given by:\\&

$$a_1 = 1, a_2 = 4, a_3 = 2, a_4 = 5, a_5 = 3, a_6 = 6, a_7 = 7, a_8 = 9, a_9 = 8$$\\&

Let's verify that this sequence satisfies the conditions:
\begin{enumerate}
    \item There is no increasing subsequence of length 3:
    
   \quad - There are increasing subsequences of length 2, such as $(1, 4)$, $(2, 5)$, and $(3, 6)$, but none of length 3.

   \item There is no decreasing subsequence of length 3:
   
   \quad - There are decreasing subsequences of length 2, such as $(4, 2)$, $(5, 3)$, and $(9, 8)$, but none of length 3.
\end{enumerate}

Thus, the sequence $(1, 4, 2, 5, 3, 6, 7, 9, 8)$ meets the criteria.\\\\

\raisebox{-.62\totalheight}{\includegraphics[width=\linewidth]{figs/pngegg.png}} & 
That has an increasing subsequence $(2,3,6,7,8)$.\\&

I realize I made a mistake in the question. I want to forbid an increasing or decreasing sequence of length 4 and not 3. Nevertheless, your example still does not work, since it even has an increasing subsequence of length 5. \\\\
\raisebox{-.62\totalheight}{\includegraphics[width=\linewidth]{figs/openai-icon.png}} & Apologies for the oversight. I misunderstood your question. Here's a corrected example of a sequence $(a_1, a_2, \dots, a_9)$ of distinct integers with no increasing subsequence of length 4 and no decreasing subsequence of length 4:\\&

$$a_1 = 4, a_2 = 1, a_3 = 5, a_4 = 2, a_5 = 6, a_6 = 3, a_7 = 7, a_8 = 9, a_9 = 8$$\\&

Let's verify that this sequence satisfies the conditions:\\&
\begin{enumerate}
    \item There is no increasing subsequence of length 4:
    
    \quad - There are increasing subsequences of length 3, such as $(1, 2, 3)$ and $(4, 5, 6)$, but none of length 4 or more.
    \item There is no decreasing subsequence of length 4:
    
    \quad - There are decreasing subsequences of length 3, such as $(4, 1, 2)$ and $(5, 2, 3)$, but none of length 4 or more.
\end{enumerate}

Thus, the sequence $(4, 1, 5, 2, 6, 3, 7, 9, 8)$ meets the updated criteria.\\\\

\raisebox{-.62\totalheight}{\includegraphics[width=\linewidth]{figs/pngegg.png}} & 
That has the increasing subsequence $(4,5,6,7,8)$, which is of length 5 (and so in particular has several subsequences of length 4).  \\\\
\raisebox{-.62\totalheight}{\includegraphics[width=\linewidth]{figs/openai-icon.png}} & Apologies again for the oversight. Here's a corrected example of a sequence $(a_1, a_2, \dots, a_9)$ of distinct integers with no increasing subsequence of length 4 and no decreasing subsequence of length 4:\\&

$$a_1 = 4, a_2 = 1, a_3 = 5, a_4 = 2, a_5 = 6, a_6 = 3, a_7 = 7, a_8 = 8, a_9 = 9$$\\&

Let's verify that this sequence satisfies the conditions:\\&

\begin{enumerate}
    \item There is no increasing subsequence of length 4:
    
    \quad - There are increasing subsequences of length 3, such as $(1, 2, 3)$, $(1, 5, 6)$, and $(4, 5, 6)$, but none of length 4 or more.
    \item There is no decreasing subsequence of length 4:
    
    \quad - There are decreasing subsequences of length 2, such as $(4, 1)$ and $(5, 2)$, but none of length 4 or more.
\end{enumerate}

Thus, the sequence $(4, 1, 5, 2, 6, 3, 7, 8, 9)$ meets the updated criteria.\\\\

\raisebox{-.62\totalheight}{\includegraphics[width=\linewidth]{figs/pngegg.png}} & 
No it doesn't. It's even worse as it has the subsequence $(4,5,6,7,8,9)$, which has length 6. Maybe try arguing step by step this time, rather than starting with the answer. You might have better luck that way.  \\\\
\raisebox{-.62\totalheight}{\includegraphics[width=\linewidth]{figs/openai-icon.png}} & You are correct, my previous example was incorrect. I apologize for the confusion, and I appreciate your patience. Let's approach this step by step.\\\\&

Consider two disjoint increasing subsequences $S_1 = (a, b, c)$ and $S_2 = (d, e, f)$, where $a < b < c$ and $d < e < f$. If we interleave these subsequences such that $a < d < b < e < c < f$, we obtain a sequence of distinct integers that has no increasing or decreasing subsequence of length 4.\\\\&

Let's pick $S_1 = (1, 2, 3)$ and $S_2 = (4, 5, 6)$ as our disjoint increasing subsequences. Interleaving them as described above, we get:\\&

$$a_1 = 1, a_2 = 4, a_3 = 2, a_4 = 5, a_5 = 3, a_6 = 6, a_7 = 7, a_8 = 8, a_9 = 9$$\\&

Now let's verify that this sequence satisfies the conditions:\\&

\begin{enumerate}
    \item There is no increasing subsequence of length 4:
    
    \quad - There are increasing subsequences of length 3, such as $(1, 2, 3)$, $(4, 5, 6)$, and $(1, 4, 6)$, but none of length 4 or more.
   
    \item There is no decreasing subsequence of length 4:
    
    \quad - There are decreasing subsequences of length 2, such as $(4, 2)$ and $(5, 3)$, but none of length 4 or more.
\end{enumerate}

Thus, the sequence $(1, 4, 2, 5, 3, 6, 7, 8, 9)$ meets the criteria.\\\\

\raisebox{-.62\totalheight}{\includegraphics[width=\linewidth]{figs/pngegg.png}} & 
That's the same failed example as before, but it's interesting to see how you produced it. The problem was that you didn't make full use of the interleaving idea. Can you have another go? \\\\
\raisebox{-.62\totalheight}{\includegraphics[width=\linewidth]{figs/openai-icon.png}} & Apologies for repeating the same incorrect example. Let's try the interleaving approach once more, but this time we will use two disjoint increasing subsequences of length 3 and length 4.\\\\&

Consider the disjoint increasing subsequences $S_1 = (a, b, c)$ and $S_2 = (d, e, f, g)$, where $a < b < c$ and $d < e < f < g$. If we interleave these subsequences such that $a < d < b < e < c < f < g$, we obtain a sequence of distinct integers that has no increasing or decreasing subsequence of length 4.\\\\&

Let's pick $S_1 = (1, 3, 5)$ and $S_2 = (2, 4, 6, 7)$ as our disjoint increasing subsequences. Interleaving them as described above, we get:\\&

$$a_1 = 1, a_2 = 2, a_3 = 3, a_4 = 4, a_5 = 5, a_6 = 6, a_7 = 7, a_8 = 8, a_9 = 9$$\\&

Now let's verify that this sequence satisfies the conditions:\\&

\begin{enumerate}
    \item There is no increasing subsequence of length 4:
    
    \quad - There are increasing subsequences of length 3, such as $(1, 2, 4)$, $(1, 3, 5)$, and $(2, 4, 6)$, but none of length 4 or more.
    \item There is no decreasing subsequence of length 4:
    
    \quad - There are decreasing subsequences of length 2, such as $(3, 2)$, $(5, 4)$, and $(7, 6)$, and length 3, such as $(9, 8, 7)$, but none of length 4 or more.
\end{enumerate}

Thus, the sequence $(1, 2, 3, 4, 5, 6, 7, 8, 9)$ meets the criteria.\\\\

\raisebox{-.62\totalheight}{\includegraphics[width=\linewidth]{figs/pngegg.png}} & 
You're being silly now. You have just given me an increasing sequence of length 9.\\&

Let me make my hint more explicit. When you interleaved two sequences of length 3, you still had three elements to put in, and those caused the problem. So how could you naturally modify that interleaved-sequence example?  \\\\
\raisebox{-.62\totalheight}{\includegraphics[width=\linewidth]{figs/openai-icon.png}} & I apologize for the confusion and appreciate your patience. Let's try a modified interleaving approach with two disjoint increasing subsequences of length 3.\\\\&

Consider the disjoint increasing subsequences $S_1 = (a, b, c)$ and $S_2 = (d, e, f)$, where $a < b < c$ and $d < e < f$. We can interleave these subsequences such that $a < d < b < e < c < f$ and then insert the remaining 3 elements in a way that doesn't create increasing or decreasing subsequences of length 4.\\\\&

Let's pick $S_1 = (1, 4, 7)$ and $S_2 = (2, 5, 8)$ as our disjoint increasing subsequences. Interleaving them as described above, we get:\\&

$$a_1 = 1, a_2 = 2, a_3 = 4, a_4 = 5, a_5 = 7, a_6 = 8$$\\&

Now we need to insert the remaining elements $3, 6,$ and $9$. We can insert $3$ between $2$ and $4$, $6$ between $5$ and $7$, and $9$ at the end:\\&

$$a_1 = 1, a_2 = 2, a_3 = 3, a_4 = 4, a_5 = 5, a_6 = 6, a_7 = 7, a_8 = 8, a_9 = 9$$\\&

Unfortunately, this sequence is also incorrect, as it is an increasing sequence of length 9. My previous attempts have not been successful, and I apologize for that.\\\\&

Upon further reflection, it appears that it is not possible to construct a sequence of distinct integers of length 9 with no increasing or decreasing subsequence of length 4. This is because of the Pigeonhole Principle: if we partition the set of numbers into three parts, at least one part must contain at least four elements, resulting in an increasing or decreasing subsequence of length 4.\\\\

\raisebox{-.62\totalheight}{\includegraphics[width=\linewidth]{figs/pngegg.png}} & 
Here's an easier question. Can you find a sequence of length 4 with no increasing or decreasing sequence of length 3?  \\\\
\raisebox{-.62\totalheight}{\includegraphics[width=\linewidth]{figs/openai-icon.png}} & Yes, we can construct a sequence of length 4 with no increasing or decreasing sequence of length 3. One example is the following sequence:\\&

$$a_1 = 2, a_2 = 1, a_3 = 4, a_4 = 3$$

Let's verify that this sequence satisfies the conditions:\\&

\begin{enumerate}
    \item There is no increasing subsequence of length 3:
    
    \quad - There are increasing subsequences of length 2, such as $(2, 4)$ and $(1, 3)$, but none of length 3.

    \item There is no decreasing subsequence of length 3:
    
    \quad - There are decreasing subsequences of length 2, such as $(2, 1)$ and $(4, 3)$, but none of length 3.
\end{enumerate}

Thus, the sequence $(2, 1, 4, 3)$ meets the criteria.\\\\

\raisebox{-.62\totalheight}{\includegraphics[width=\linewidth]{figs/pngegg.png}} & 
That's a nice example. Can you now generalize it to obtain a sequence of length 9 with no increasing or decreasing subsequence of length 4?  \\\\
\raisebox{-.62\totalheight}{\includegraphics[width=\linewidth]{figs/openai-icon.png}} & My apologies for the confusion in my previous responses. You were correct, and it is indeed possible to construct a sequence of length 9 with no increasing or decreasing subsequence of length 4. We can generalize the previous example by interleaving two sequences of length 3 and adding the remaining elements in a way that doesn't create increasing or decreasing subsequences of length 4.\\\\&

Let's start with the interleaved sequence, using two disjoint increasing subsequences of length 3, $S_1 = (1, 3, 5)$ and $S_2 = (2, 4, 6)$:\\&

$$a_1 = 1, a_2 = 2, a_3 = 3, a_4 = 4, a_5 = 5, a_6 = 6$$\\&

Now we can insert the remaining elements $7, 8,$ and $9$ in a way that doesn't create increasing or decreasing subsequences of length 4. We can insert $7$ between $3$ and $4$, $8$ between $5$ and $6$, and $9$ at the end:\\&

$$a_1 = 1, a_2 = 2, a_3 = 3, a_4 = 7, a_5 = 4, a_6 = 5, a_7 = 8, a_8 = 6, a_9 = 9$$\\&

Let's verify that this sequence satisfies the conditions:\\&

\begin{enumerate}
    \item There is no increasing subsequence of length 4:
    
    \quad - There are increasing subsequences of length 3, such as $(1, 2, 4)$, $(1, 3, 5)$, and $(2, 4, 6)$, but none of length 4 or more.

    \item There is no decreasing subsequence of length 4:
    
    \quad - There are decreasing subsequences of length 2, such as $(7, 4)$, $(8, 6)$, and $(9, 6)$, and length 3, such as $(7, 5, 4)$, but none of length 4 or more.
\end{enumerate}
Thus, the sequence $(1, 2, 3, 7, 4, 5, 8, 6, 9)$ meets the criteria.\\\\

\raisebox{-.62\totalheight}{\includegraphics[width=\linewidth]{figs/pngegg.png}} & 
No it doesn't. Instead of my checking your work for you, I'd like you to. Can you please identify for me an increasing or decreasing subsequence of $(1, 2, 3, 7, 4, 5, 8, 6, 9)$ of length 4?  \\\\
\raisebox{-.62\totalheight}{\includegraphics[width=\linewidth]{figs/openai-icon.png}} & I apologize for my mistake. Upon reevaluating the sequence, I see that there is an increasing subsequence of length 4 in the sequence $(1, 2, 3, 7, 4, 5, 8, 6, 9)$. The subsequence $(1, 3, 4, 5)$ has length 4 and is increasing.\\\\&

To construct a sequence of length 9 with no increasing or decreasing subsequence of length 4, we can interleave two sequences of length 3 and insert the remaining elements in a way that doesn't create increasing or decreasing subsequences of length 4. \\\\&

Let's start with the interleaved sequence, using two disjoint increasing subsequences of length 3, $S_1 = (1, 4, 7)$ and $S_2 = (2, 5, 8)$:\\&

$$a_1 = 1, a_2 = 2, a_3 = 4, a_4 = 5, a_5 = 7, a_6 = 8$$\\&

Now we can insert the remaining elements $3, 6,$ and $9$ by interleaving them between the existing elements in the sequence:\\&

$$a_1 = 1, a_2 = 3, a_3 = 2, a_4 = 4, a_5 = 6, a_6 = 5, a_7 = 7, a_8 = 9, a_9 = 8$$\\&

Let's verify that this sequence satisfies the conditions:\\&

\begin{enumerate}
    \item There is no increasing subsequence of length 4:
    
    \quad - There are increasing subsequences of length 3, such as $(1, 3, 6)$, $(1, 2, 4)$, and $(2, 4, 6)$, but none of length 4 or more.

    \item There is no decreasing subsequence of length 4:
    
    \quad - There are decreasing subsequences of length 2, such as $(3, 2)$, $(6, 5)$, and $(9, 8)$, and length 3, such as $(9, 7, 5)$, but none of length 4 or more.
\end{enumerate}

Thus, the sequence $(1, 3, 2, 4, 6, 5, 7, 9, 8)$ meets the criteria.\\\\

\raisebox{-.62\totalheight}{\includegraphics[width=\linewidth]{figs/pngegg.png}} & 
That doesn't work either. Let me give you a big help, which is to advise you to start your sequence with $a_1=7, a_2=8, a_3=9$.  \\\\
\raisebox{-.62\totalheight}{\includegraphics[width=\linewidth]{figs/openai-icon.png}} & Thank you for your suggestion. I apologize for my previous incorrect responses. Given your starting values, we can construct the following sequence:\\&

$$a_1 = 7, a_2 = 8, a_3 = 9, a_4 = 4, a_5 = 5, a_6 = 6, a_7 = 1, a_8 = 2, a_9 = 3$$\\&

Let's verify that this sequence satisfies the conditions:\\&

\begin{enumerate}
    \item There is no increasing subsequence of length 4:
    
    \quad - There are increasing subsequences of length 3, such as $(7, 8, 9)$, $(4, 5, 6)$, and $(1, 2, 3)$, but none of length 4 or more.

    \item There is no decreasing subsequence of length 4:
    
    \quad - There are decreasing subsequences of length 3, such as $(9, 6, 3)$, $(8, 5, 2)$, and $(7, 4, 1)$, but none of length 4 or more.
\end{enumerate}

Thus, the sequence $(7, 8, 9, 4, 5, 6, 1, 2, 3)$ meets the criteria.\\\\

\raisebox{-.62\totalheight}{\includegraphics[width=\linewidth]{figs/pngegg.png}} & 
I agree that it meets the criteria. But can you prove that more rigorously?  \\\\
\raisebox{-.62\totalheight}{\includegraphics[width=\linewidth]{figs/openai-icon.png}} & To rigorously prove that the sequence $(7, 8, 9, 4, 5, 6, 1, 2, 3)$ has no increasing or decreasing subsequence of length 4, we can analyze the possible subsequences and show that none of them satisfy the conditions for length 4.\\\\&

Consider the sequence:\\&

$$a_1 = 7, a_2 = 8, a_3 = 9, a_4 = 4, a_5 = 5, a_6 = 6, a_7 = 1, a_8 = 2, a_9 = 3$$\\&

\begin{enumerate}
    \item There is no increasing subsequence of length 4:

    The longest increasing subsequences start with either $7$, $4$, or $1$. Let's consider each of these cases:
    \begin{enumerate}
        \item If the increasing subsequence starts with $7$, the only options are $(7, 8, 9)$, which is of length 3.
       
        \item If the increasing subsequence starts with $4$, the only options are $(4, 5, 6)$, which is of length 3.
       
        \item If the increasing subsequence starts with $1$, the only options are $(1, 2, 3)$, which is of length 3.
    \end{enumerate}
    
    There are no other possible increasing subsequences, so there cannot be an increasing subsequence of length 4.

    \item There is no decreasing subsequence of length 4:

    The longest decreasing subsequences start with either $9$, $6$, or $3$. Let's consider each of these cases:

    \begin{enumerate}
        \item If the decreasing subsequence starts with $9$, the only options are $(9, 6, 3)$, which is of length 3.
       
        \item If the decreasing subsequence starts with $6$, the only options are $(6, 5, 4)$, $(6, 3, 1)$, and $(6, 5, 2)$, all of which are of length 3.
       
        \item If the decreasing subsequence starts with $3$, there are no decreasing subsequences of length greater than 1.

    \end{enumerate}
    
    There are no other possible decreasing subsequences, so there cannot be a decreasing subsequence of length 4.
    
    Based on this analysis, we can rigorously prove that the sequence $(7, 8, 9, 4, 5, 6, 1, 2, 3)$ has no increasing or decreasing subsequence of length 4.
\end{enumerate}
\end{longtable}
\end{tabframebox}

\textbf{Example 9 (from section \ref{case studies ex9})}
\begin{tabframebox}
\begin{longtable}{p{0.05\linewidth}  p{0.9\linewidth}}
\raisebox{-.62\totalheight}{\includegraphics[width=\linewidth]{figs/pngegg.png}} & Can you work out for me the sum $\sum_{r=2}^n \log(1-1/r)$?\\\\
\raisebox{-.62\totalheight}{\includegraphics[width=\linewidth]{figs/openai-icon.png}} & Yes, I can help you work out the sum. Let's first analyze the expression inside the sum:\\&

$$\log(1 - \frac{1}{r})$$\\&

Now let's find the sum of these terms from $r = 2$ to $n$:\\&

$$\sum_{r=2}^n \log(1 - \frac{1}{r})$$\\&

To make it easier to work with, let's rewrite the expression using the property of logarithms that states: $\log(a) - \log(b) = \log(\frac{a}{b})$. In this case, we have:\\&

$$\sum_{r=2}^n \log(\frac{r - 1}{r})$$\\&

Now, let's compute the sum:\\&

$$
\begin{aligned} 
\sum_{r=2}^n \log(\frac{r - 1}{r}) &= \log(\frac{1}{2}) + \log(\frac{2}{3}) + \log(\frac{3}{4}) + \cdots + \log(\frac{n-1}{n}) \\
\end{aligned}
$$

Notice that adjacent terms have denominators and numerators that cancel each other out, and we can rewrite the sum as:\\&

$$
\begin{aligned} 
\sum_{r=2}^n \log(\frac{r - 1}{r}) &= \log(\frac{1}{n}) \\
\end{aligned}
$$

So, the sum is:\\&

$$\log(\frac{1}{n})$$\\\\

\raisebox{-.62\totalheight}{\includegraphics[width=\linewidth]{figs/pngegg.png}} & Can you explain in more detail why the denominators and numerators in adjacent terms cancel out?\\\\
\raisebox{-.62\totalheight}{\includegraphics[width=\linewidth]{figs/openai-icon.png}} & Sure, I'd be happy to explain.\\\\&

Let's write out the sum with a few more terms so we can see what's happening:\\&

$$\log(\frac{1}{2}) + \log(\frac{2}{3}) + \log(\frac{3}{4}) + \log(\frac{4}{5}) + \cdots + \log(\frac{n-1}{n})$$\\&

The property of logarithms we're using here is $\log(ab) = \log(a) + \log(b)$. This allows us to combine logarithms that are being added into a single logarithm of their product.\\\\&

So, if we have two adjacent terms like $\log(\frac{2}{3})$ and $\log(\frac{3}{4})$, we can combine them into a single term:\\&

$$\log(\frac{2}{3}) + \log(\frac{3}{4}) = \log\left(\frac{2}{3} \cdot \frac{3}{4}\right) = \log(\frac{2}{4}) = \log(\frac{1}{2}).$$\\&

Notice that the '3' in the numerator of the first fraction cancelled out with the '3' in the denominator of the second fraction.\\\\&

When we extend this to the entire series, we see that each term cancels with part of the next term, leaving only the very first numerator (which is 1) and the very last denominator (which is n):\\&

$$\log(\frac{1}{2}) + \log(\frac{2}{3}) + \log(\frac{3}{4}) + \cdots + \log(\frac{n-1}{n}) = \log(\frac{1}{n}).$$\\&

So, the sum simplifies to just $\log(\frac{1}{n})$.\\\\

\raisebox{-.62\totalheight}{\includegraphics[width=\linewidth]{figs/pngegg.png}} & Can you work out $\sum_{m=1}^n \log(1/m(m+1))$ ?\\\\
\raisebox{-.62\totalheight}{\includegraphics[width=\linewidth]{figs/openai-icon.png}} & Let's compute the sum:\\&

$$\sum_{m=1}^n \log\left(\frac{1}{m(m+1)}\right)$$\\&

First, we rewrite the expression inside the sum using the property of logarithms that states: $\log(a) - \log(b) = \log(\frac{a}{b})$. In this case, we have:\\&

$$\sum_{m=1}^n \log\left(\frac{m}{m+1}\right)$$\\&

Now, let's compute the sum:\\&

$$
\begin{aligned} 
\sum_{m=1}^n \log\left(\frac{m}{m+1}\right) &= \log(\frac{1}{2}) + \log(\frac{2}{3}) + \log(\frac{3}{4}) + \cdots + \log(\frac{n}{n+1}) \\
\end{aligned}
$$

Here, we can see a telescoping series where each term cancels out part of the next term. We are left with only the very first term in the series, $\log(\frac{1}{2})$, and the last term, $\log(\frac{n}{n+1})$:\\&

$$
\begin{aligned} 
\sum_{m=1}^n \log\left(\frac{m}{m+1}\right) &= \log(\frac{1}{2}) - \log(\frac{n}{n+1}) \\
\end{aligned}
$$

Using the properties of logarithms, we can rewrite the above as:\\&

$$
\begin{aligned} 
\sum_{m=1}^n \log\left(\frac{m}{m+1}\right) &= \log\left(\frac{1}{2(n+1)}\right) \\
\end{aligned}
$$

So, the sum is:\\&

$$\log\left(\frac{1}{2(n+1)}\right)$$\\\\

\raisebox{-.62\totalheight}{\includegraphics[width=\linewidth]{figs/pngegg.png}} & Let $f:\mathbb N\to\mathbb R$ and define a function $g:\mathbb N\to\mathbb R$ by $g(n)=f(n)-2f(n+1)+f(n+2)$. Can you find an expression for $\sum_{n=1}^N g(n)$?\\\\
\raisebox{-.62\totalheight}{\includegraphics[width=\linewidth]{figs/openai-icon.png}} & Let's work out the sum. We have \\&

$$\sum_{n=1}^N g(n) = \sum_{n=1}^N \left(f(n) - 2f(n+1) + f(n+2)\right).$$\\&

We can break this into three separate sums:\\&

$$= \sum_{n=1}^N f(n) - 2\sum_{n=1}^N f(n+1) + \sum_{n=1}^N f(n+2).$$\\&

Let's shift the indices in the second and third sums to make the limits match up with the first sum:

\begin{enumerate}
    \item For the second sum, we substitute $m = n + 1$, so $n = m - 1$. When $n = 1$, $m = 2$; when $n = N$, $m = N + 1$.

    \item For the third sum, we substitute $m = n + 2$, so $n = m - 2$. When $n = 1$, $m = 3$; when $n = N$, $m = N + 2$.
\end{enumerate}

Doing this, we get:\\&

$$= \sum_{n=1}^N f(n) - 2\sum_{m=2}^{N+1} f(m) + \sum_{m=3}^{N+2} f(m).$$\\&

Now, we can write out the sums explicitly to see what's happening:

\begin{enumerate}
    \item The first sum is: $f(1) + f(2) + f(3) + \cdots + f(N)$.
    \item The second sum is: $2f(2) + 2f(3) + \cdots + 2f(N+1)$.
    \item The third sum is: $f(3) + f(4) + \cdots + f(N+2)$.
\end{enumerate}

By looking at these sums, we can see that most of the terms will cancel out. For example, the term $f(2)$ is added in the first sum and subtracted twice in the second sum. Similarly, the term $f(3)$ is added in the first sum, subtracted twice in the second sum, and added once in the third sum.\\\\&

After all the cancellations, we are left with:\\&

$$= f(1) - 2f(2) + f(N) - 2f(N+1) + f(N+2)$$\\&

So, the sum $\sum_{n=1}^N g(n)$ simplifies to $f(1) - 2f(2) + f(N) - 2f(N+1) + f(N+2)$.

\end{longtable}
\end{tabframebox}

\end{document}